\documentclass[final,5p,times,twocolumn]{elsarticle}

\usepackage{graphicx}
\usepackage{subfigure}

\usepackage{amssymb}
\usepackage{amsfonts}
\usepackage{amsmath}
\usepackage{xfrac}

\usepackage{url}
\usepackage[dvipsnames]{xcolor}
\makeatletter
\newif\if@restonecol
\makeatother

\usepackage{bm}
\usepackage{multirow}
\usepackage{mathrsfs}
\usepackage{amsmath}

\usepackage[ruled,linesnumbered,lined]{algorithm2e}

\usepackage{boolexpr,pdftexcmds,trace}
\usepackage{array} 
\usepackage{pdflscape}

\newcommand{\N}{\ensuremath{\mathbb{N}}}
\newcommand{\R}{\ensuremath{\mathbb{R}}}

\allowdisplaybreaks

\begin{document}

\begin{frontmatter}



\title{The Responsibility Weighted Mahalanobis Kernel for Semi-Supervised Training of Support Vector Machines for Classification}
\author{Tobias Reitmaier}
\author{Bernhard Sick}

\address{Intelligent Embedded Systems Lab, University of Kassel\\
Wilhelmsh\"oher Allee 71 -- 73, 34121 Kassel, Germany\\ 
(e-mail: $\{\text{tobias.reitmaier,bsick}\}\text{@uni-kassel.de}$)}

\begin{abstract}
Kernel functions in support vector machines (SVM) are needed to assess the similarity of input samples in order to classify these samples, for instance. Besides standard kernels such as Gaussian (i.e., radial basis function, RBF) or polynomial kernels, there are also specific kernels tailored to 
consider
structure in the data for similarity assessment. In this article, we will capture structure in data by means of probabilistic mixture density models, for example Gaussian mixtures in the case of real-valued input spaces. From the distance measures that are inherently contained in these models, e.g., Mahalanobis distances in the case of Gaussian mixtures, we derive a new kernel, the responsibility weighted Mahalanobis (RWM) kernel. Basically, this kernel emphasizes the influence of model components from which any two samples that are compared are assumed to originate (that is, the ``responsible'' model components). 
We will see that this kernel outperforms the RBF kernel and other kernels capturing structure in data (such as the LAP kernel in Laplacian SVM) in many applications where partially labeled data are available, i.e., for semi-supervised training of SVM. Other key advantages are that the RWM kernel can easily be used with standard SVM implementations and training algorithms such as sequential minimal optimization, and heuristics known for the parametrization of RBF kernels in a C-SVM can easily be transferred to this new kernel. Properties of the RWM kernel are demonstrated with 20 benchmark data sets and an increasing percentage of labeled samples in the training data.
\end{abstract}


\end{frontmatter}

\section{Introduction}
\label{sec:intro}

\textit{Support vector machines (SVM)} are a standard technique for pattern classification \cite{duda01,Ham09,STC00}. 
Often, kernel functions such as \textit{radial basis functions (RBF)}, sigmoid functions, or polynomials are taken to build a \textit{kernel matrix} that basically assesses the similarity (e.g., by means of a distance measure) of any two samples in a training data set \cite{Bur98}. 
This kernel matrix is needed by training techniques such as \textit{sequential minimal optimization (SMO)} in order to parametrize an SVM, i.e., in order to find the support vectors and their respective weights. 

In specific application domains, such as time series classification or document classification, various attempts have been made to define appropriate kernel functions for the specific tasks. Also, a number of attempts have been made to capture structure in the training data and to consider that information in the training process, for example, by modifying the kernel matrix appropriately or by defining a data dependent kernel function. What do we mean with ``capturing structure in data''? Essentially, we want to identify the hidden mechanisms underlying the data generation process, e.g., by describing a certain manifold embedded within the feature space in which the sample data lives \cite{Bis06} or by clustering the sample data.

In this article, we propose a new approach to consider structure in sample data in the training process. This approach is based on a \textit{parametric density model} of the training data, i.e., a Gaussian mixture model in the case of a continuous (real-valued) input space of the classifier. From all Mahalanobis distances being part of the various Gaussian components in this parametric density model 
we derive a new similarity measure for any two points in the input space of the classifier (a dissimilarity measure, to be precise, as it yields low values for similar samples). This measure, which we call \textit{responsibility weighted Mahalanobis (RWM) similarity}, 
considers structure in the data captured by means of the density model which is trained in an unsupervised way (for example with a maximum likelihood approach such as \textit{expectation maximization} or \textit{variational Bayesian inference} \cite{Bis06}). In order to call a function ``distance'', literature often requires the properties of a metric. Thus, we call our new measure \textit{similarity} as the triangle inequality does not hold.

The key property of the new RWM similarity is that it emphasizes the influence of those model components of the mixture density models from which the two samples that we want to compare (i.e., assess their similarity) to determine the kernel matrix are assumed to originate. These components are termed to be ``\textit{responsible}'' for the respective samples. 

Then, the Euclidean distance in an RBF kernel is simply replaced by the new RWM (dis-)similarity to define a new kernel, the RWM kernel. This kernel can be used in a C-SVM for classification tasks, for instance.

As the parametric density models are built in an unsupervised way, the RWM kernel is perfectly suited for \textit{semi-\-super\-vised} \textit{learning (SSL)}, i.e., training of SVM with partially labeled data sets.

\begin{figure*}[htb]
\centering
\subfigure[Euclidean distance to labeled sample $\mathbf{x}_1$.]{\label{fig:exampleA}\includegraphics[width=0.31\textwidth]{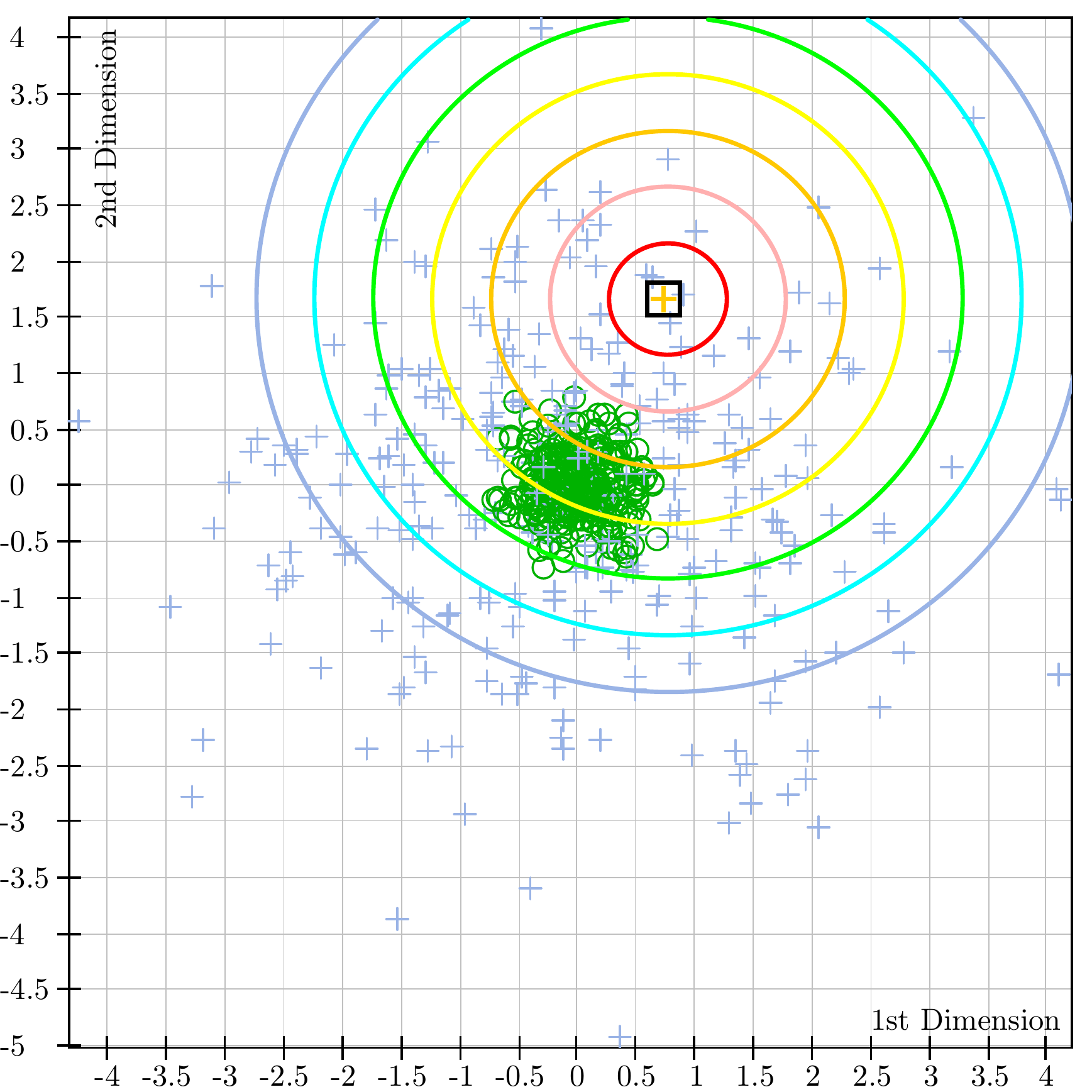}}\hfill
\subfigure[Euclidean distance to labeled sample $\mathbf{x}_2$.]{\label{fig:exampleB}\includegraphics[width=0.31\textwidth]{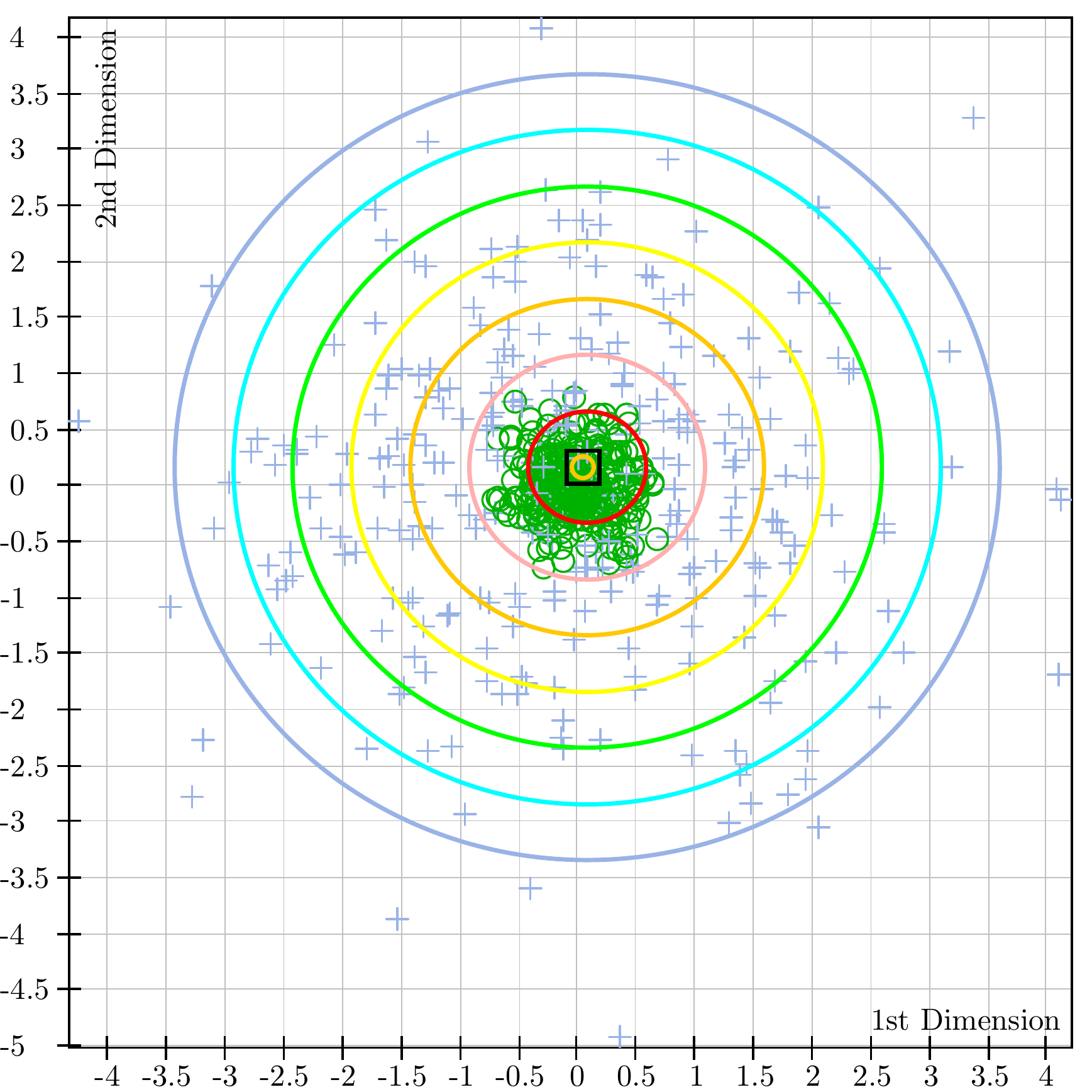}}\hfill
\subfigure[Resulting SVM classifier.]{\label{fig:exampleC}\includegraphics[width=0.31\textwidth]{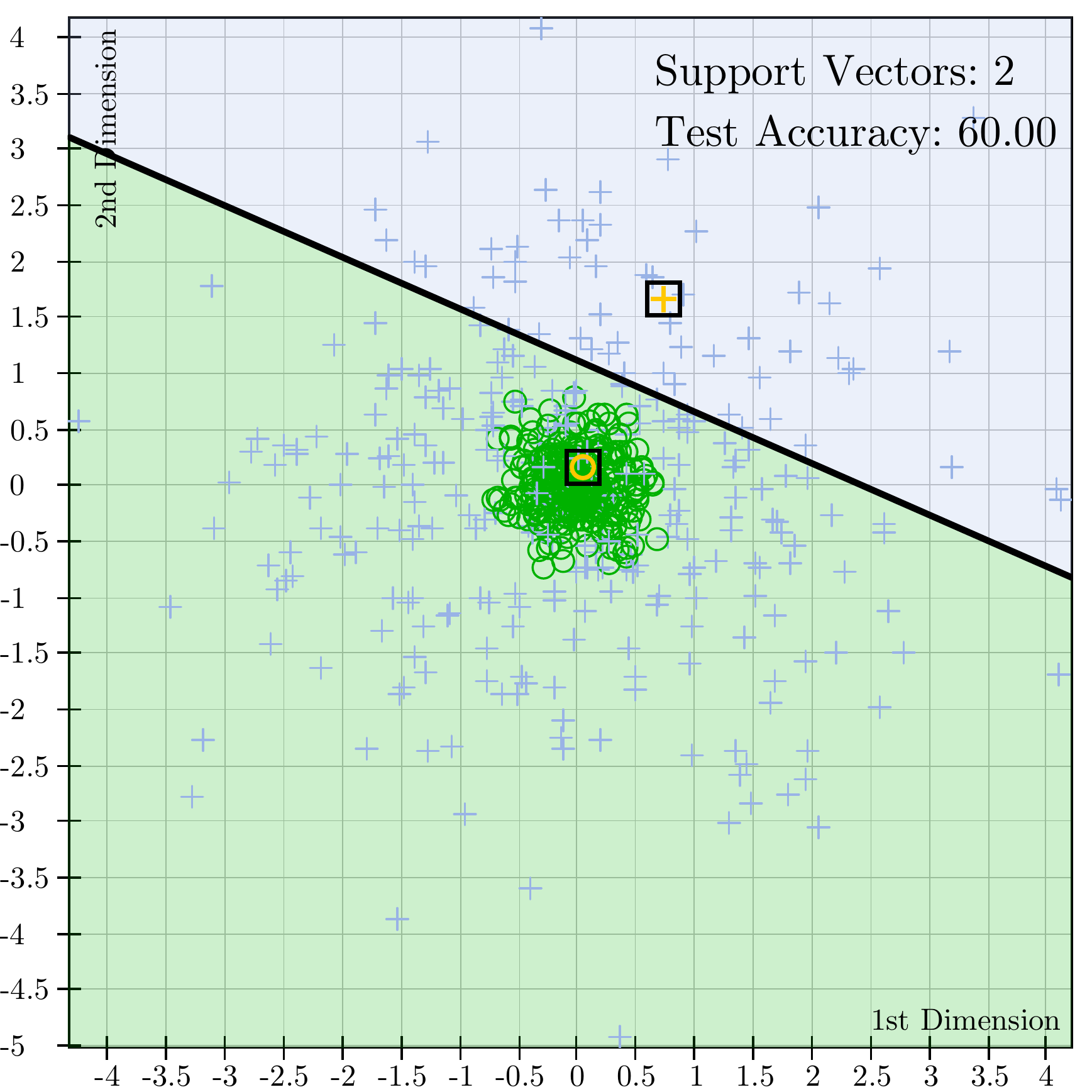}}
\vfill
\subfigure[RWM similarity to labeled sample $\mathbf{x}_1$.]{\label{fig:exampleD}\includegraphics[width=0.31\textwidth]{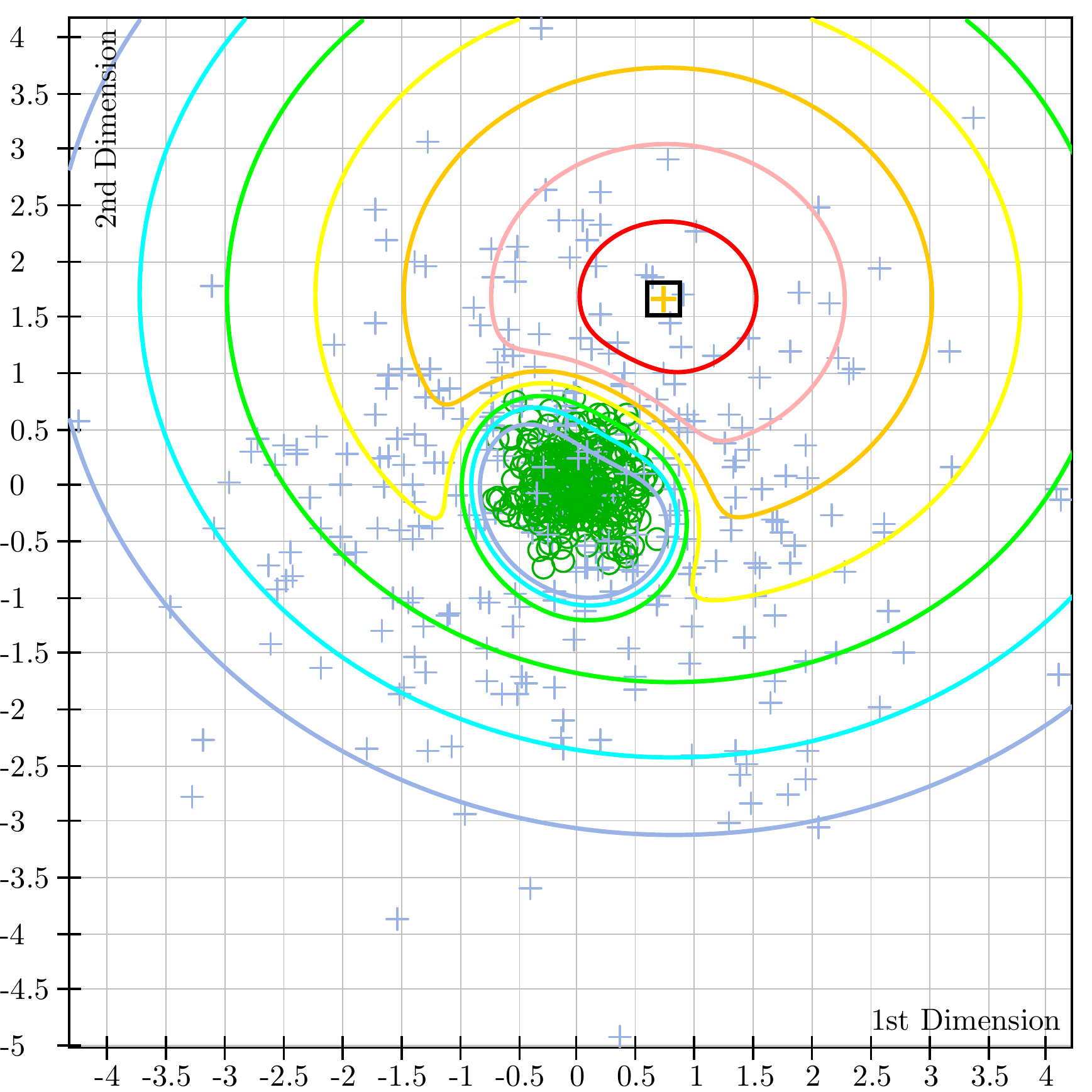}}\hfill
\subfigure[RWM similarity to labeled sample $\mathbf{x}_2$.]{\label{fig:exampleE}\includegraphics[width=0.31\textwidth]{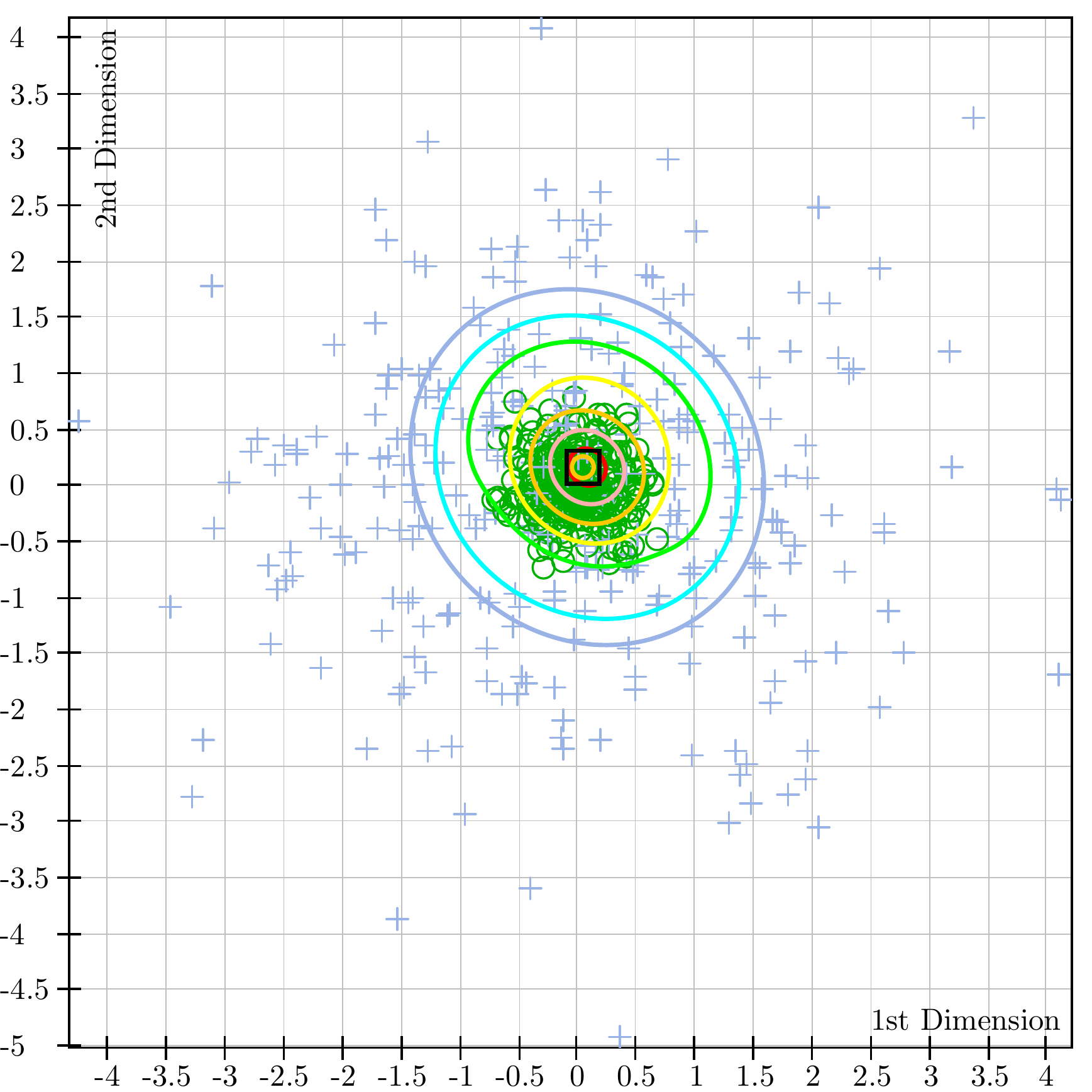}}\hfill
\subfigure[Resulting SVM classifier.]{\label{fig:exampleF}\includegraphics[width=0.31\textwidth]{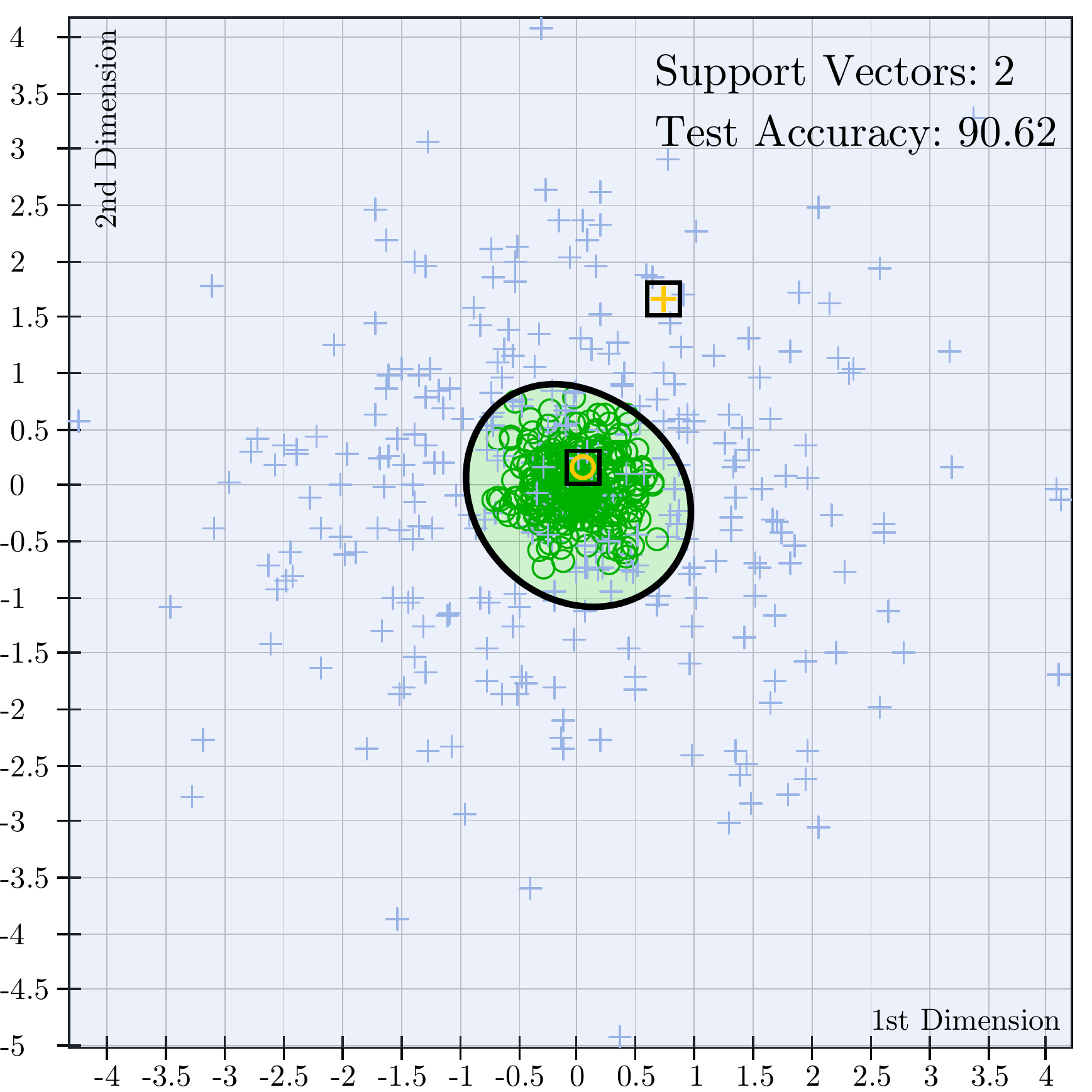}}
\caption{Binary classification problem: Two Gaussian processes (800 samples, 640 for training and 160 for test in a 5-fold cross-validation) of different classes (green circles and blue plus signs). The curves in parts (a), (b), (d), and (e) correspond to certain similarity or distance values between the two labeled samples $\mathbf{x}_i$  (orange colored) and all samples $\mathbf{y}$ with $\Delta(\mathbf{x}_i,\mathbf{y}) \in \left\{ 0.5, 1.0, 1.5,\dots,3.5 \right\}$ with $i \in \left\{1, 2\right\}$. The black solid line shown in parts (c) and (f) is the decision boundary of the resulting SVM classifier; the classification accuracy is the accuracy on test data.}
\label{intro_example}
\end{figure*}

In the following, we will illustrate the properties of the RWM similarity and the RWM kernel with a simple example.

Assume we observe two processes (or, depending on the point of view, one process consisting of two components) producing data in a two-dimensional input space (small blue plus signs and green circles; the two classes we want to recognize). For our example, we generated a set of samples using a Gaussian mixture model (GMM) with two components (GMM$_{gen}$: $\boldsymbol{\mu}_1 = \boldsymbol{\mu}_2 = (0.00,0.00)^{\text{T}}$, $\boldsymbol{\Sigma}_1 =\left(\begin{smallmatrix}0.07 & 0.00 \\ 0.00 & 0.07 \end{smallmatrix}\right)$, $\boldsymbol{\Sigma}_2 =\left(\begin{smallmatrix}1.93 & 0.00 \\ 0.00 & 1.93 \end{smallmatrix}\right)$, $\pi_{1}=\pi_{2}=0.50$). The component densities highly overlap and we also say that the respective processes overlap. Assume that, initially, we do not have any label information for the observed samples and we reconstruct the data generating model from the sample data in a completely unsupervised way. The model of this estimate is again a GMM (GMM$_{est}$: $\boldsymbol{\mu}_1 = (0.00,0.00)^{\text{T}}$, $\boldsymbol{\mu}_2 = (-0.05,-0.05)^{\text{T}}$, $\boldsymbol{\Sigma}_1 =\left(\begin{smallmatrix}0.00 & -0.01 \\ -0.01 & 0.10 \end{smallmatrix}\right)$, $\boldsymbol{\Sigma}_2 =\left(\begin{smallmatrix}2.23 & -0.02 \\ -0.02 & 1.86 \end{smallmatrix}\right)$, $\pi_{1}=0.55$ and $\pi_{2}=0.45$). Both, the generating model and the estimated model are not shown in Fig.~\ref{intro_example} for sake of simplicity. The class labels are shown but not considered by the unsupervised training step. Then, assume we get labels for only two samples, one for each class (shown in orange color). We train two SVM using these two labeled samples, an SVM with RBF kernel based on a Euclidean distance and an SVM with RWM kernel based on the RWM similarity that considers the information from the unlabeled samples, too (by means of the GMM$_{est}$). In both cases, the two labeled samples which build the training data set become support vectors (small black rectangles). To illustrate the differences between the Euclidean distance and the RWM similarity, Figs.~\ref{intro_example}\subref{fig:exampleA} and \ref{intro_example}\subref{fig:exampleD} show some curves of constant distance / similarity of all points in the input space with regard to sample $\mathbf{x}_1$. Figs.~\ref{intro_example}\subref{fig:exampleB} and \ref{intro_example}\subref{fig:exampleE} show these curves for sample $\mathbf{x}_2$. It can clearly be seen how the RWM similarity considers the structure information contained in the GMM$_{est}$, while the Euclidean distance does not rely on this information. And so do the corresponding kernels as shown in Figs.~\ref{intro_example}\subref{fig:exampleC} and \ref{intro_example}\subref{fig:exampleF}. These figures illustrate the resulting SVM classifiers and the accuracy on the test data. Essentially, according to the Bayesian principle of risk minimization \cite{Bis06}, the decision boundaries (solid black lines) can be constructed from the intersection of corresponding distance / similarity curves with regard to the two support vectors. The RBF kernel does not use structure information and, thus, the decision boundary corresponds to the perpendicular bisector of the connecting line of the two labeled samples. The RWM kernel uses structure information and, thus, the decision boundary becomes a nearly ring-shaped closed curve. The SVM with RWM kernel clearly outperforms the SVM with RBF kernel regarding classification accuracy (about $91\%$ vs. $60\%$).

The RWM kernel has a number of advantages:
\begin{itemize}
	\item In the case of semi-supervised learning (SSL) it outperforms some other kernels that capture structure in data such as the Laplacian kernel (Laplacian SVM) \cite{MB11} that can be regarded as being based on non-parametric density estimates.
	\item Standard training techniques such as SMO and standard implementations of SVM such as \textsf{libsvm} \cite{libsvm} can be used with RWM kernels without any algorithmic adjustments or extensions as only the kernel matrices have to be provided.
	\item Such as C-SVM with RBF kernels, C-SVM with RWM kernels can easily be parametrized using existing heuristics relying on line search strategies in a two-dimensional parameter space. This does not hold for the Laplacian kernel, for example.
\end{itemize}

The remainder of this article is structured as follows: Section~\ref{sec:overview} gives an overview of related work. Section~\ref{sec:rwmkernel} sketches the density model, defines the RWM (dis-)similarity, proposes the new RWM kernel, and investigates their respective properties. Results of simulation experiments with 20 benchmark data sets are set out in Section~\ref{sec:results}. Finally, Section~\ref{sec:conclusion} summarizes the key findings and gives an outlook to future work.

\section{Related Work}
\label{sec:overview}

The RWM kernel is particularly advantageous for SSL of SVM. Thus, we focus on this aspect here.

In an SSL setting there is, typically, a large amount of unlabeled data (also referred to as set of instances, observations, or samples without classification targets, i.e., desired outputs) in conjunction with only a small subset of labeled data. SSL aims to find a classification function by considering both sets (labeled and unlabeled). A large number of algorithms have been proposed that capture structure information in unlabeled data to improve the classification performance, e.g., \cite{CSZ06,ZGBD09}. Many SSL algorithms make, explicitly or implicitly, at least one of the following two common assumptions on the marginal distribution (i.e., the distribution of the unlabeled data), that is used to determine the classification function \cite{MB11}. 

The first assumption, called \textit{cluster assumption} \cite{CWS02}, claims that two samples in the ``same" cluster (high density region) are more likely to have the same class label. One major class of algorithms that follows this idea are the \textit{distance metric learning} algorithms. These algorithms require a distance metric to compare samples. Often, distances between classes based on the Euclidean distance or its generalization, the Mahalanobis distance \cite{mahalanobis36}, are used. Distance metric learning algorithms share the idea to move similar input samples closer and dissimilar ones further away where similarity is generally defined through class membership \cite{XWC12}. For this purpose, convex optimization with pairwise constraints \cite{WL09} or gradient descent with soft neighborhood assignments \cite{SHWP02} are used. This often leads to a two-step approach: First the metric is learned, then it is used to train the classifier of choice, e.g., an SVM. Support Vector Metric Learning (SVML) \cite{XWC12} differs from this two-step scheme in that it learns a Mahalanobis metric to minimize the validation error of the SVM prediction at the same time the SVM is trained. A similar approach, presented in \cite{NY08}, solves the metric learning problem by quadratic programming with local neighborhood constraints based on the SVM framework. In addition, the cluster assumption implies that the decision boundary between two classes lies in lower density regions of the input space \cite{CWS02}. This conclusion is underlying the category of \textit{low-density separation} methods that try to place decision boundaries into lower density regions. One of the most frequently used algorithms in this class are transductive SVM \cite{Vapnik95} and their various implementations, e.g., TSVM \cite{Joachims99} and S$^3$VM~\cite{CSK08, RSM11, FGQZ11, AC11}. 

The second assumption, called \textit{manifold assumption} \cite{BNS06}, 
claims that the marginal distribution underlying the data can be described by means of
a manifold of much lower dimension than the input space, so that the distances and densities defined on this manifold can be used for learning \cite{BNS06}. A lot of \textit{graph based methods}, another major class of SSL techniques, have been proposed, but most of them only perform transductive inference \cite{Joachims03, BN02, ZGL03}, which means that they classify only the unlabeled training data. The Laplacian support vector machines (LapSVM) \cite{BNS06,WL09,NCW15} provide a natural out-of-sample extension to classify data that become available after the training process, without having to retrain the classifier. LapSVM follow the principle of manifold regularization by incorporating an ``intrinsic regularizer'' \cite{MB11} into the learning process that is empirically estimated from the labeled and unlabeled data using a Laplacian graph (nonparametric density estimator). It has been shown that LapSVM yield very good performance in semi-supervised classification \cite{MB11}. The last major class of methods are \textit{generative models}. SSL with generative models can be viewed as an extension of unsupervised learning (clustering plus some class label information). Here, often adapted versions of the well-known $c$-means algorithm or the more general expectation maximization (EM) algorithm for Gaussian mixture models are used. A detailed description of training algorithms and applications of various kinds of probabilistic mixture models is given in \cite{MP00}.

What do we intend to make better or in an other way? Our new SSL approach considers structure information provided by the unlabeled data by means of a parametric density model, i.e., a Gaussian mixture model in case of a continuous (real-valued) input space of the classifier. From all Mahalanobis distances being part of the Gaussians we derive a new kernel function, called responsibility weighted Mahalanobis (RWM) kernel. Basically, this kernel is based on Mahalanobis distances but it reinforces the impact of model components from which any two samples that are compared are assumed to originate.

 
\section{The RWM Kernel}
\label{sec:rwmkernel}

In this section we will first describe the density model which is based on mixtures of Gaussians for real-valued dimensions of the input space of the classifier. This model is basis of the RWM similarity which will then be defined and investigated. Next, we integrate this similarity into the new RWM kernel and explore its properties. Finally, we show how this approach can be extended to categorical input dimensions.

\subsection{Density Models Based on Gaussian Mixtures}
To capture structure information contained in (unlabeled) sample data, we build a density model from these sample data. We start from the assumption that we have a $D$-dimensional real-valued input space of the classifier and the training samples are realizations of a $D$-dimensional random variable $\mathbf{x} \in \R^D$. Then, the density function $p(\mathbf{x})$ will be modeled with $K$ \textit{components}
\begin{align}\label{eq:density}
p(\mathbf{x}) &= \sum_{k=1}^K p(\mathbf{x}, k)
= \sum_{k=1}^K p(k) p(\mathbf{x}|k)
\end{align}
using sum and product rules of probabilities. The $p(\mathbf{x}|k)$  are the component densities. Here, we assume that these conditional densities are modeled with \textit{multivariate normal densities} which can be motivated by the \textit{generalized central limit theorem} in many applications \cite{duda01}. That is, we realize Eq.~\eqref{eq:density} with
\begin{equation} 
p(\mathbf{x}|\boldsymbol\pi,\boldsymbol{\mu},\mathbf{\Sigma}) = \sum_{k=1}^{K}{\pi_{k} \, \mathcal{N}(\mathbf{x}|\boldsymbol{\mu}_{k},\mathbf{\Sigma}_{k})}
\end{equation}
where we have \textit{mixing coefficients} $\pi_k$ and multivariate Gaussians
\begin{equation}
\mathcal{N}(\mathbf{x}|\boldsymbol{\mu}_{k},\mathbf{\Sigma}_{k}) = \frac{1}{(2\pi)^{\frac{D}{2}} |\mathbf{\Sigma}_{k}|^{\frac{1}{2}}} \, \exp{\left(-\frac{1}{2}\left( \Delta_{\mathbf{\Sigma}_{k}}\left(\mathbf{x},\boldsymbol{\mu}_{k}\right)\right)^2\right)}
\end{equation}
with \textit{mean vectors} $\boldsymbol{\mu}_{k} \in \R^D$ and \textit{covariance matrices} $\boldsymbol{\Sigma}_{k} \in \R^{D\times D}$ ($\boldsymbol\pi,\boldsymbol{\mu},\mathbf{\Sigma}$ in $p(\mathbf{x}|\boldsymbol\pi,\boldsymbol{\mu},\mathbf{\Sigma})$ summarize all $\pi_k$, $\boldsymbol{\mu}_k$, and $\boldsymbol{\Sigma}_{k}$, respectively). Here, $|\cdot |$ denotes the determinant of a matrix. The matrix distance
\begin{equation}
\Delta_{\mathbf{M}}(\mathbf{x}_{i},\mathbf{x}_{j}) = \sqrt{\left(\mathbf{x}_{i} - \mathbf{x}_{j}\right)^{\mathrm{T}}\mathbf{M}^{-1}\left(\mathbf{x}_{i} - \mathbf{x}_{j}\right)}
\end{equation}
with $\mathbf{M} = \boldsymbol{\Sigma}_{k}$ is known as \textit{Mahalanobis distance} of vectors  $\mathbf{x}_{i}, \mathbf{x}_{j} \in \R^D$.
If $\boldsymbol{\Sigma}_{k}$ is the unit matrix (and, therefore, $\boldsymbol{\Sigma}_{k}^{-1}$ too), $\Delta_{\boldsymbol{\Sigma}_{k}}(\mathbf{x}_{i},\mathbf{x}_{j})$ is the \textit{Euclidean distance} which shows that the Mahalanobis distance contains the Euclidean distance as a special case. If $\boldsymbol{\Sigma}_{k}$ is a diagonal matrix, we get a \textit{scaled Euclidean distance}.

\begin{figure}[htb]
\centering
\includegraphics[width=0.45\textwidth]{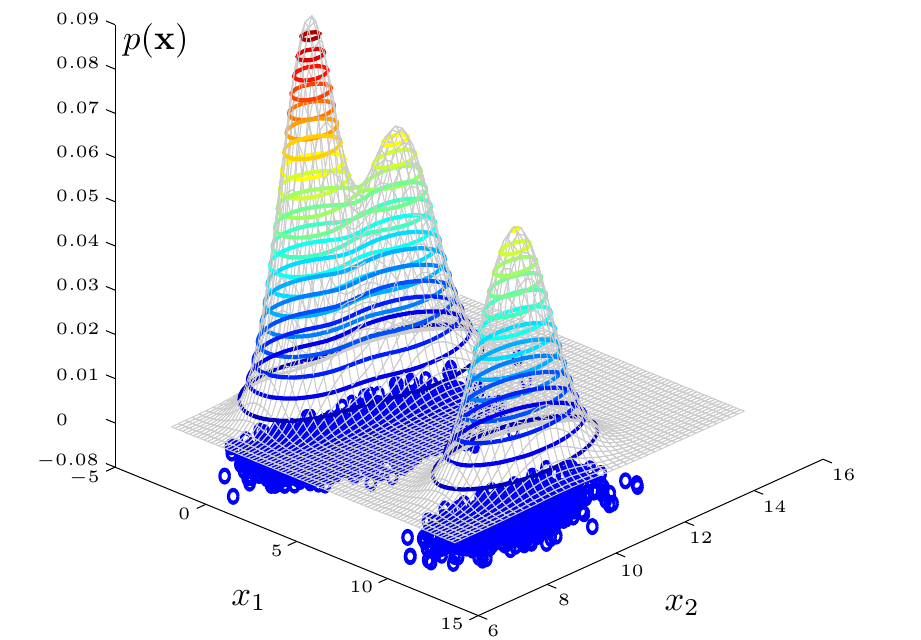}


\caption{Example for a GMM trained from sample data.}
\label{gmm-example}
\end{figure}

Fig.\ \ref{gmm-example} shows an example for a Gaussian mixture model (GMM) for $p(\mathbf{x}) = \sum_{k=1}^{3} p(k) p(\mathbf{x}|k)$ in a two-dimensional input space. The level curves correspond to the surfaces of constant density of $p(\mathbf{x})$. For a single Gaussian, such level curves have the shape of an ellipse (or a circle if the covariance matrix is isotropic).


With the $K$ components of such a model we aim at modeling $K$ processes in the real world that are said to ``generate'' the samples that we observe. For a given sample $\mathbf{x}' \in \R^D$ we do usually not know by which process it has been generated but we can estimate that by means of so-called \textit{responsibilities} (note the use of Bayes' theorem):
\begin{align}
\label{eq:resp}
 \rho_{\mathbf{x}',k} &= p(k|\mathbf{x}') \\
& = \frac{p(k) p(\mathbf{x}'|k)}{p(\mathbf{x}')} \nonumber \\
& =\frac{\pi_{k}\mathcal{N}(\mathbf{x}'|\boldsymbol{\mu}_{k},\mathbf{\Sigma}_{k})}{\sum_{j=1}^{K}\pi_{j}\mathcal{N}(\mathbf{x}'|\boldsymbol{\mu}_{j},\mathbf{\Sigma}_{j})}. \nonumber 
\end{align}
Thus, a responsibility $\rho_{\mathbf{x}',k}$ of a component $k$ for a sample $\mathbf{x}'$ can be seen as a gradual assignment of the sample $\mathbf{x}'$ to the component $k$ considering structure in the data. Note that $\sum_{k=1}^K  \rho_{\mathbf{x}',k} = 1$ and $\rho_{\mathbf{x}',k} > 0$.

The well-known $c$-means clustering, for example, can be seen as a special case of such an approach (cf.~\cite{Bis06} for details) where we have a unique assignment of samples to components (i.e., clusters).

How can the various parameters of the Gaussian mixture density model be determined in an \textit{unsupervised learning} approach, i.e., without using class labels of samples? Here, we perform the parameter estimation not with a standard \textit{expectation maximization (EM)} technique \cite{Bis06} but with a technique called \textit{variational Bayesian inference (VI)} which realizes the Bayesian idea of regarding the model parameters as random variables whose distributions must be trained. This approach has two important advantages. First, the estimation process is more robust, i.e., it avoids ``collapsing'' components, so-called singularities whose variance in one or more dimensions vanishes. Second, VI optimizes the number of components by its own. It starts with a large number of components and prunes components \textit{automatically} until a sufficient number of components $K$ is achieved. For a detailed discussion of Bayesian inference, and, particularly, the VI algorithm see \cite{Bis06, FS09, RCS14}. 


\subsection{The RWM Similarity Measure}
\label{subsec:similarity}

With the Mahalanobis distance measure described above we can determine a distance of any two samples in the $D$-di\-men\-sion\-al input space with respect to a process modeled by a single Gaussian component with given mean and covariance matrix. In general, however, we need a number of $K > 1$ components to model densities with sufficient accuracy. 

Assume we are given a density model GMM based on Gaussian mixtures as described above. Then, the following distance measure for any two samples $\mathbf{x}_i, \mathbf{x}_j \in \R^D$ with $i,j \in \N$ can be defined as described in \cite{Bis06}:
\begin{equation}
\Delta_{\mathbf{GMM}}(\mathbf{x}_{i},\mathbf{x}_{j}) = \sum_{k=1}^{K}\left(\pi_k  \Delta_{\mathbf{\boldsymbol{\Sigma}}_{k}}(\mathbf{x}_{i},\mathbf{x}_{j})  \right).
\end{equation}
This measure is zero from a sample to itself and positive for two different samples (positive definiteness), symmetric, and it fulfills the triangle inequality. Thus, this distance function is a \textit{metric}. A proof must exploit the fact that $\Delta_{\mathbf{\boldsymbol{\Sigma}}_{k}}(\mathbf{x}_{i},\mathbf{x}_{j})$ is a metric, too \cite{Run12}. In contrast to a Euclidean distance $\Delta_{\mathbf{EUC}}(\mathbf{x}_{i},\mathbf{x}_{j}) = \|\mathbf{x}_{i}-\mathbf{x}_{j}\|^2$, this \textit{GMM distance measure} considers the distances of two samples with respect to all processes contained in the Gaussian mixture model, weighted with their respective mixing coefficients. These mixing coefficients are related to the responsibilities as follows:
$ \pi_k = \sum_{n=1}^{N} \rho_{\mathbf{x}_n,k}, $
%
%
%
i.e., they are determined from $N$ training samples $\mathbf{x}_1, \ldots, \mathbf{x}_N$ in an unsupervised step (e.g., using VI).

In our new RWM similarity, however, we want to give more emphasis to the individual responsibilities of components for two given samples we want to assess. Thus, the Mahalanobis distance $\Delta_{\boldsymbol{\Sigma}_{k}}(\mathbf{x}_{i},\mathbf{x}_{j})$ gets a weight that depends on the responsibilities of $k$ for the two considered samples. According to this, the new \textit{responsibility weighted Mahalanobis (RWM) measure} can be defined with
\begin{equation}
\Delta_{\mathbf{RWM}}(\mathbf{x}_{i},\mathbf{x}_{j}) = \sum_{k=1}^{K}\left(\frac{1}{2} \left(\rho_{\mathbf{x}_i,k} + \rho_{\mathbf{x}_j,k}\right) \Delta_{\mathbf{\boldsymbol{\Sigma}}_{k}}(\mathbf{x}_{i},\mathbf{x}_{j})  \right).
\end{equation}
Basically, this measure is a dissimilarity measure as it yields high values for very distinct input samples. It can easily be shown that this measure is a \textit{semi-metric} according to \cite{Wil31}
as the properties of non-negativity, identity of indiscernibles, and symmetry still hold. The triangle inequality is dropped here. For a proof of the former properties it must be considered that responsibilities are non-negative for Gaussian mixtures. 

We will now investigate the properties of the RWM similarity in more detail.

\begin{figure*}[htbp!]
\centering
\subfigure[Euclidean distance.]{\includegraphics[width=0.31\textwidth]{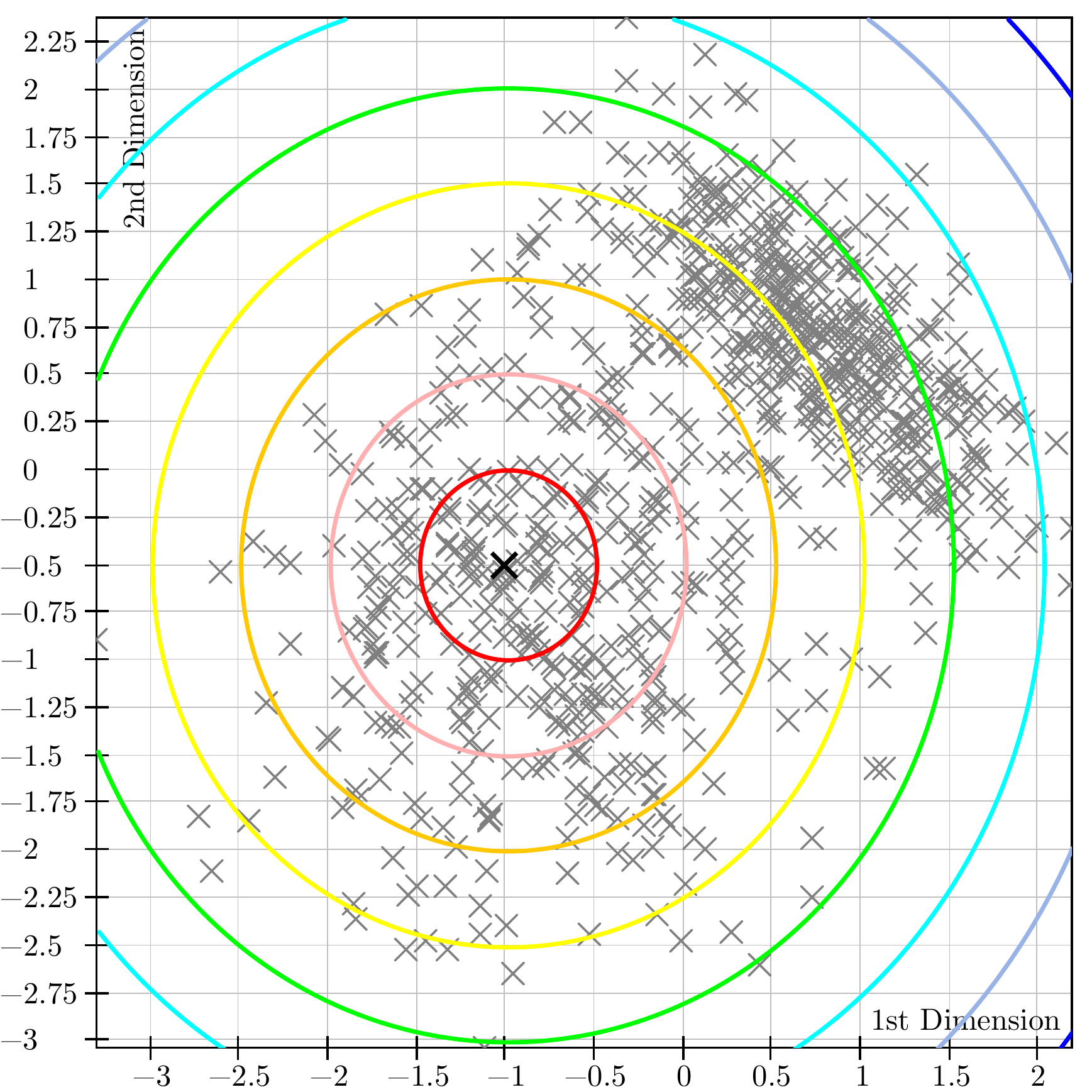}}\hfill
\subfigure[GMM distance.]{\includegraphics[width=0.31\textwidth]{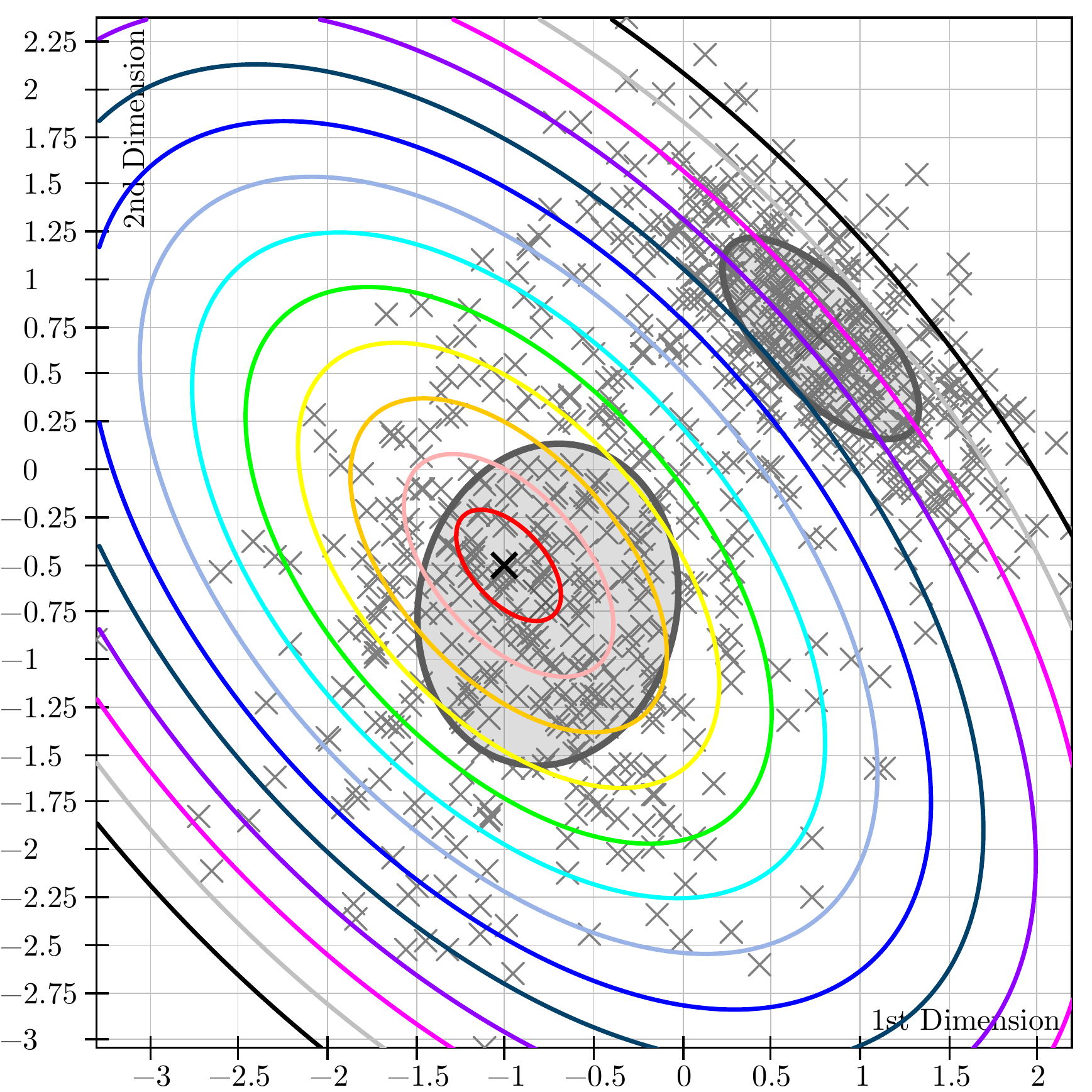}}\hfill
\subfigure[RWM similarity.]{\includegraphics[width=0.31\textwidth]{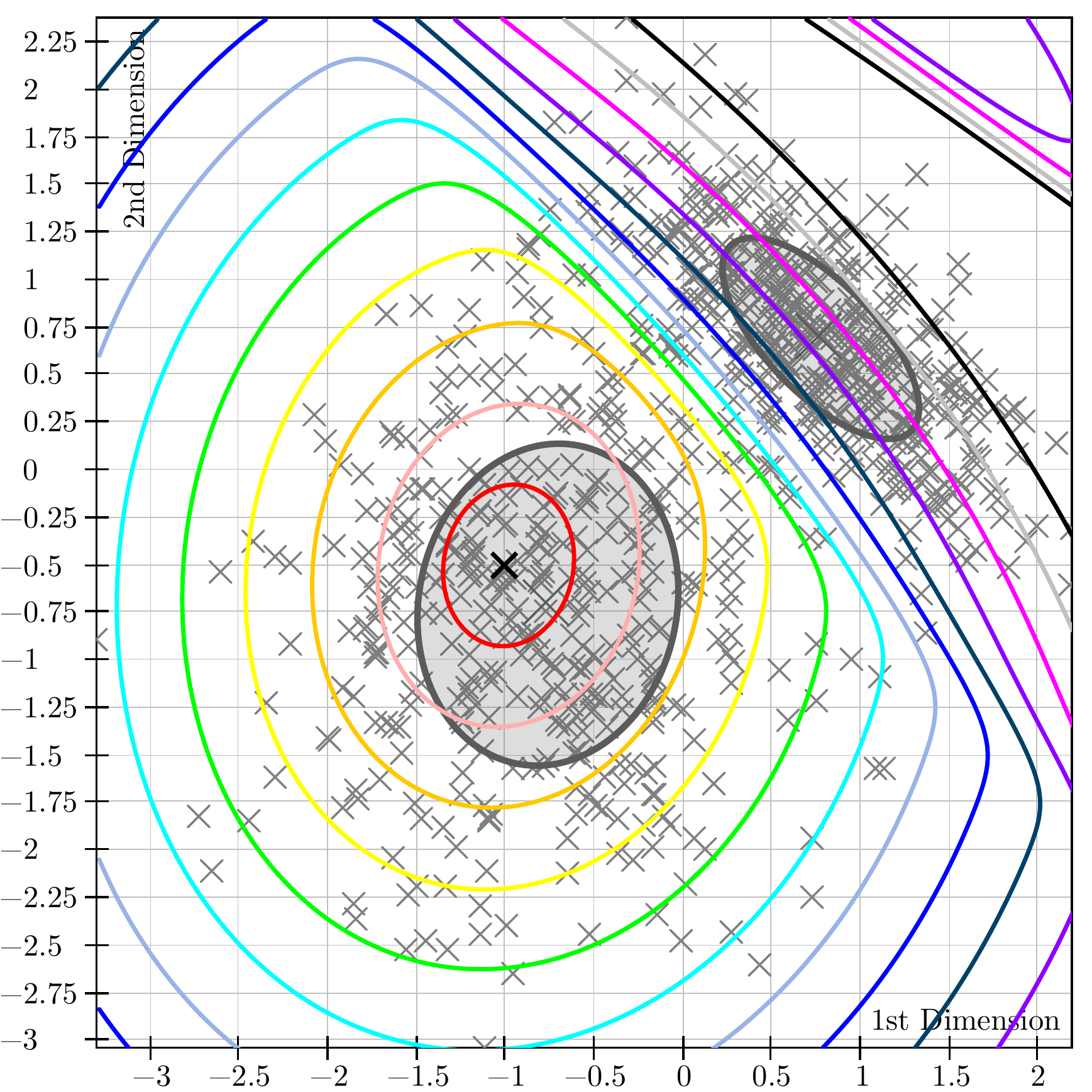}}\hfill
\caption{Comparison of measures that do not use structure information or use structure information in different ways. The level curves correspond to distances or similarity values between the sample $\mathbf{x}$ at position $(-1,-0.5)$ (thick black cross) and all samples $\mathbf{y}$ with $\Delta(\mathbf{x},\mathbf{y}) \in \left\{ 0.5, 1.0, 1.5,\dots,6.5 \right\}$.}
\label{distance_comparison}
\end{figure*}

First, we want to compare the Euclidean distance to the GMM distance and the RWM similarity.
For this purpose we use an synthetic data set generated by a mixture model consisting of two Gaussians with $\boldsymbol{\Sigma}_1 =\left(\begin{smallmatrix}0.46 & 0.04 \\ 0.04 & 0.54 \end{smallmatrix}\right)$, $\boldsymbol{\Sigma}_2 =\left(\begin{smallmatrix}-0.34 & -0.24 \\ -0.24 & 0.33 \end{smallmatrix}\right)$, $\boldsymbol{\mu}_{1}=(-0.78,-0.76)^{\text{T}}$, $\boldsymbol{\mu}_{2}=(-0.76,0.75)^{\text{T}}$ and $\pi_1=\pi_2=0.50$. Fig.~\ref{distance_comparison} shows the different behavior of three measures, two of them use structure information for similarity or distance measurement. The depicted ellipses with gray background are level curves of the two Gaussian components of a mixture model estimated from the sample data that are located at centers indicated by large $\times$s. All samples on such a level curve have a Mahalanobis distance of one to the respective center. In the area between the two Gaussians, the gradient of the RWM similarity function is higher than the gradient of the GMM distance (the level curves of the similarity measure are more dense). The reason is that the RWM similarity considers the local structure of the data as it emphasizes the responsibilities (cf.\ Eq.~\eqref{eq:resp}) of the two samples under consideration.

\begin{figure*}[phtb!]
\centering
\subfigure[\scriptsize$\mathbf{\Sigma}_{1}=\mathbf{\Sigma}_{2}=\mathbf{\Sigma}_{3}=\mathbf{I}$.]{\includegraphics[width=0.31\textwidth]{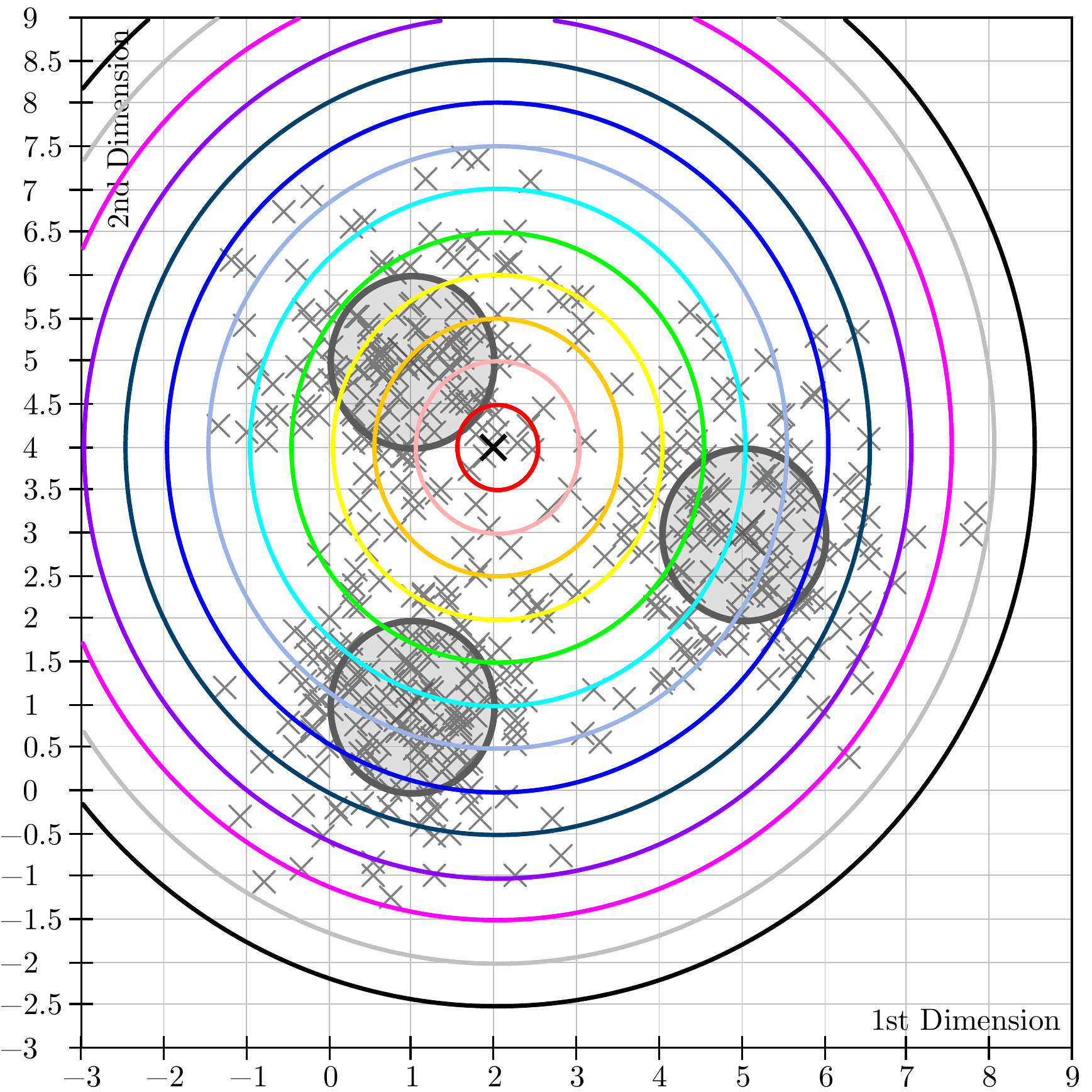}}\hfill
\subfigure[\scriptsize$\mathbf{\Sigma}_{1}=\mathbf{I};\mathbf{\Sigma}_{2}=\mathbf{\Sigma}_{3}=0.25\mathbf{I}$.]{\includegraphics[width=0.31\textwidth]{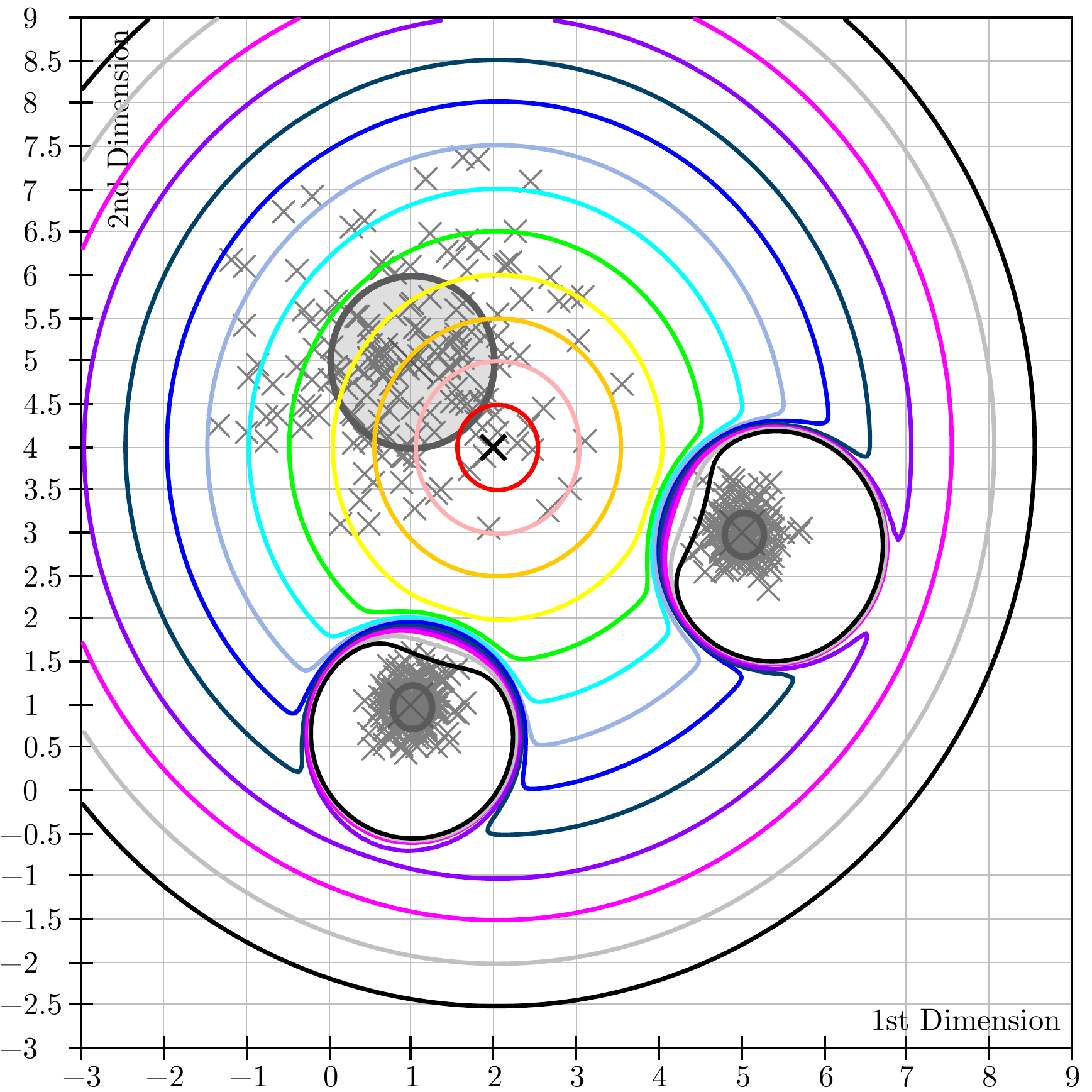}}\hfill
\subfigure[\scriptsize$\mathbf{\Sigma}_{1}=\mathbf{I};\mathbf{\Sigma}_{2}=0.5\mathbf{I};\mathbf{\Sigma}_{3}=2.5\mathbf{I}$.]{\includegraphics[width=0.31\textwidth]{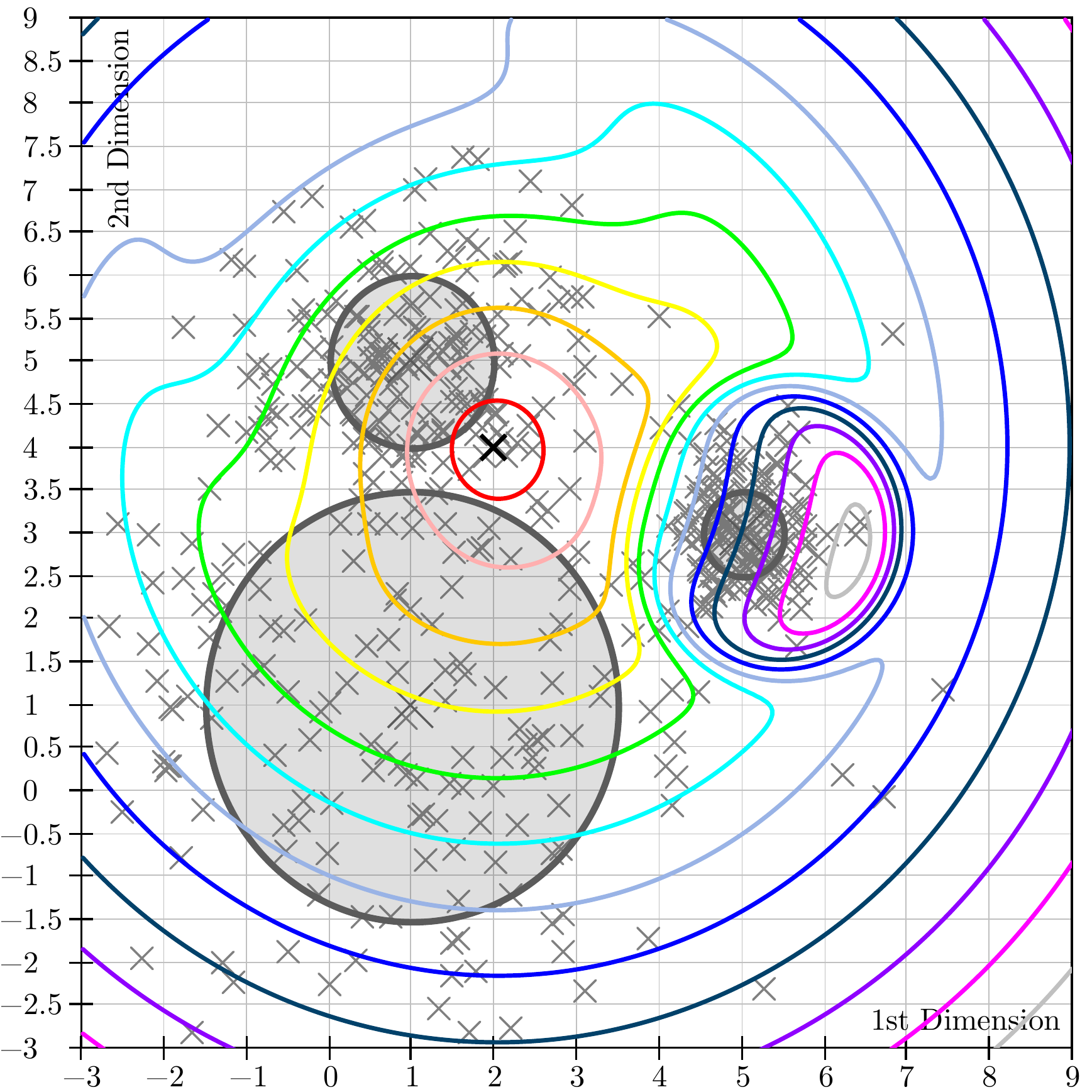}}
\caption{Influence of the scaling of isotropic covariance matrices on the RWM similarity. The level curves correspond to RWM similarity values between the sample $\mathbf{x}$ at position $(2,4)$ (thick black cross) and all samples $\mathbf{y}$ with $\Delta_{\mathbf{RWM}}(\mathbf{x},\mathbf{y}) \in \left\{ 0.5, 1.0, 1.5,\dots,6.5 \right\}$.}
\label{hmm2}
\end{figure*}


%

\begin{figure*}[phtb!]
\centering
\subfigure[\scriptsize $\pi_{1}=\pi_{2}=\pi_{3}=0.33$.]{\includegraphics[width=0.31\textwidth]{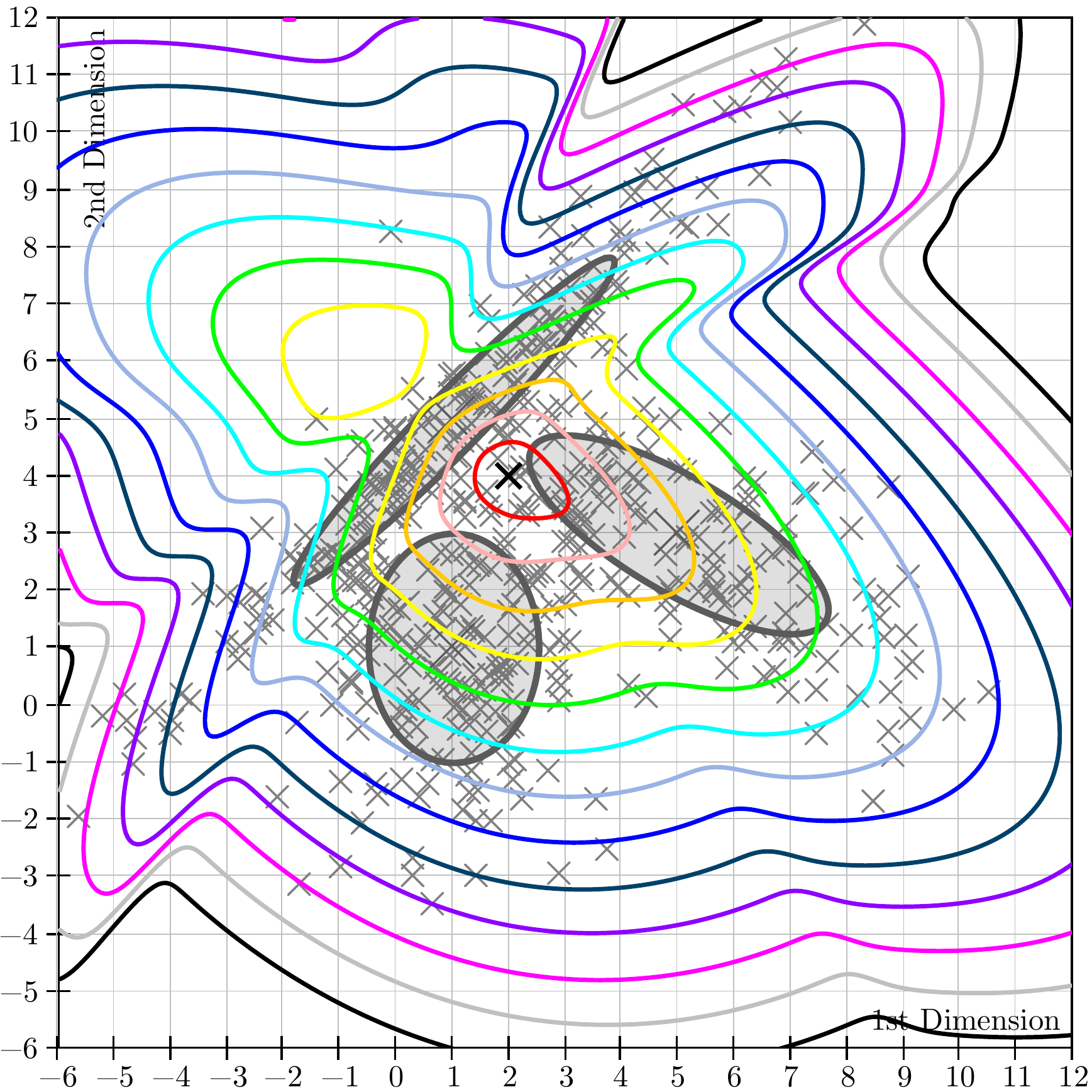}}\hfill
\subfigure[\scriptsize $\pi_{1}=0.9;\pi_{2}=\pi_{3}=0.05$.]{\includegraphics[width=0.31\textwidth]{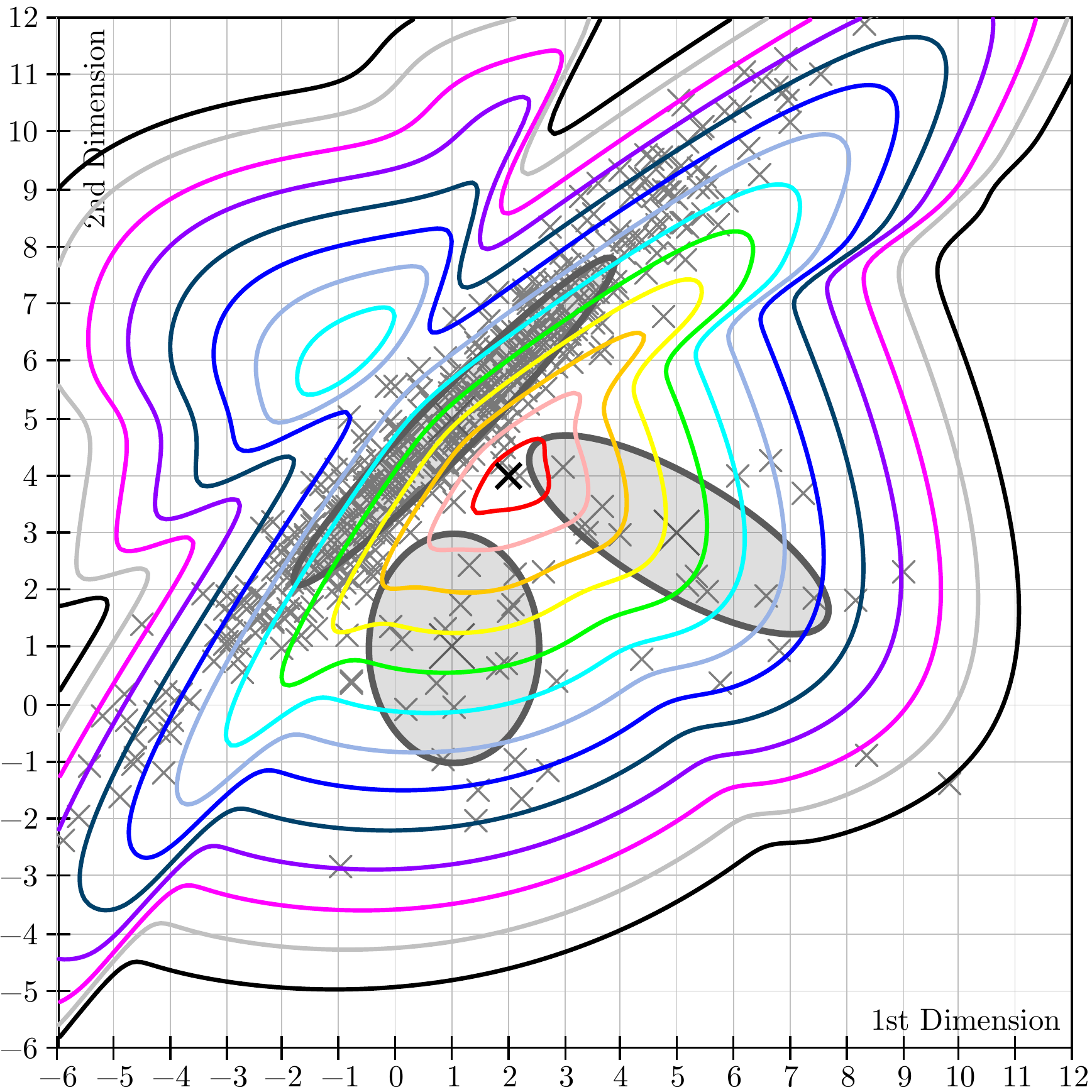}}\hfill
\subfigure[\scriptsize $\pi_{1}=0.05;\pi_{2}=\pi_{3}=0.475$.]{\includegraphics[width=0.31\textwidth]{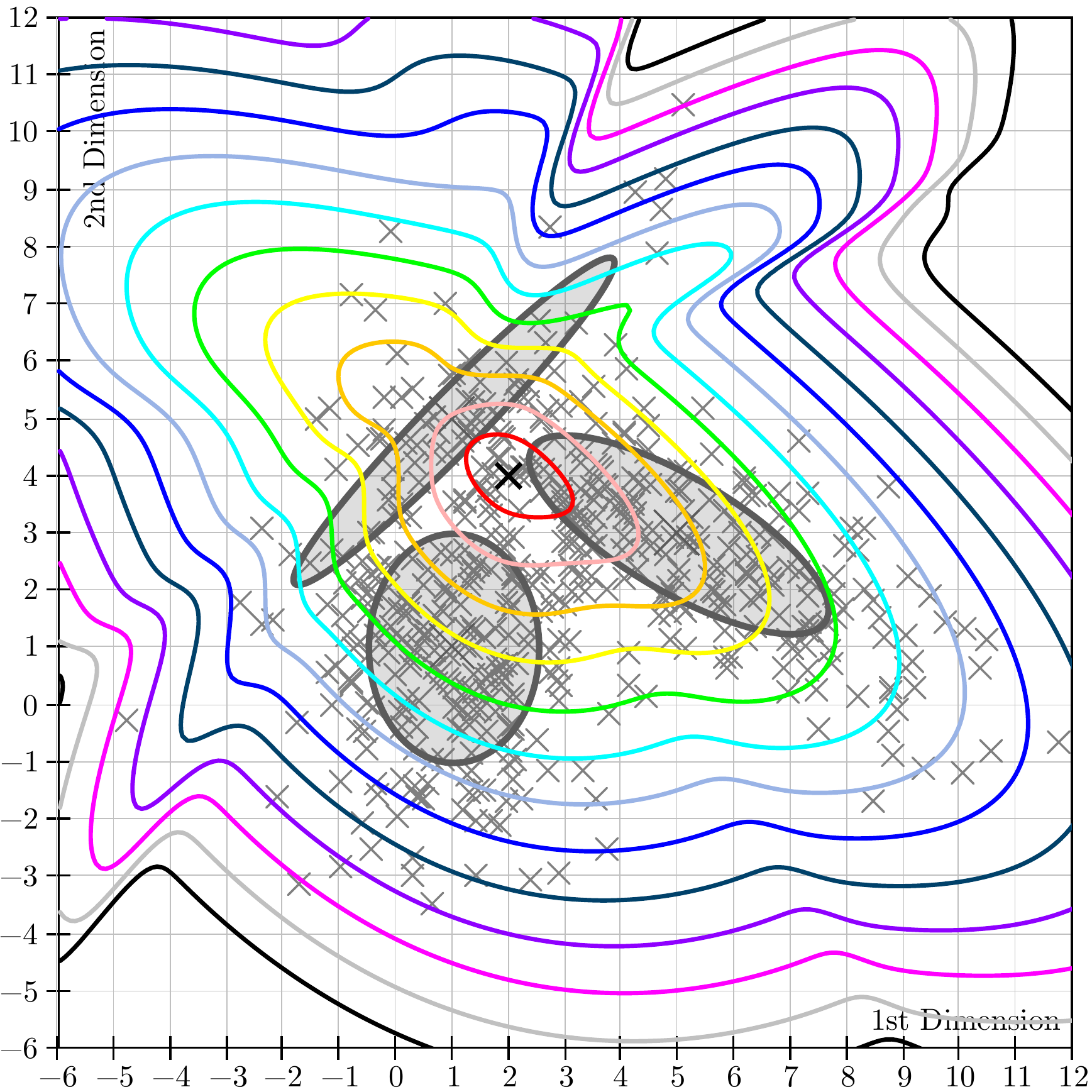}}
\caption{Influence of mixing coefficients on the RWM similarity. The level curves correspond to RWM similarity values between the sample $\mathbf{x}$ at position $(2,4)$ (thick black cross) and all samples $\mathbf{y}$ with $\Delta_{\mathbf{RWM}}(\mathbf{x},\mathbf{y}) \in \left\{ 0.5, 1.0, 1.5,\dots,6.5 \right\}$.}
\label{hmm3}
\end{figure*}

Figs.~\ref{hmm2} and \ref{hmm3} investigate the influence of different scaling factors of covariance matrices $\boldsymbol{\Sigma}_{k}$ and the influence of different values of mixing coefficients $\pi_k$, respectively. In each of the figures we have Gaussian mixtures consisting of three components with $\boldsymbol{\mu}_{1}=(1.00,5.00)^{\mathrm{T}}$, $\boldsymbol{\mu}_{2}=(5.00,3.00)^{\mathrm{T}}$, and $\boldsymbol{\mu}_{3}=(1.00,1.00)^{\mathrm{T}}$ in a two-dimensional input space. Fig.~\ref{hmm2} illustrates that the RWM similarity corresponds to the Euclidean distance if all covariance matrices are isotropic and equal (left). If scaling factors of the matrices are different, we
see how the RWM similarity is influenced by local distortions. Here, the mixing coefficients of the Gaussians are fixed to $\pi_1=\pi_2=\pi_3=0.33$. If we allow different mixing coefficients, as shown in Fig.~\ref{hmm3}, we see that the distortions are also influenced by these. Here, we use $\boldsymbol{\Sigma}_1 =\left(\begin{smallmatrix}8.13 & 7.88 \\ 7.88 & 8.13 \end{smallmatrix}\right)$, $\boldsymbol{\Sigma}_2 =\left(\begin{smallmatrix}7.00 & -3.46 \\ -3.46 & 3.00 \end{smallmatrix}\right)$ and $\boldsymbol{\Sigma}_3 =\left(\begin{smallmatrix}2.25 & 0.00 \\ 0.00 & 4.00 \end{smallmatrix}\right)$.

\begin{figure*}[phtb!]
\centering
\subfigure[$w_0$ = 4.50.]{\includegraphics[width=0.31\textwidth]{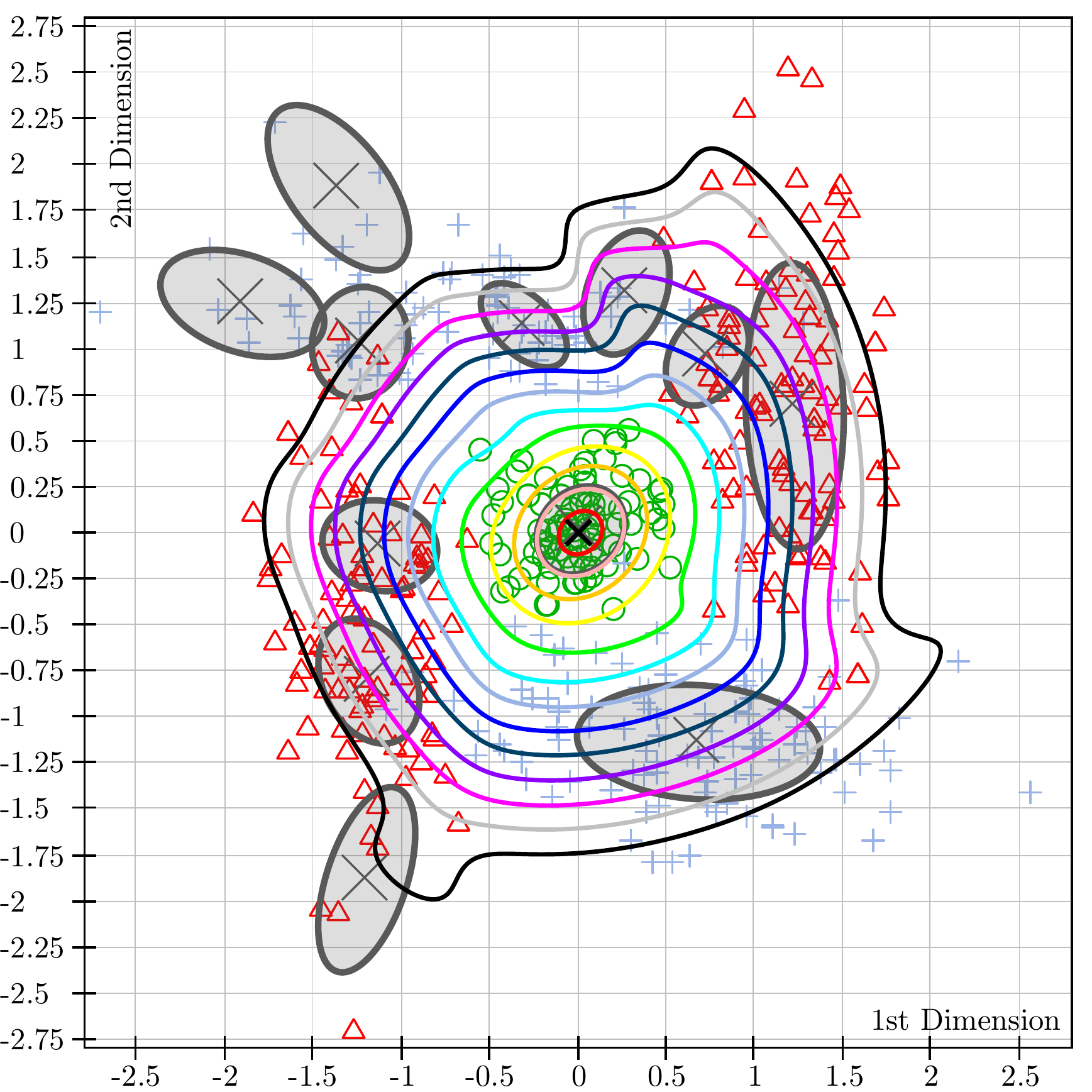}}\hfill
\subfigure[$w_0$ = 2.25.]{\includegraphics[width=0.31\textwidth]{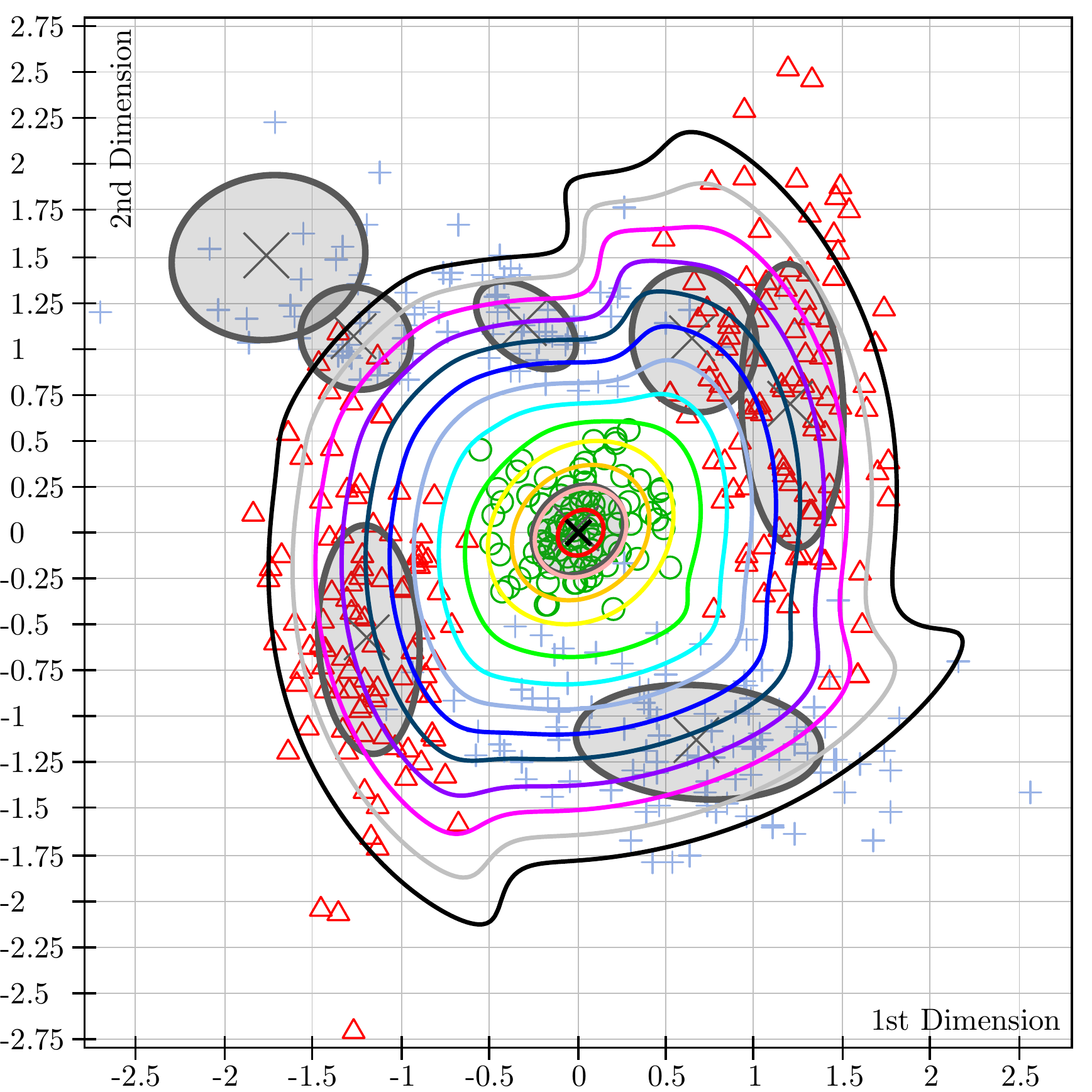}}\hfill
\subfigure[$w_0$ = 0.75.]{\includegraphics[width=0.31\textwidth]{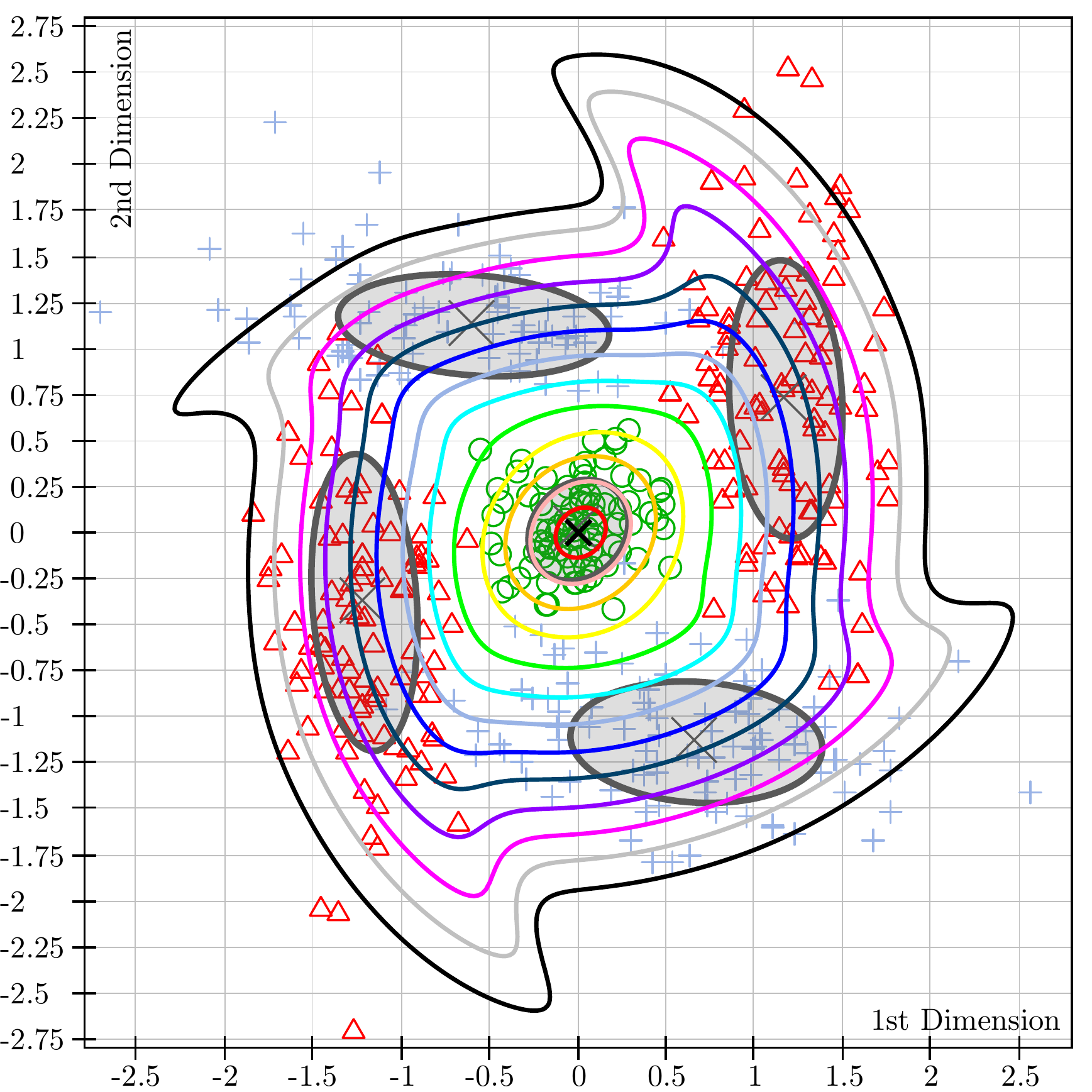}}
\caption{Robustness of the RWM similarity measure: The parts (a) -- (c) show different GMM resulting from a VI training on a synthetic data set by variation of the $w_0$ parameter. The data set is produced by five processes that are uniquely assigned to one of three classes (green circles, blue plus signs or red rectangles). The level curves correspond to RWM similarities between the sample $\mathbf{x}$ at position $(0,0)$ (thick black cross) and all samples $\mathbf{y}$ with $\Delta_{\mathbf{RWM}}(\mathbf{x},\mathbf{y}) \in \left\{ 0.5, 1.0, 1.5,\dots,6.5 \right\}$. To assess the modeling result, signs and colors reflect the class information, but this information is not used by the VI modeling technique.}
\label{robustness_oli_data}
\end{figure*}

\begin{figure*}[phtb!]
\centering
\subfigure[1st disjoint subset.]{\includegraphics[width=0.31\textwidth]{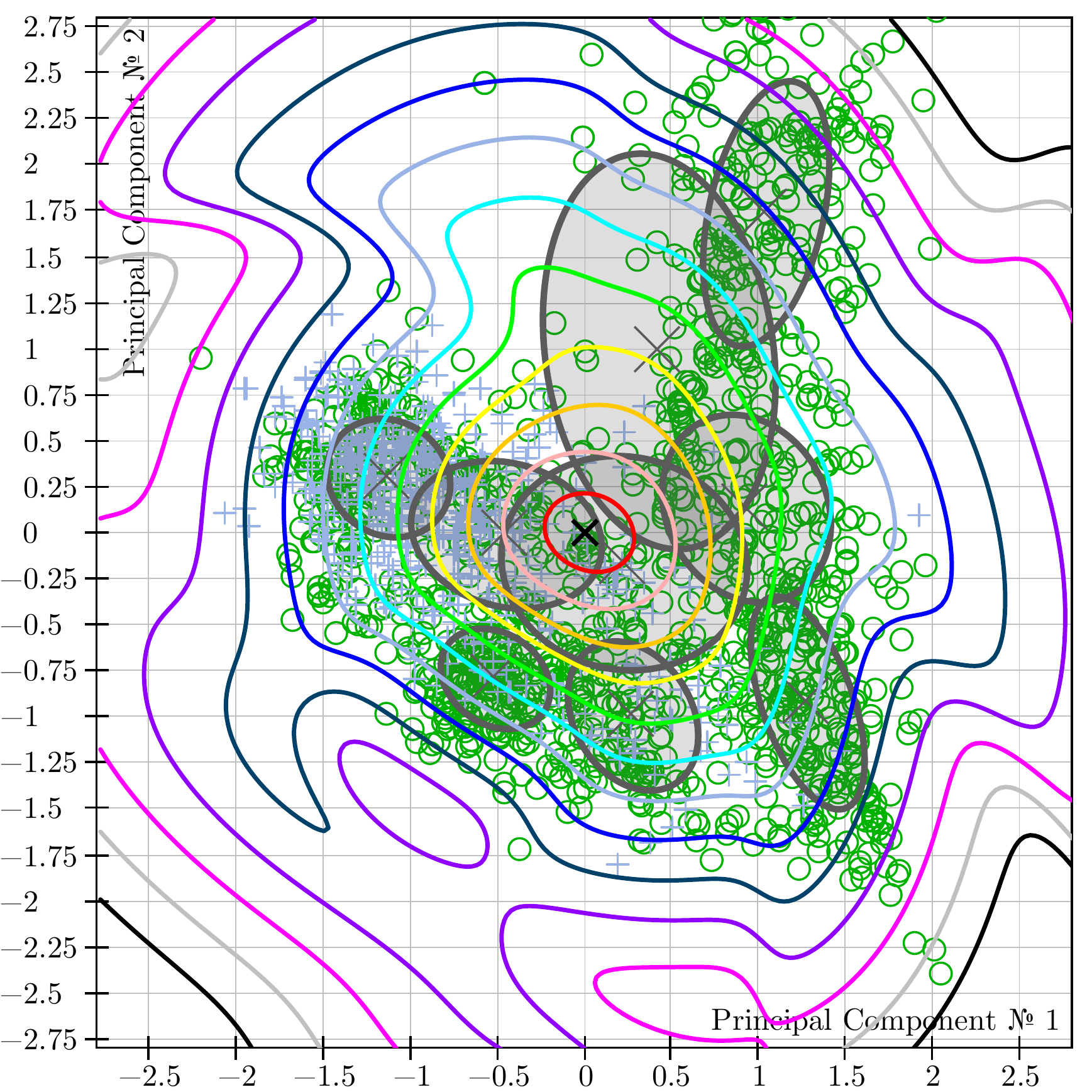}}\hfill
\subfigure[2nd disjoint subset.]{\includegraphics[width=0.31\textwidth]{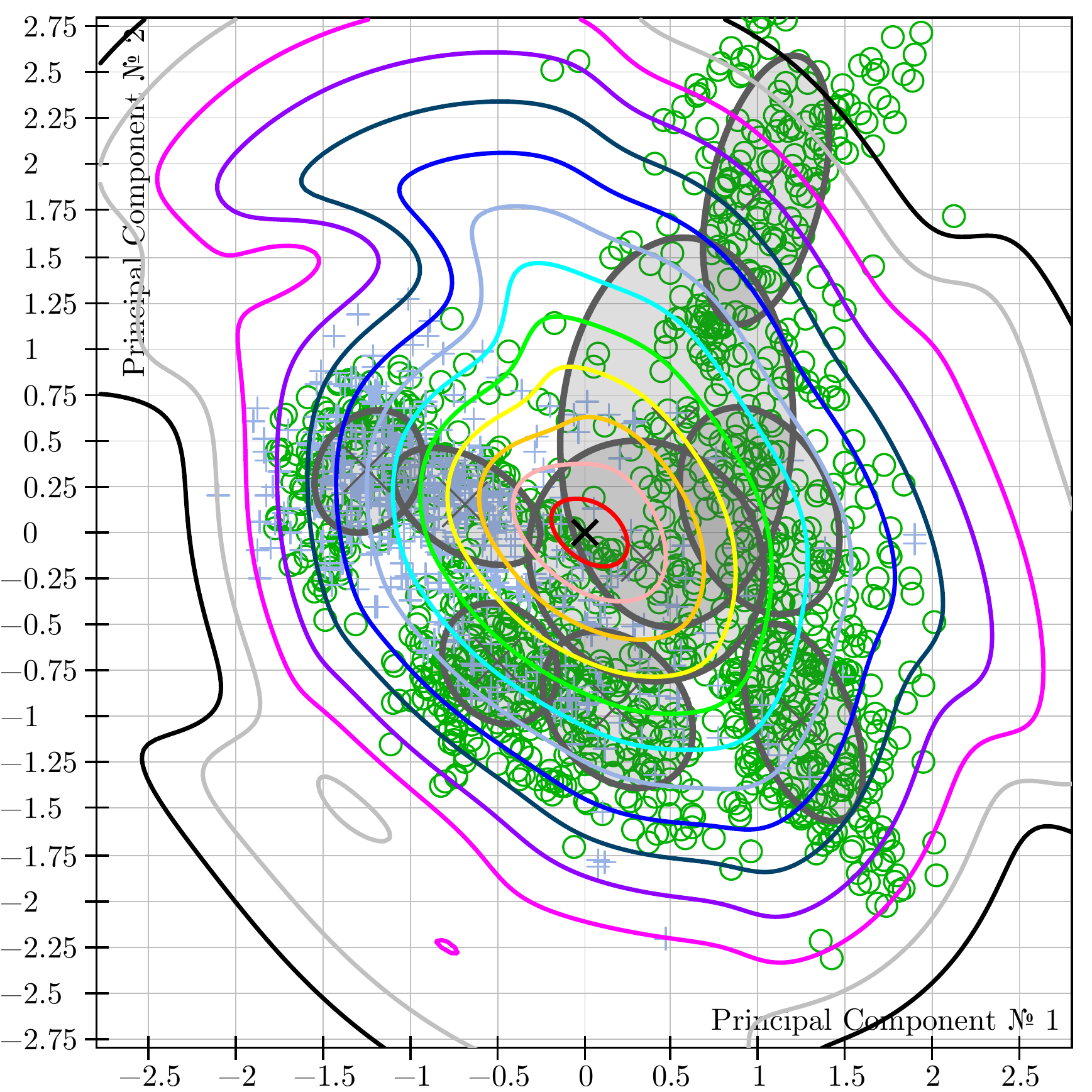}}\hfill
\subfigure[3rd disjoint subset.]{\includegraphics[width=0.31\textwidth]{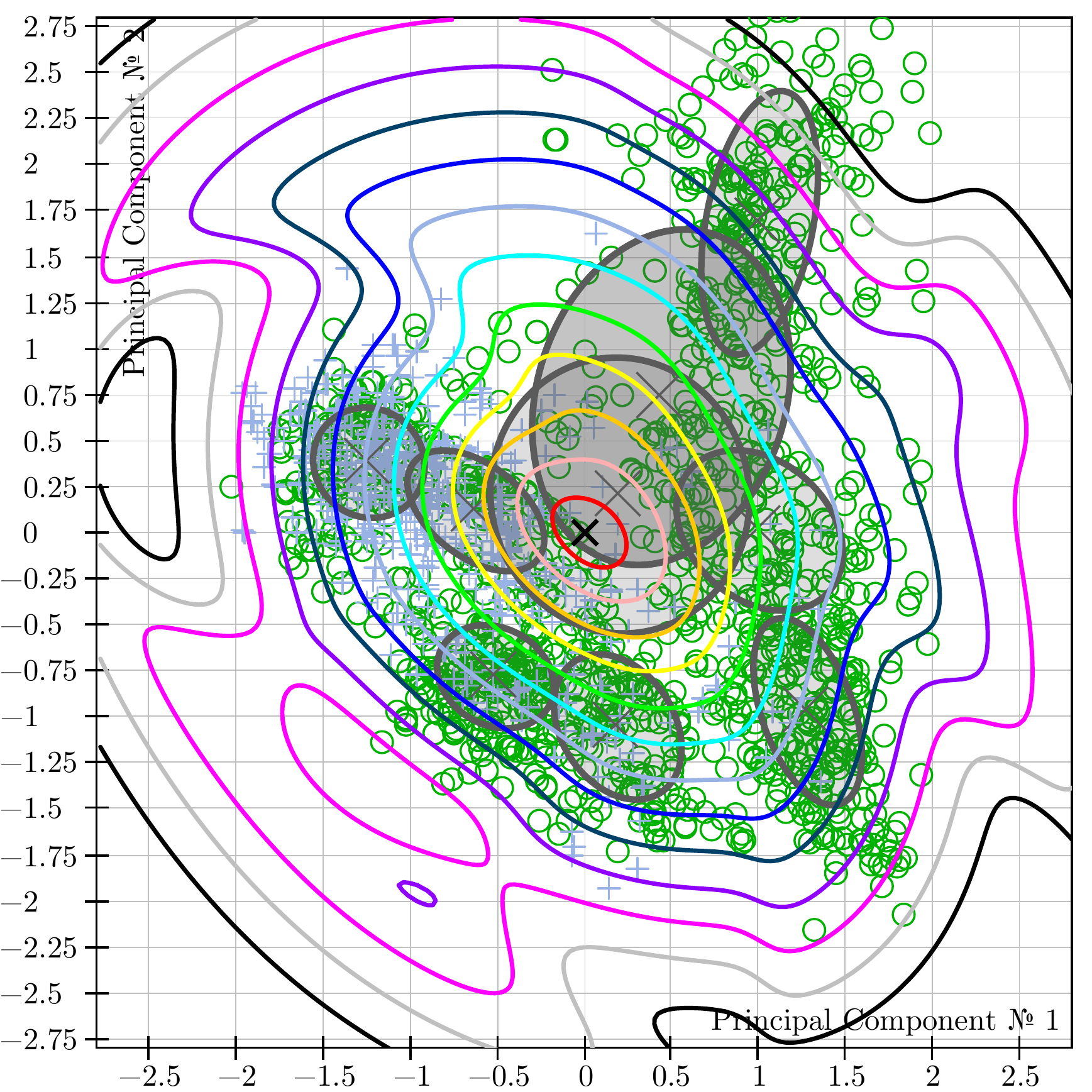}}
\caption{Robustness of the RWM similarity measure: The parts (a) -- (c) show the GMM resulting from a VI training for three disjoint subset of the Phoneme data set. The level curves correspond to RWM similarity values between the sample $\mathbf{x}$ at position $(0,0)$ (thick black cross) and all samples $\mathbf{y}$ with $\Delta_{\mathbf{RWM}}(\mathbf{x},\mathbf{y}) \in \left\{ 0.5, 1.0, 1.5,\dots,6.5 \right\}$. To assess the modeling result, signs and colors reflect the class information, but this information is not used by the VI modeling technique.}
\label{robustness_phoneme}
\end{figure*}

One might ask the question whether the RWM similarity is influenced by (pseudo-)random influences of the unsupervised modeling process such as parameters of the VI training algorithm or the choice of training samples, e.g., in a cross-validation approach. Figs.~\ref{robustness_oli_data} and \ref{robustness_phoneme} show that the RWM similarity is quite robust regarding these influences. 


In the case of continuous input dimensions, the VI algorithm has three hyper-parameters $\alpha_0$, $\beta_0$, and $w_0$. The values of these parameters are determined in an unsupervised fashion \cite{RCS14}. The hyper-parameter $\alpha_0$ controls how easy components are pruned and it has direct impact on the resulting number of components $K$. The larger the value of $\alpha_0$, the less components are pruned. For the RWM similarity the value of $\alpha_0$ is not critical because even with very small values never too many components are pruned. For very large values of $\alpha_0$ the resulting model describes the data very well, but it contains almost identical components that model together the same processes. The hyper-parameter $\beta_0$ controls how much the component centers get attracted by a prior center. The value of $\beta_0$ is also not critical for the RWM similarity. The last hyper-parameter $w_0$ controls variances (i.e., the shapes) of the components. The larger the value of $w_0$, the smaller the shapes of the resulting Gaussians, and, as a consequence, the larger the number of components are used to model the data. To analyze the influence of $w_0$ on the RWM similarity, we varied $w_0$ on a synthetic data set consisting of five processes generating data, that are uniquely assigned to one of three classes (green circles, blue plus signs or red rectangles). 
In Fig.~\ref{robustness_oli_data} it can be seen that larger values of $w_0$ result in density models with a larger number of components which cover smaller regions. However, the resulting level curves of the RWM similarity are not very different for different values of $w_0$. Nevertheless, the question is: What impact has the number of components $K$ regarding the classification results of an SVM with RWM kernel. This question will be answered in the following section.
 
In Fig.~\ref{robustness_phoneme}, we see the outcome of a VI training for three disjoint subsets of the Phoneme data set from the UCL Machine Learning Group \cite{UCL14} and the resulting RWM similarities. Training and classification are done in a two-dimensional space spanned by the two principal components of the data in order to project the data into a two-dimensional space for visualization purposes. It should also be noted that these data are not normally distributed. 

All the examples in this section only give a first impression of the behavior of the RWM similarity. In Section~\ref{sec:results} we will investigate the RWM similarity (then integrated into an SVM kernel) by means of 20 publicly available benchmark data sets. Most of these data sets are real data sets where we cannot expect that clusters in the data are normally distributed.

\subsection{The RWM Kernel}
\label{subsec:rwmKernel}


The RWM similarity measure leads to the definition of a kernel function in a straightforward way if we take the standard RBF kernel
\begin{equation}
 K_{\mathbf{RBF}}(\mathbf{x}_i,\mathbf{x}_j) = \exp\left(-\gamma (\|\mathbf{x}_{i}-\mathbf{x}_{j}\|)^2 \right) 
\end{equation}
for any two samples $\mathbf{x}_{i},\mathbf{x}_{j} \in \R^D$ with the parameter $\gamma = \frac{1}{2\sigma^2} \in \R^+$ being the kernel width and $i,j \in \N$. In this kernel we simply replace the Euclidean distance $\| \cdot \|$ with the RWM similarity:
\begin{equation} 
K_{\mathbf{RWM}}(\mathbf{x}_i,\mathbf{x}_j) = \exp\left(-\gamma \left(\Delta_{\mathbf{RWM}}(\mathbf{x}_{i},\mathbf{x}_{j})\right)^2\right). 
\end{equation}
This \textit{responsibility weighted Mahalanobis (RWM) kernel} considers structure in the data captured by a mixture density model as described in the previous section. This is also the case with a kernel based on the GMM based measure, 
\begin{equation}
K_{\mathbf{GMM}}(\mathbf{x}_i,\mathbf{x}_j) = \exp\left(-\gamma \left(\Delta_{\mathbf{GMM}}(\mathbf{x}_{i},\mathbf{x}_{j})\right)^2 \right), 
\end{equation}
to which we will compare our approach.

A key advantage of the RWM kernel is that it can be used in combination with any standard implementation of SVM such as \textsf{libsvm} \cite{libsvm} as it is only necessary to construct the kernel matrix needed as input for an optimization procedure such as SMO. 
One might ask the question whether the use of the RWM kernel always leads to a \textit{positive semi-definite (PSD) kernel matrix} or not. Though this question is of some theoretical interest, we postpone it here, as we exploit a specific property of the SMO version realized in the \textsf{libsvm} library: This SMO version (for details see \cite{CFL06,FCL05}) is able to cope with non-PSD kernels.

\begin{figure}[htbp!]
\centering
\includegraphics[width=0.45\textwidth]{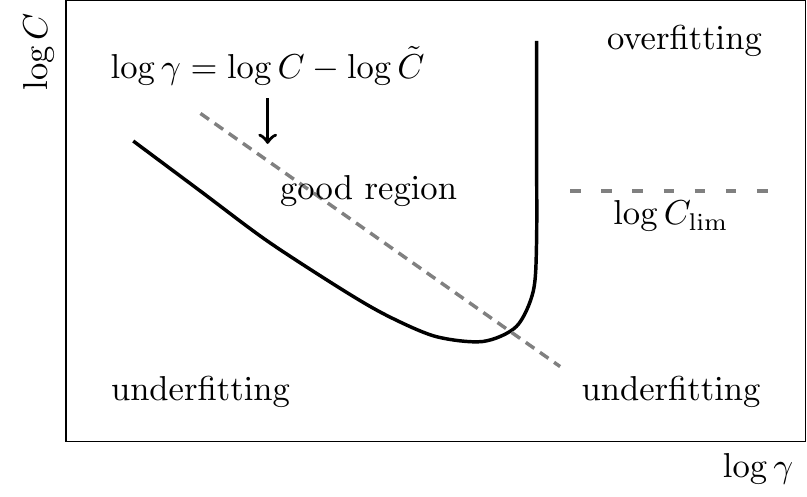}
\vspace{-0.5em}
\caption{The two-dimensional $(\log \gamma,\log C)$ parameter space typically contains two regions, an overfitting / underfitting region and a region with good parameter combinations.}
\label{fig:keerthiLin}
\end{figure}

The parameters of a C-SVM with RBF kernel -- the penalty parameter $C$ and the kernel width $\gamma$ -- can be found by applying a heuristic search method presented in \cite{KL03} instead of doing an exhaustive search. Keerthi and Lin show in \cite{KL03} that the two-dimensional $(\log \gamma,\log C)$ parameter space typically contains two regions, an overfitting / underfitting region and a region with good parameter combinations (cf.\ Fig.~\ref{fig:keerthiLin}), that have similar shapes for all data sets (a property that cannot be observed, e.g., for polynomial or sigmoid kernels).

\begin{figure*}[htbp!]
\centering
\subfigure[Data set Two Moons.]{\includegraphics[width=0.31\textwidth]{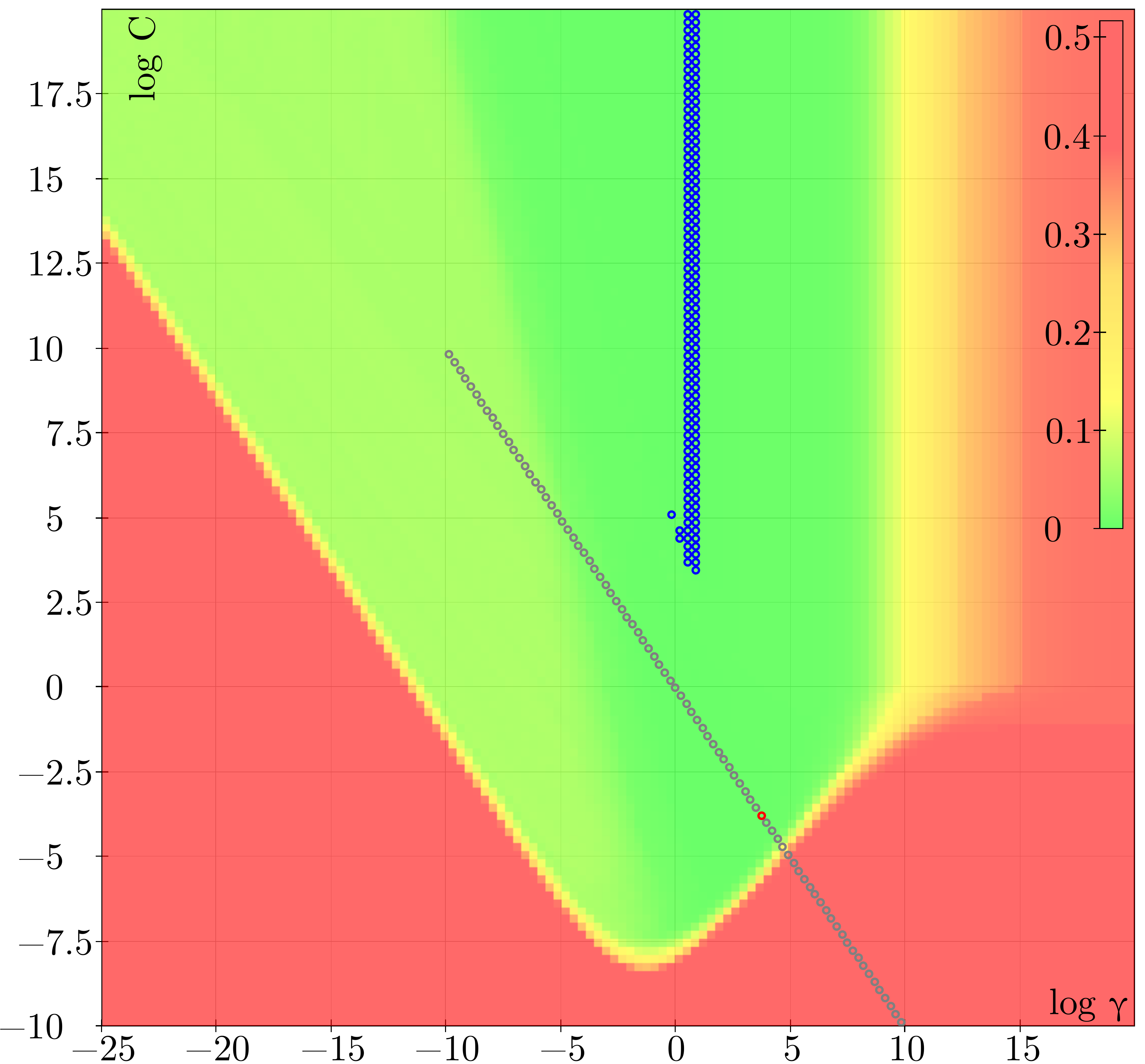}}\hfill
\subfigure[Data set Wine.]{\includegraphics[width=0.31\textwidth]{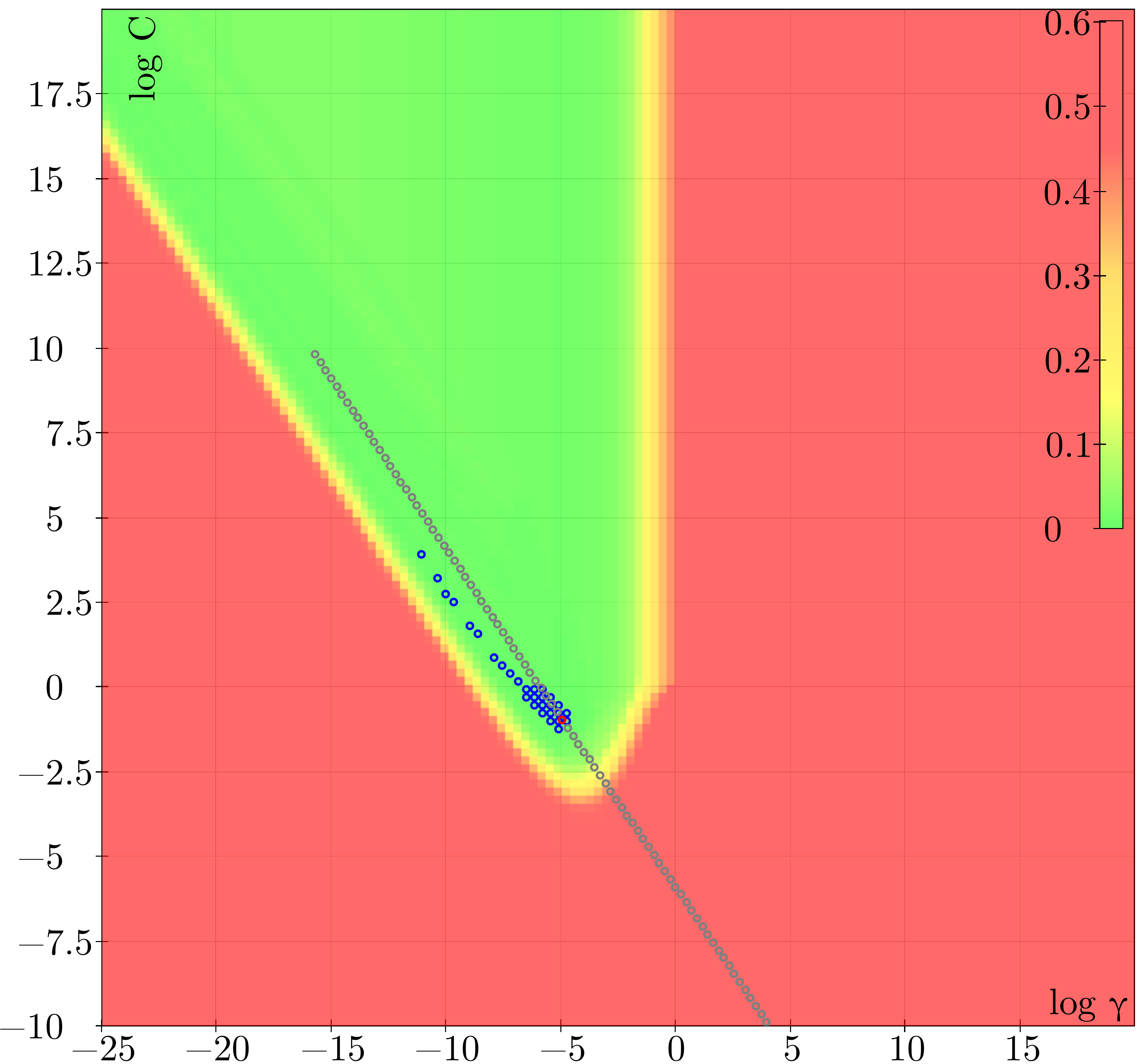}}\hfill
\subfigure[Data set Iris.]{\includegraphics[width=0.31\textwidth]{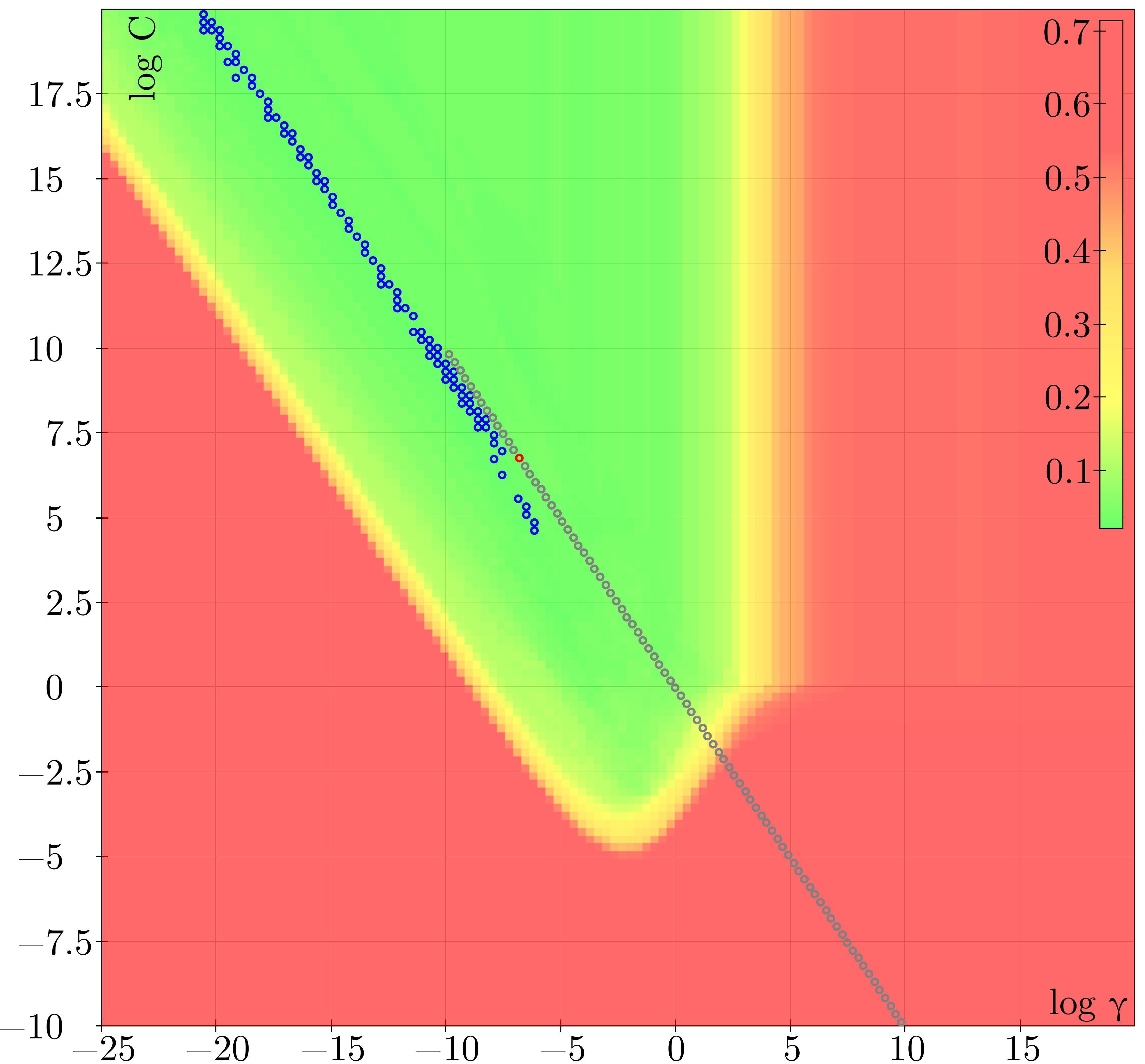}}
\caption{Grid search for RBF kernel in the parameter space spanned by $\log \gamma$ (horizontal) and $\log C$ (vertical). The blue circles represent the parameter combinations with the smallest classification error (averaged over a $5$-fold cross-validation) on the validation set found with exhaustive searching. The gray circles correspond to the parameter combinations which were analyzed by means of the search heuristic of Keerthi and Lin. The combination with the smallest error is colored red.}
\label{gridsearch_rbf}
\end{figure*}

\begin{figure*}[htbp!]
\centering
\subfigure[Data set Two Moons.]{\includegraphics[width=0.31\textwidth]{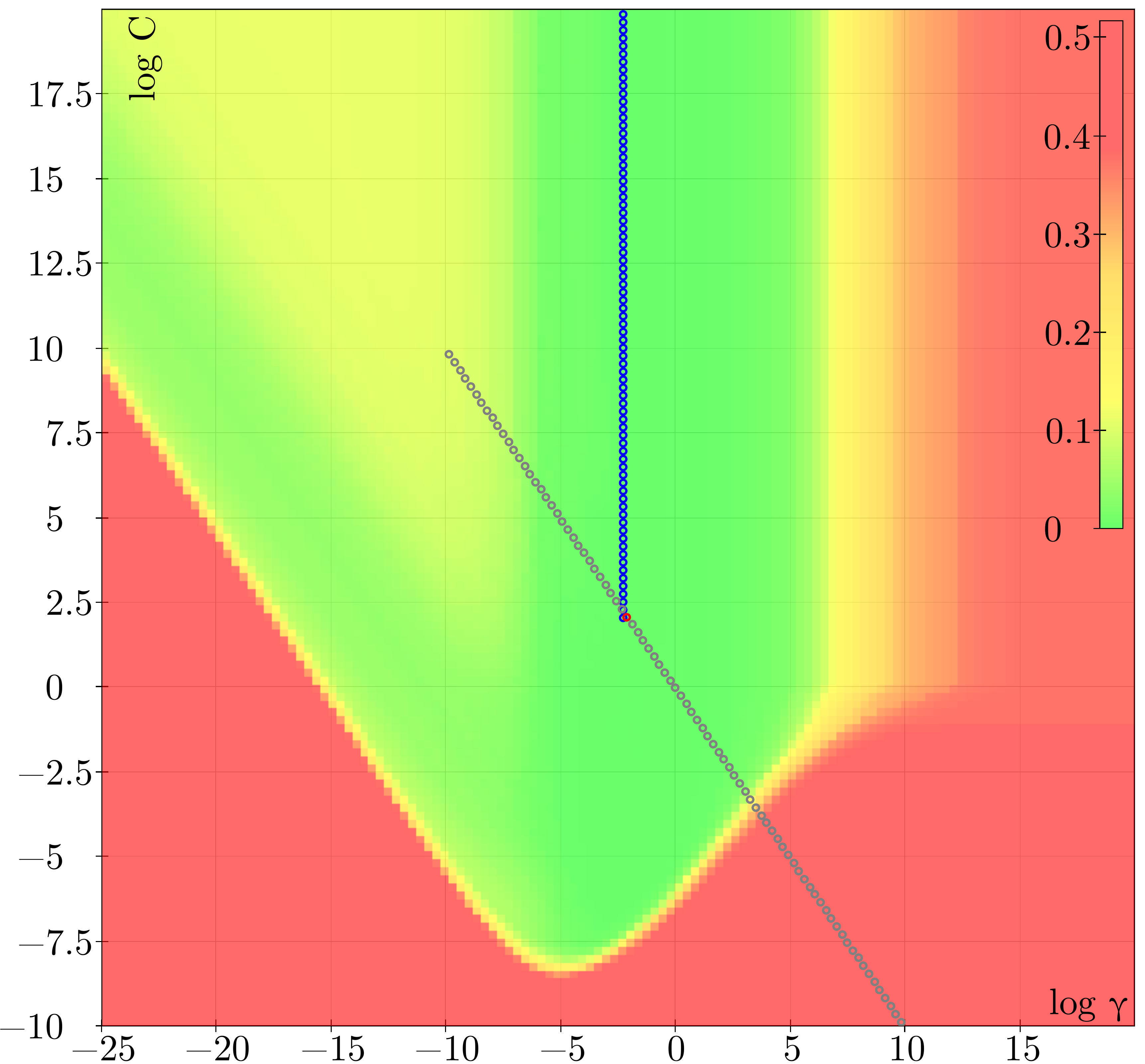}}\hfill
\subfigure[Data set Wine.]{\includegraphics[width=0.31\textwidth]{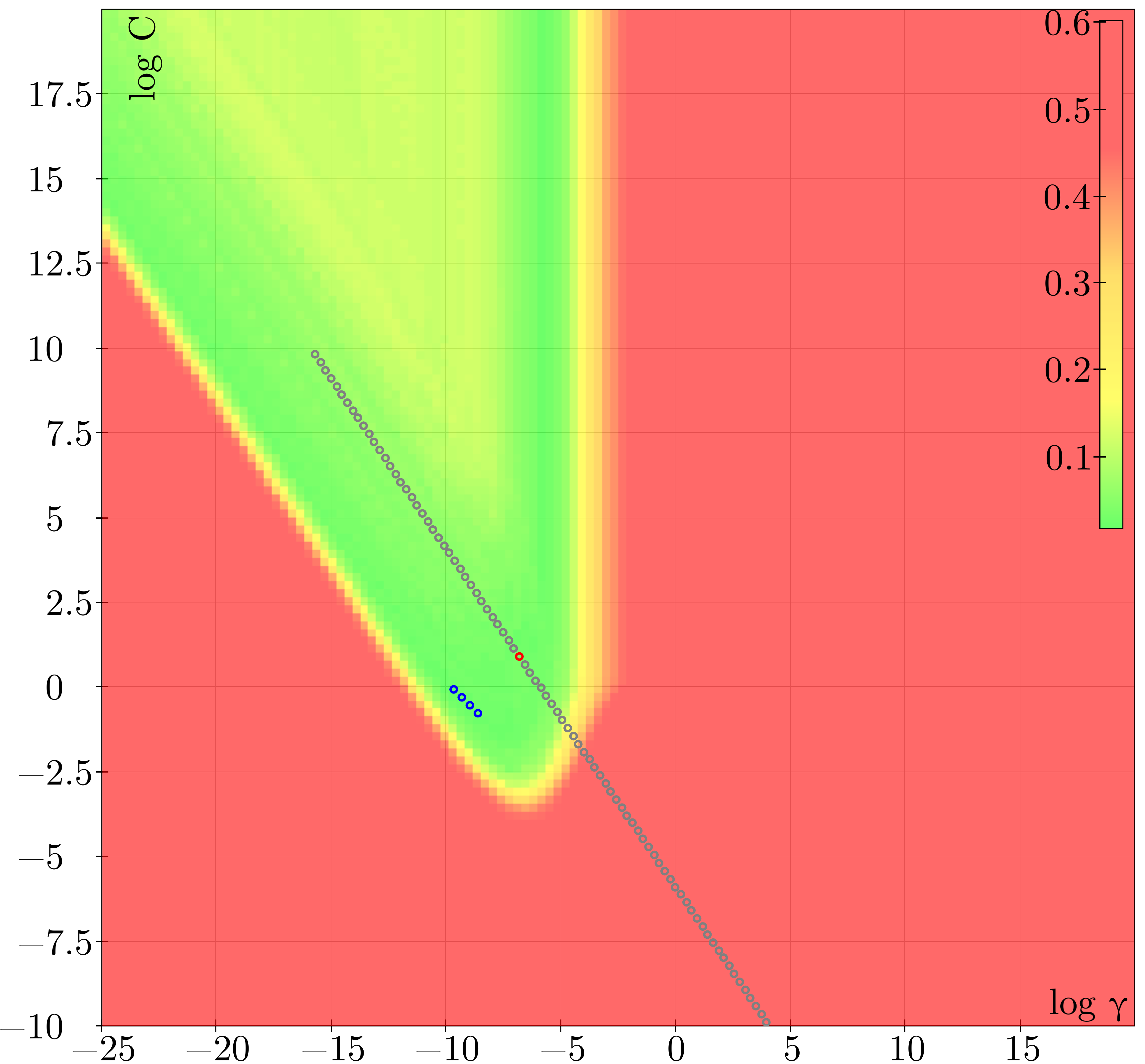}}\hfill
\subfigure[Data set Iris.]{\includegraphics[width=0.31\textwidth]{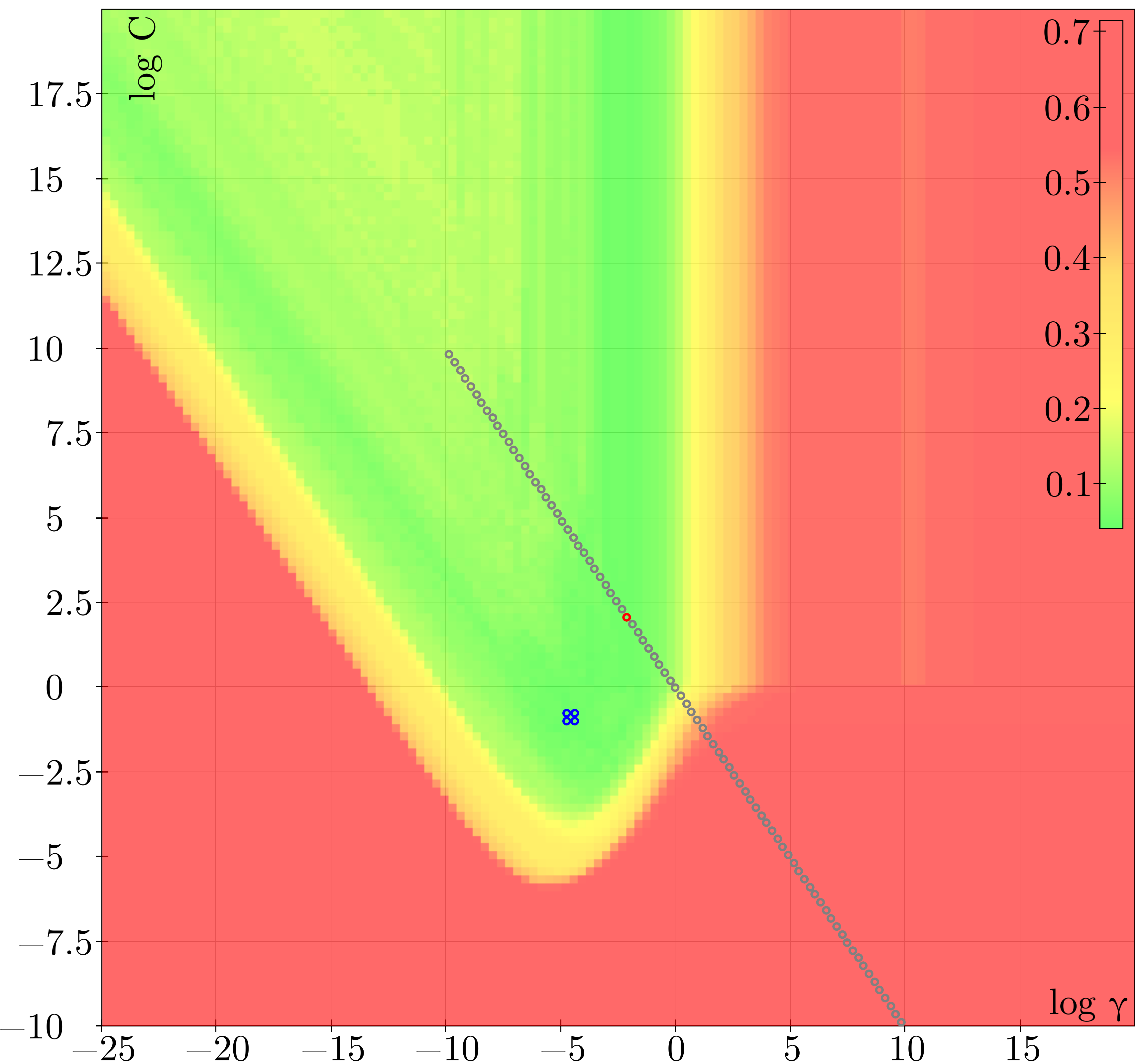}}
\caption{Grid search for RWM kernel in the parameter space spanned by $\log \gamma$ (horizontal) and $\log C$ (vertical). The blue circles represent the parameter combinations with the smallest classification error (averaged over a $5$-fold cross-validation) on the validation set found with exhaustive searching. The gray circles correspond to the parameter combinations which were analyzed by means of the search heuristic of Keerthi and Lin. The combination with the smallest error is colored red.}
\label{gridsearch_rwm}
\end{figure*}

This leads to an efficient heuristic to find parameter combinations with small generalization error: First, the optimal penalty parameter $\tilde{C}$ for an SVM with linear kernel is determined, because  $\tilde{C}$ defines a straight line $\log\gamma = \log C  -\log \tilde{C}$ with a slope of minus one which cuts the region with good parameter combinations (cf.~Fig.~\ref{gridsearch_rbf}, the line defined by the gray circles that cuts the green colored region). Second, the best combination of $C$ and $\gamma$ along this line is determined using an SVM with RBF kernel (cf.~Fig.~\ref{gridsearch_rbf}, the red circle corresponds to the best combination on the line of otherwise gray circles). 

Figs.~\ref{gridsearch_rbf} and \ref{gridsearch_rwm} show the results of an exhaustive search in the parameter space spanned by $\log \gamma $ and $\log C$ for three data sets from the UCI Machine Learning Repository \cite{AN07}. The blue circles correspond to the parameter combinations with the smallest generalization error found by an exhaustive search and the red circles represent the best parameter combinations found by the explained heuristic.
While Fig.~\ref{gridsearch_rbf} shows the results for an RBF kernel, Fig.~\ref{gridsearch_rwm} demonstrates that the heuristic works with the RWM kernel, too. Thus, a C-SVM with
RWM kernel can parametrized just as easily as a C-SVM with RBF kernel. 

As the Gaussian mixtures that define the RWM similarity can be trained in an unsupervised way, the RWM kernel is able to consider structure in data even if the data are only partially labeled. It has already been shown in Fig.~\ref{intro_example} that the RWM kernel is well-suited for SSL. The same holds, however, for other kernels such as the GMM kernel or the LapSVM (see Section~\ref{sec:overview}) to which we will compare the RWM kernel in Section~\ref{sec:results}.

\begin{figure*}[htbp!]
\centering
\subfigure[12 components.]{\includegraphics[width=0.31\textwidth]{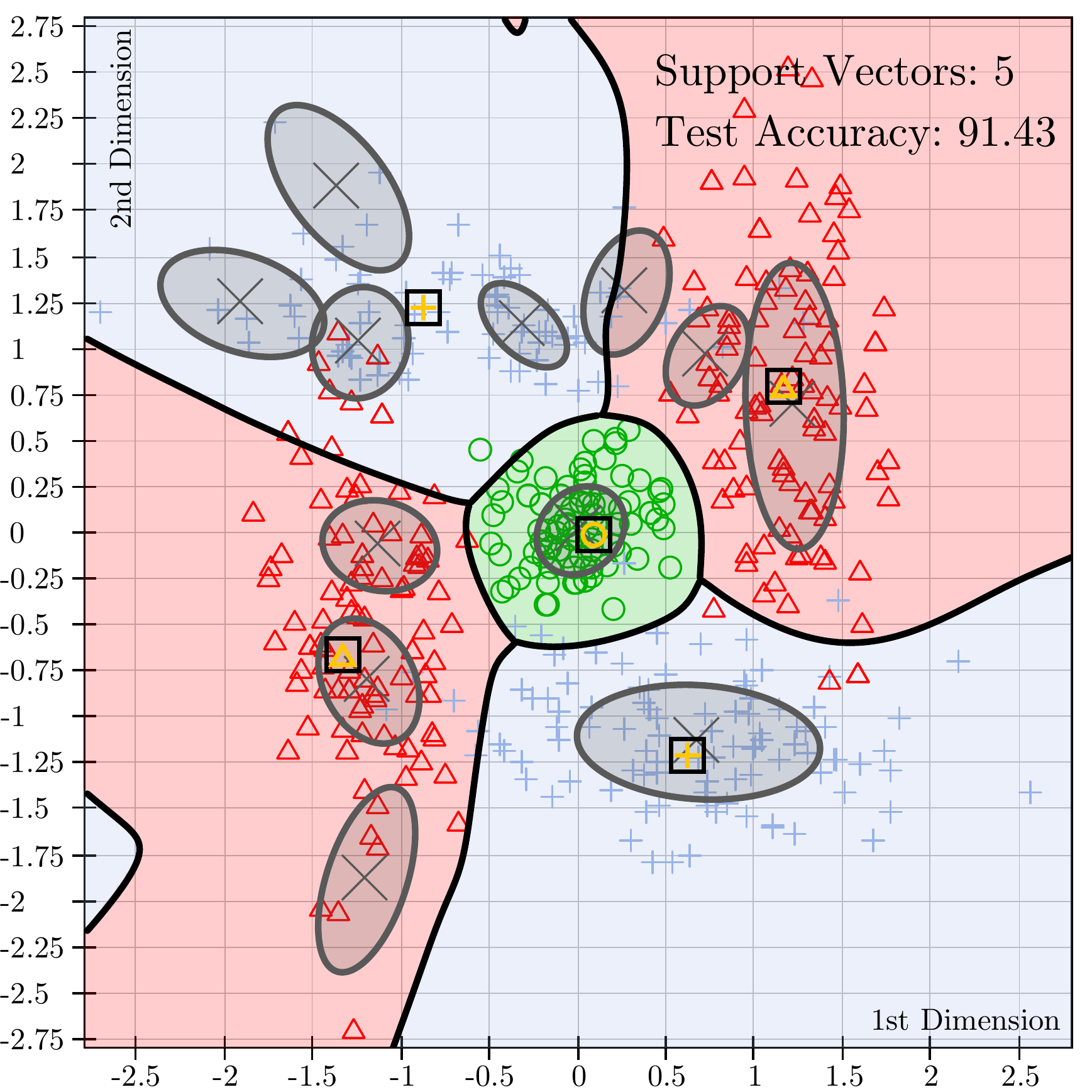}}\hfill
\subfigure[8 components.]{\includegraphics[width=0.31\textwidth]{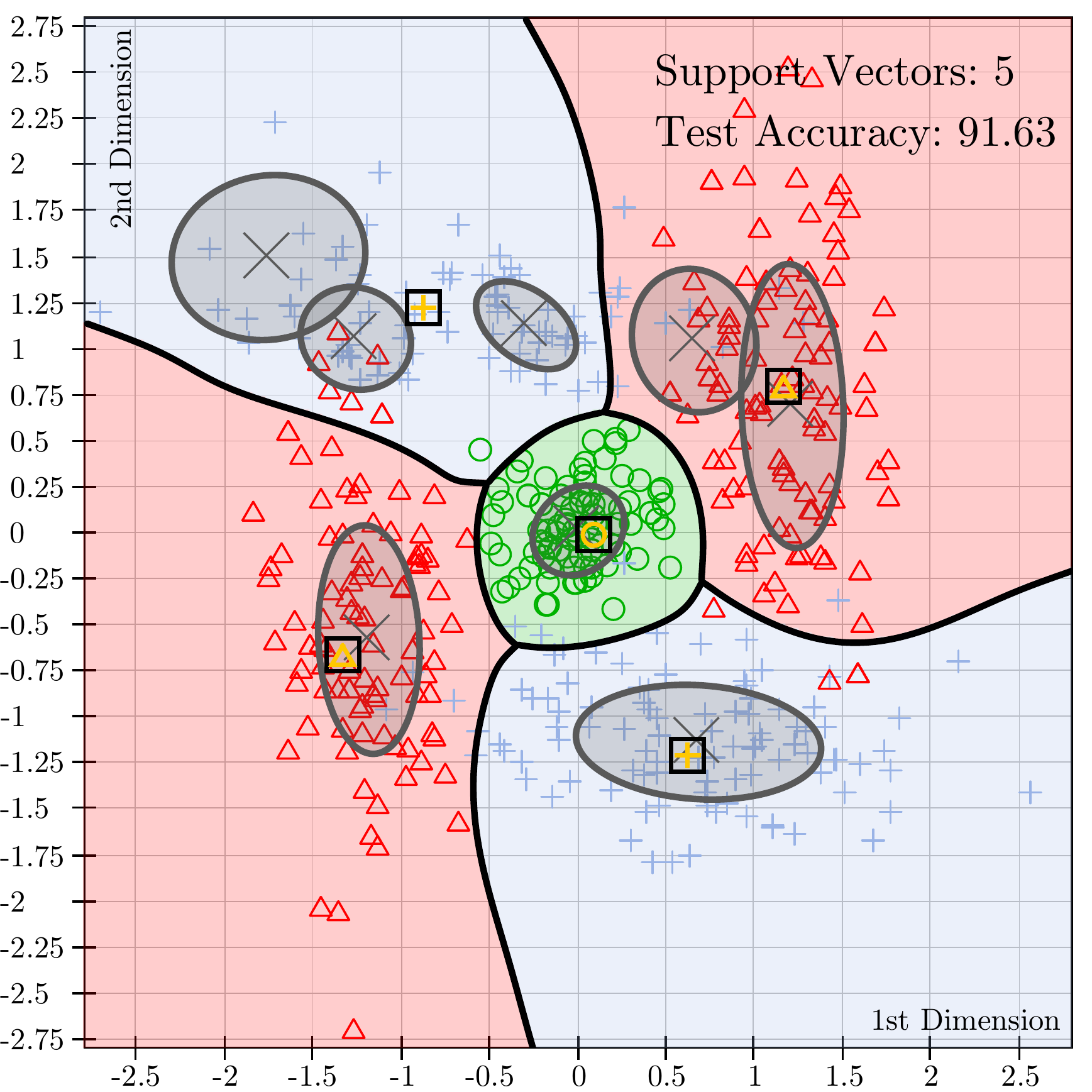}}\hfill
\subfigure[5 components.]{\includegraphics[width=0.31\textwidth]{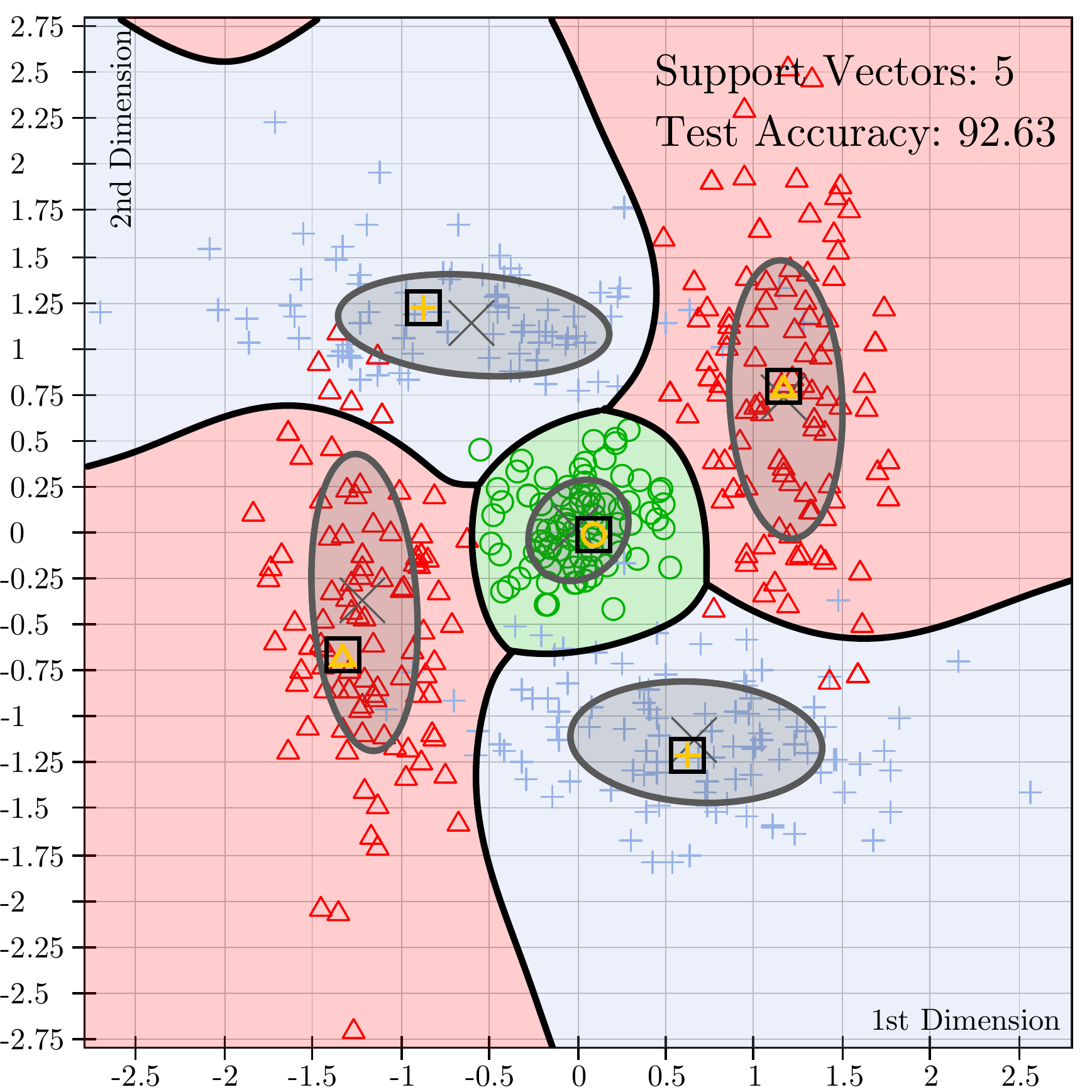}}
\caption{Influence of a large number of model components on the classification results of an SVM with RWM kernel. The training samples framed by a black square are support vectors.}
\label{oli_1}
\end{figure*}

Finally, the following question shall be answered with an experiment: How does an SVM with RWM kernel behave if the underlying model has a number of components that is different from the number of data generating processes? For this purpose, we use the data set shown in Fig.~\ref{robustness_oli_data}, for which we varied the parameter $w_0$ of the VI algorithm (see above). Now, we use the respective density models to train SVM with RWM kernel. Here, only five samples (orange colored), one from each process, are labeled and used to solve the classification problem. Fig.~\ref{oli_1} shows the resulting SVM classifiers. It can be seen that the classification accuracies on the test data decrease slightly with an increasing  number of components.

\begin{figure*}[htbp!]
\centering
\subfigure[4 components.]{\includegraphics[width=0.31\textwidth]{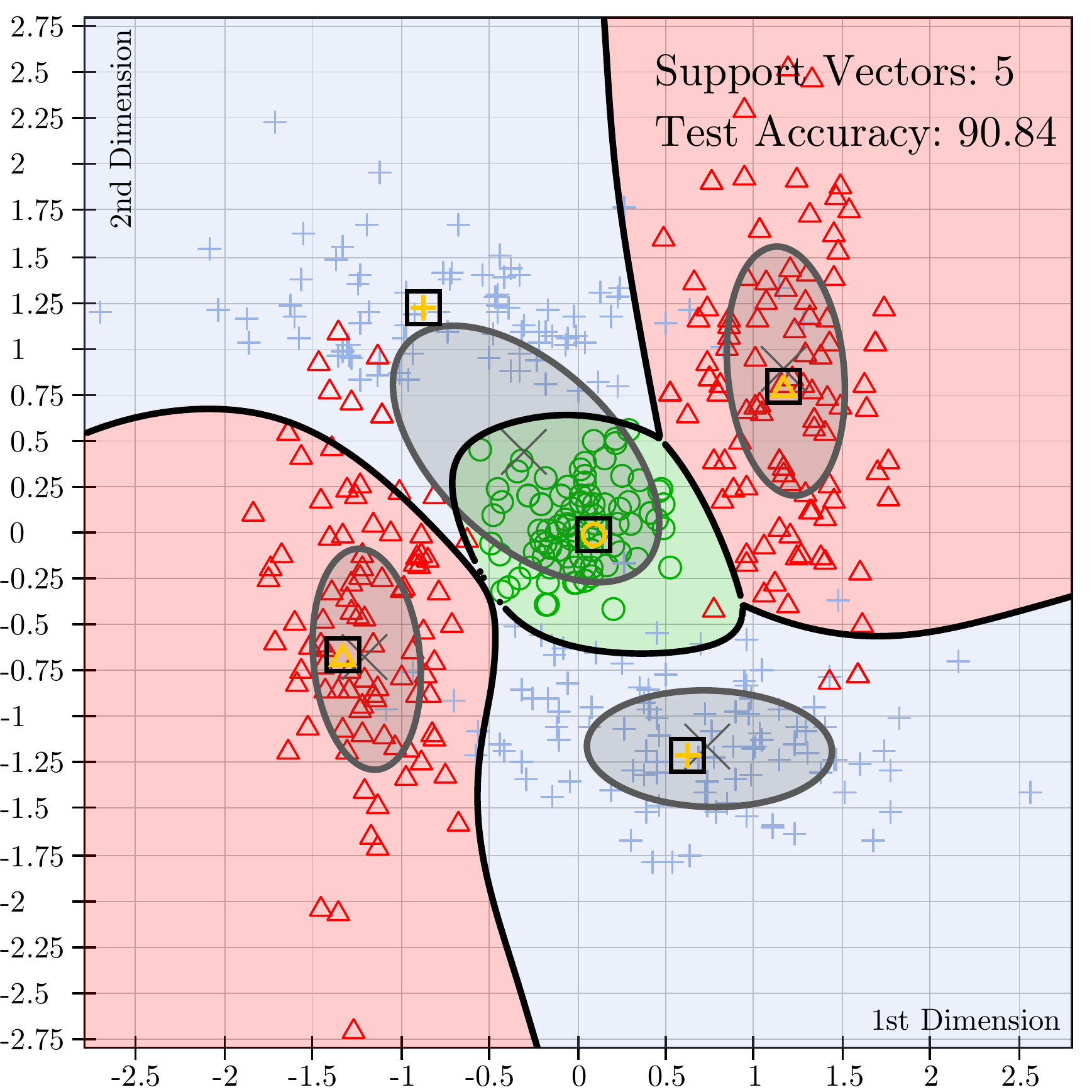}}\hfill
\subfigure[3 components.]{\includegraphics[width=0.31\textwidth]{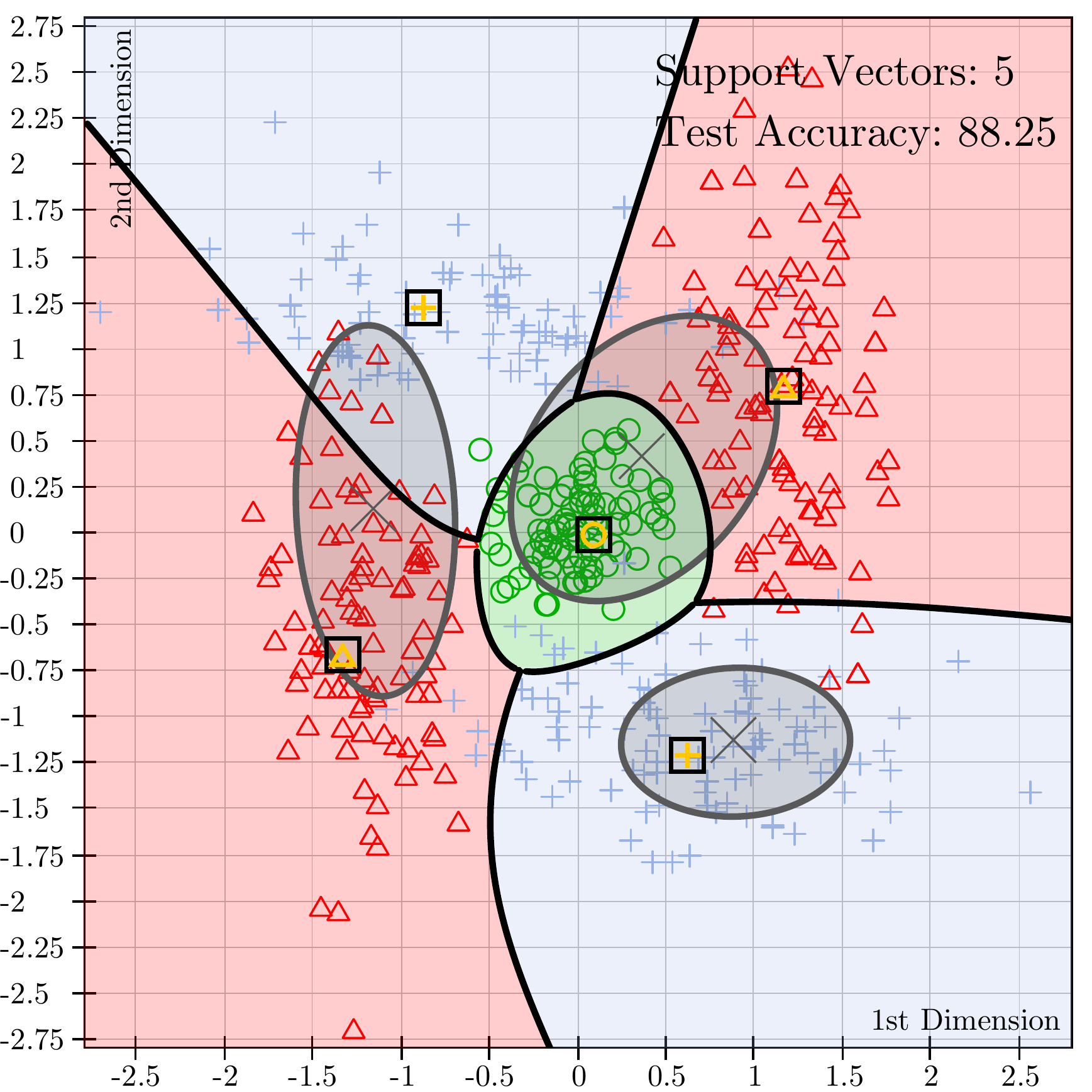}}\hfill
\subfigure[1 component.]{\label{fig:oli_2C}\includegraphics[width=0.31\textwidth]{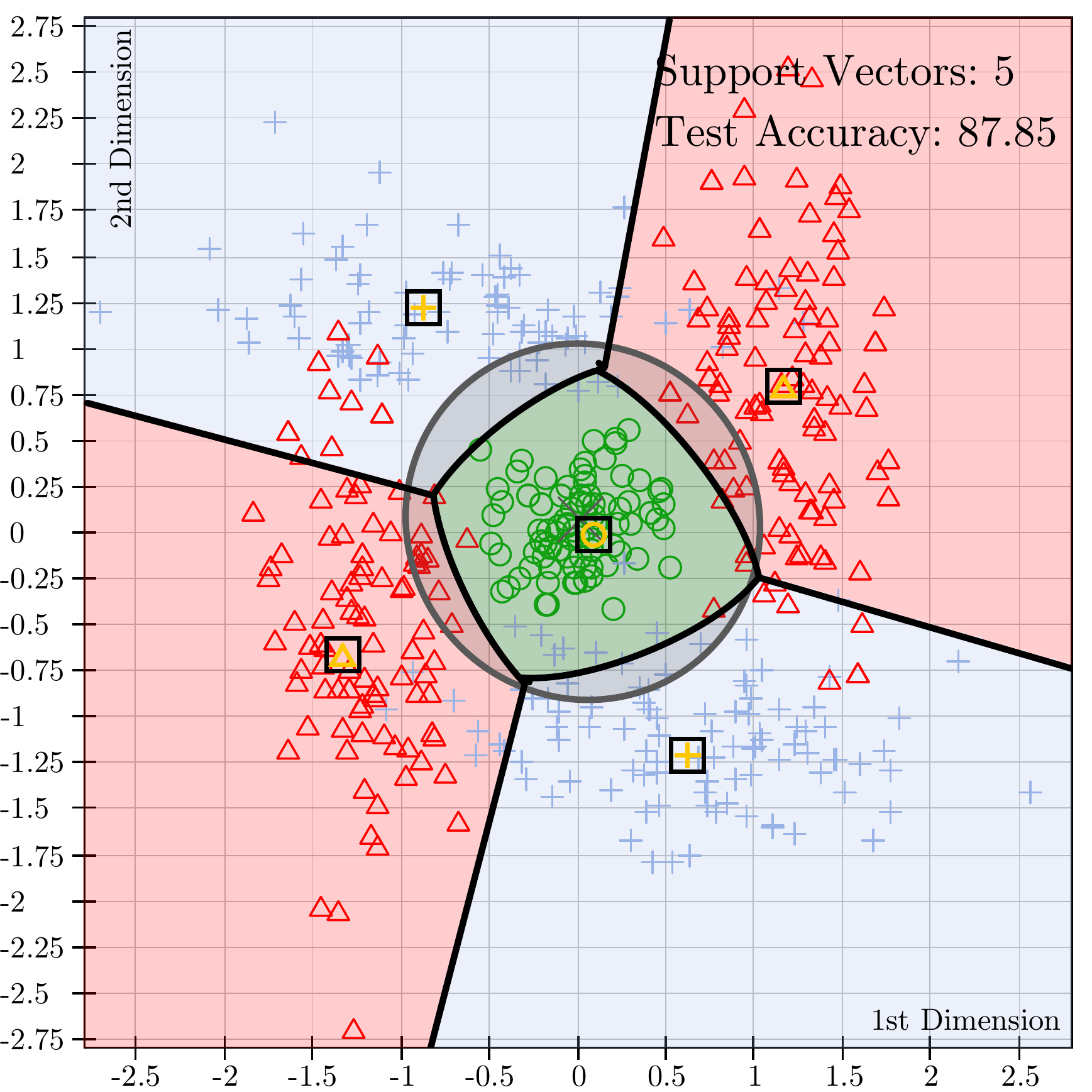}}
\caption{Influence of a small number of model components on the classification results of an SVM with RWM kernel. The training samples framed by a black square are support vectors.}
\label{oli_2}
\end{figure*}

Usually, it is not possible to parameterize the VI algorithm in the way such that the resulting density model contains a number of components that is to small. However, we limit the number of components such that the VI can use only four, three, or one to model the data. The resulting models and trained SVM with RWM kernels are depicted in Fig.~\ref{oli_2}. In case of one component (see Fig.~\ref{oli_2}\subref{fig:oli_2C}) the RWM kernel cannot extract much information from model and, therefore, it behaves like the RBF kernel. Consequently, the SVM with that RWM kernel yields roughly the same test accuracy as an SVM with RBF kernel, namely $87.67\%$. With an increasing number $K$, the classification results of the corresponding SVM with RWM kernel also increase. In summary, we can state for this experiment that an SVM with RWM kernel, which is parameterized appropriately, yields at least the classification results of an SVM with RBF kernel.

\subsection{Extension of the RWM Kernel for Categorical Input Dimensions}
\label{sec:extensionRWM}

In real classification problems we usually not only have continuous (real-valued) input dimensions. While integer dimensions are often handled such as continuous ones, this is typically not possible for categorical (non-ordinal) inputs. Assume we have a set $X$ of samples where each sample has $D$ continuous and $E$ categorical input dimensions. Each of the $E$ categorical input dimensions has $K_{e}$ different categories for which we use a one-out-of-$K_{e}$ coding scheme. Then, we extend the RWM kernel as follows:
\ifpreprint
\begin{equation}
K_{\mathbf{RWM}}(\mathbf{x}_i,\mathbf{x}_j) = 
\exp\left(-\gamma \left(\alpha\left(\Delta_{\mathbf{RWM}}(\mathbf{x}_{i}',\mathbf{x}_{j}')\right)^2 + \beta\left(\Delta_{\mathbf{0/1}}(\mathbf{x}_{i}'',\mathbf{x}_{j}'')\right)^2\right) \right) \nonumber
\end{equation}
\else
\begin{equation}
K_{\mathbf{RWM}}(\mathbf{x}_i,\mathbf{x}_j) = 
\exp\left(-\gamma \left(\alpha\left(\Delta_{\mathbf{RWM}}(\mathbf{x}_{i}',\mathbf{x}_{j}')\right)^2\!+ \beta\left(\Delta_{\mathbf{0/1}}(\mathbf{x}_{i}'',\mathbf{x}_{j}'')\right)^2\right) \right) \nonumber
\end{equation}
\fi
with weighting factors $\alpha, \beta\in [0,1]$, $\mathbf{x}_i,\mathbf{x}_j \in X$ and $\mathbf{x}_i',\mathbf{x}_j' \in \R^{D}$, $\mathbf{x}_i'',\mathbf{x}_j'' \in \mathbb{B}^{E'}$ (with $E'=\sum_{e=1}^{E} K_{e}$) only containing the values of the respective continuous and (binary encoded) categorical dimensions. For the categorical dimensions, we define
\ifpreprint
\begin{equation}
 \Delta_{\mathbf{0/1}}(\mathbf{x}_{i}'',\mathbf{x}_{j}'') = \sum_{e = 1}^{E} (1 - \delta_{e,ij}), 
\text{ with }
\delta_{e,ij} = 
\begin{cases}
1 & \text{ for } \left(x_i''\right)_e = \left(x_j''\right)_e\\
0 & \text{ otherwise } \end{cases},
\end{equation}
\else
\begin{equation}
 \Delta_{\mathbf{0/1}}(\mathbf{x}_{i}'',\mathbf{x}_{j}'') = \sum_{e = 1}^{E} (1 - \delta_{e,ij}),
 \end{equation} 
with 
\begin{equation}
\delta_{e,ij} = 
\begin{cases}
1 & \text{ for } \left(x_i''\right)_e = \left(x_j''\right)_e\\
0 & \text{ otherwise } \end{cases},
\end{equation}
\fi
i.e., simply by checking the values in the different dimensions for equality. If necessary, it is also possible to weight the categorical part and the continuous part differently by means of the parameters $\alpha, \beta \in [0,1]$. If $\alpha$ and $\beta$ are both set to $1$ and the covariance matrix of each model component corresponds to the identity matrix, then the RWM kernel behaves like an RBF kernel with binary encoded, categorical dimensions.

\section{Simulation Experiments}
\label{sec:results}
This section compares the new RWM kernel to some other kernels -- RBF, GMM, and LAP kernels -- in particular for SSL. For the SVM with LAP kernel (also called LapSVM), we ported the MATLAB implementation of Melacci \cite{Melacci12} to Java and adapt\-ed it to cope with multi-class problems. First, we visualize the behavior of the kernels for five data sets with two-dimensional input spaces. 
Second, simulation experiments are performed on 20 benchmark data sets to compare the mentioned kernels numerically and in some more detail. Third, we briefly summarize the ``lessons learned'' from our experimental studies.

\begin{figure*}[htbp!]
\centering
\vspace{-0.3cm}
\subfigure[RWM kernel.]{\includegraphics[width=0.24\textwidth]{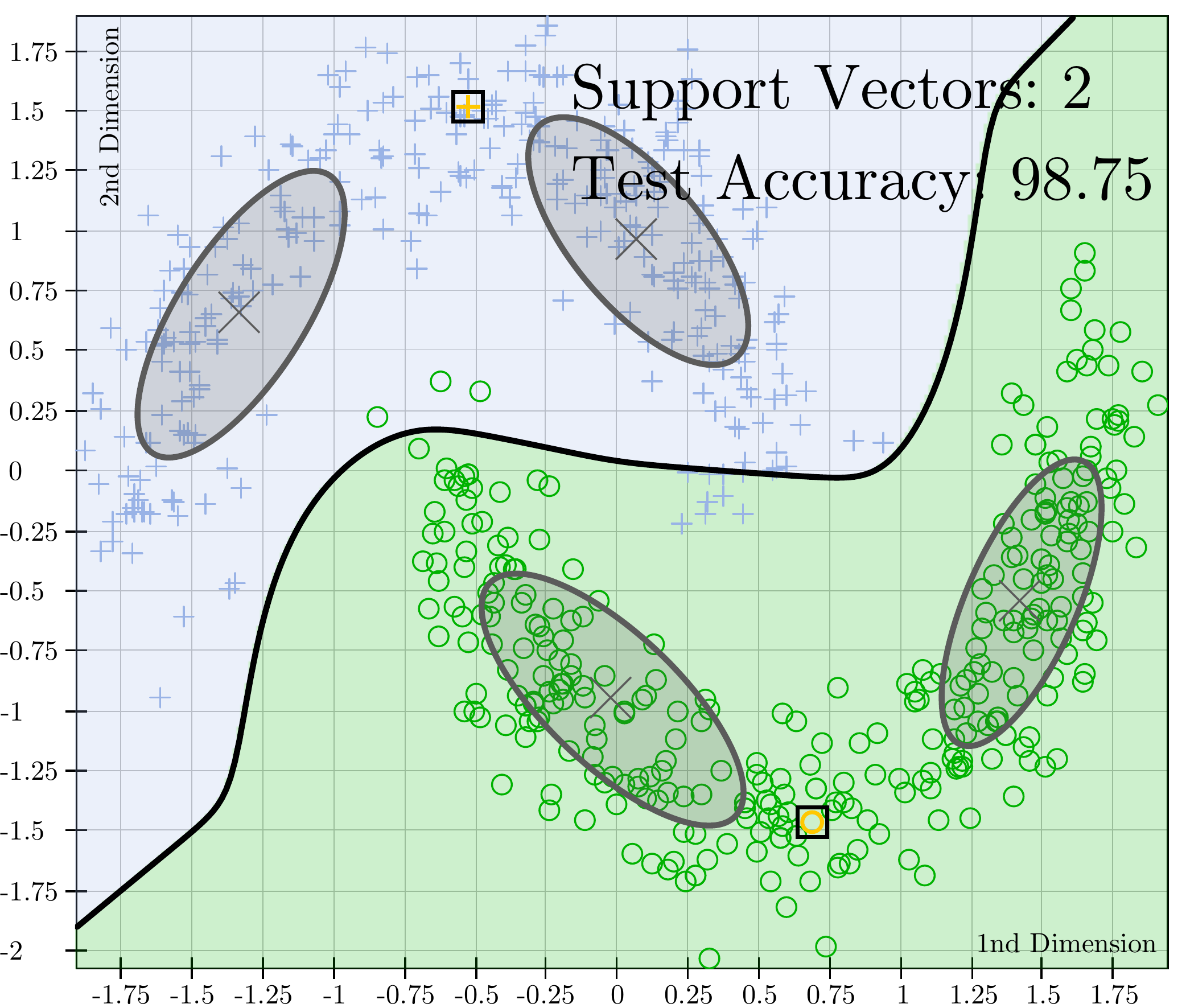}}\hfill
\subfigure[LAP kernel.]{\includegraphics[width=0.24\textwidth]{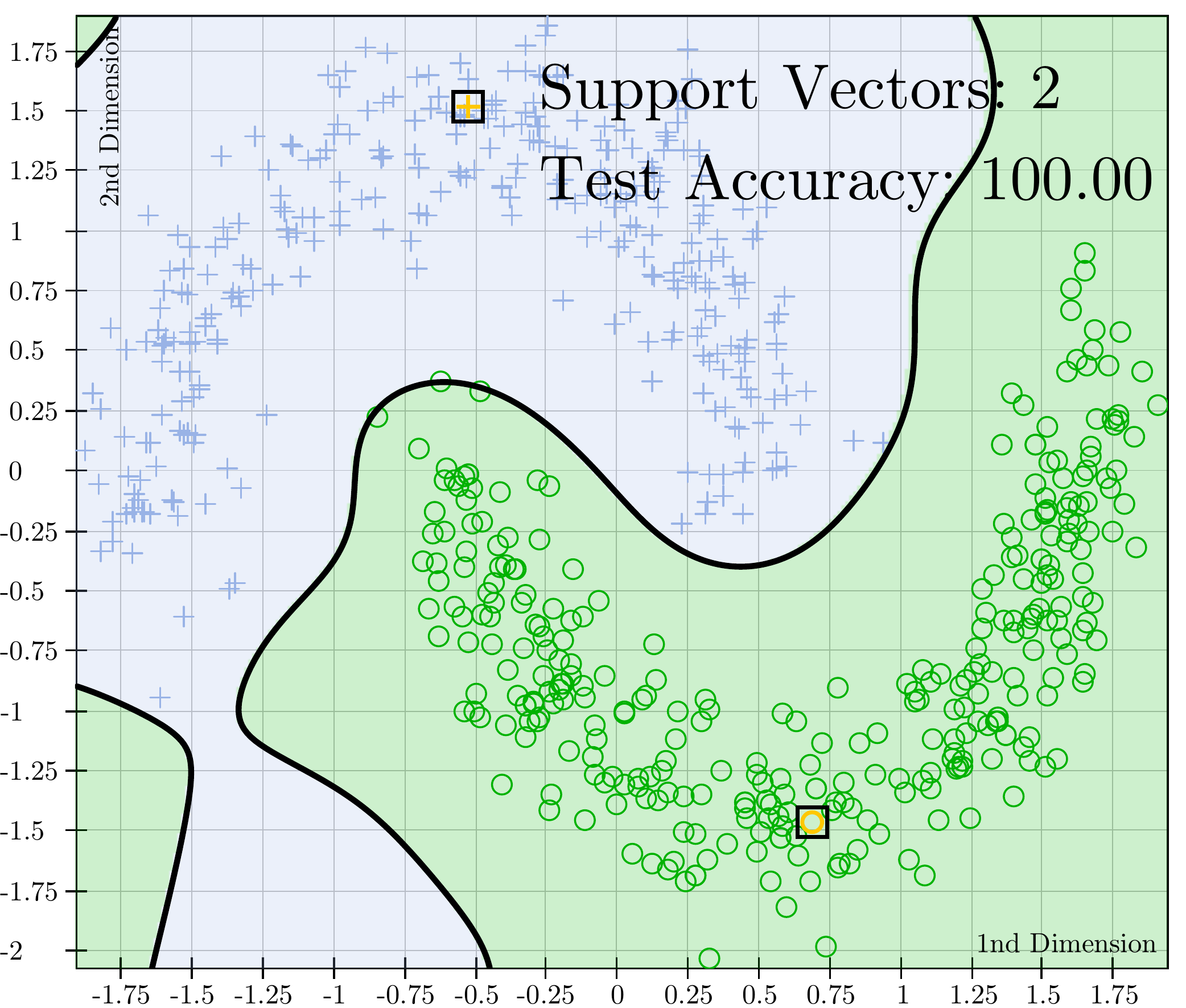}}\hfill
\subfigure[GMM kernel.]{\includegraphics[width=0.24\textwidth]{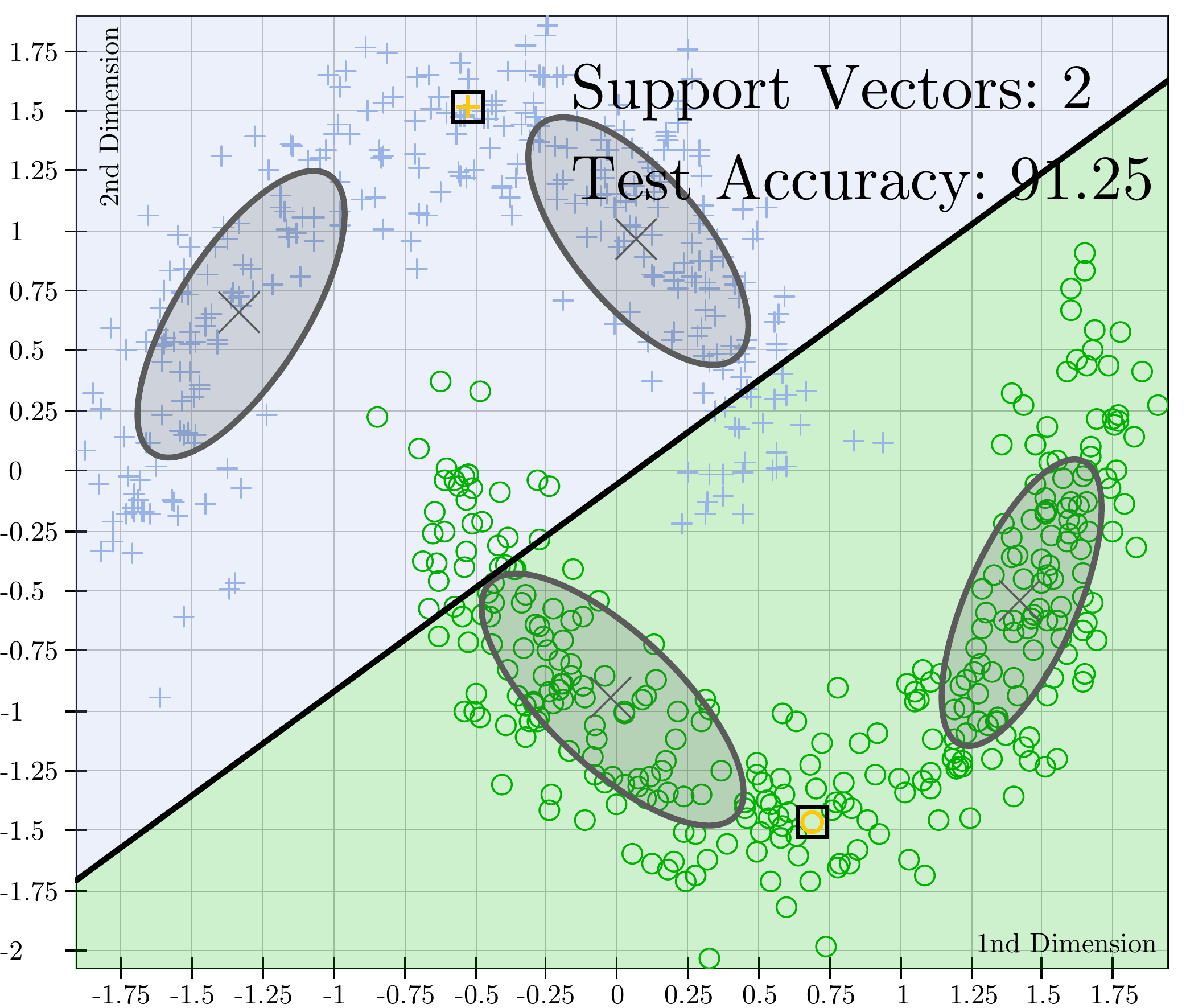}}\hfill
\subfigure[RBF kernel.]{\includegraphics[width=0.24\textwidth]{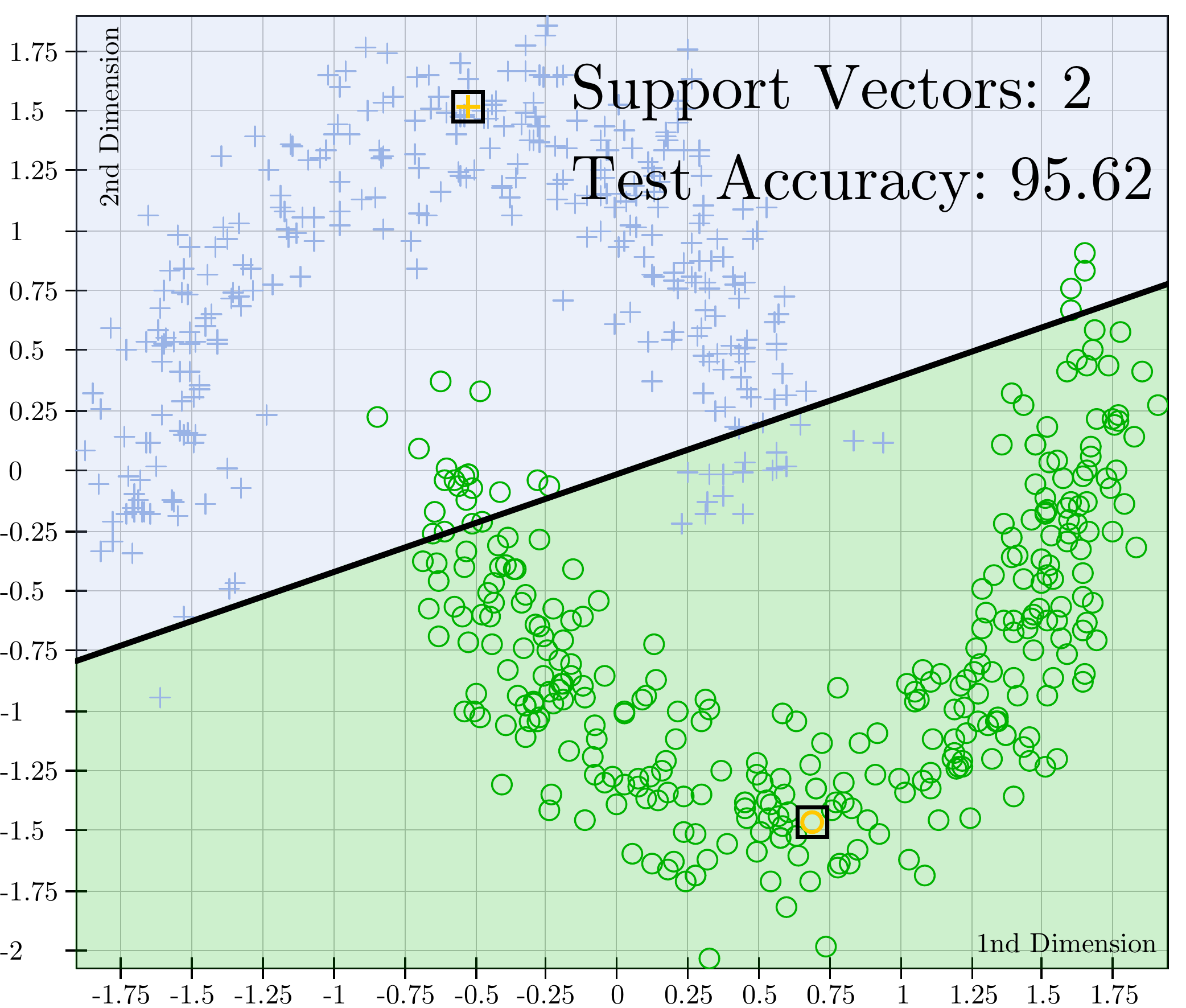}}
\vfill
\vspace{-0.1em}
\subfigure[RWM kernel.]{\includegraphics[width=0.24\textwidth]{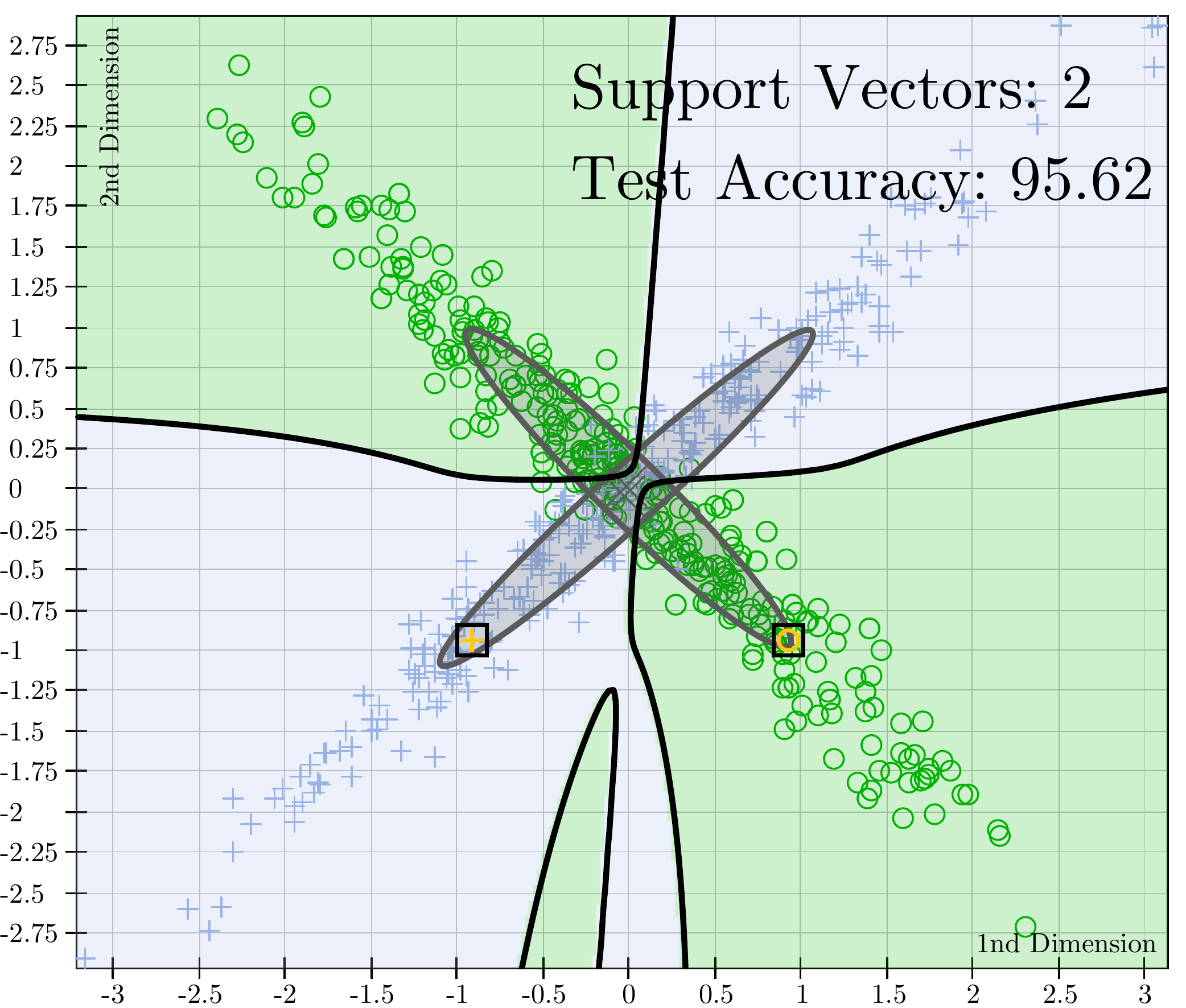}}\hfill
\subfigure[LAP kernel.]{\includegraphics[width=0.24\textwidth]{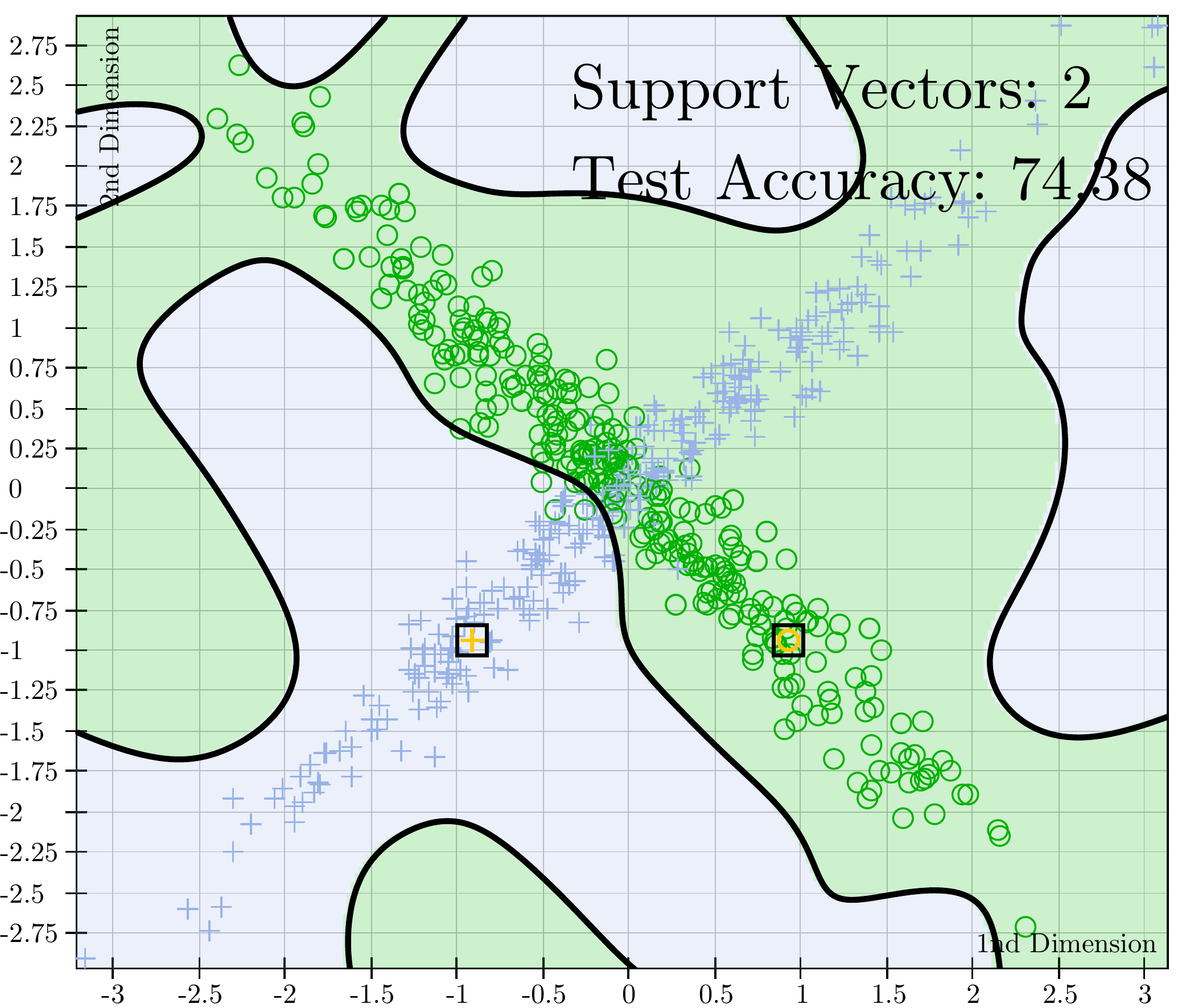}}\hfill
\subfigure[GMM kernel.]{\includegraphics[width=0.24\textwidth]{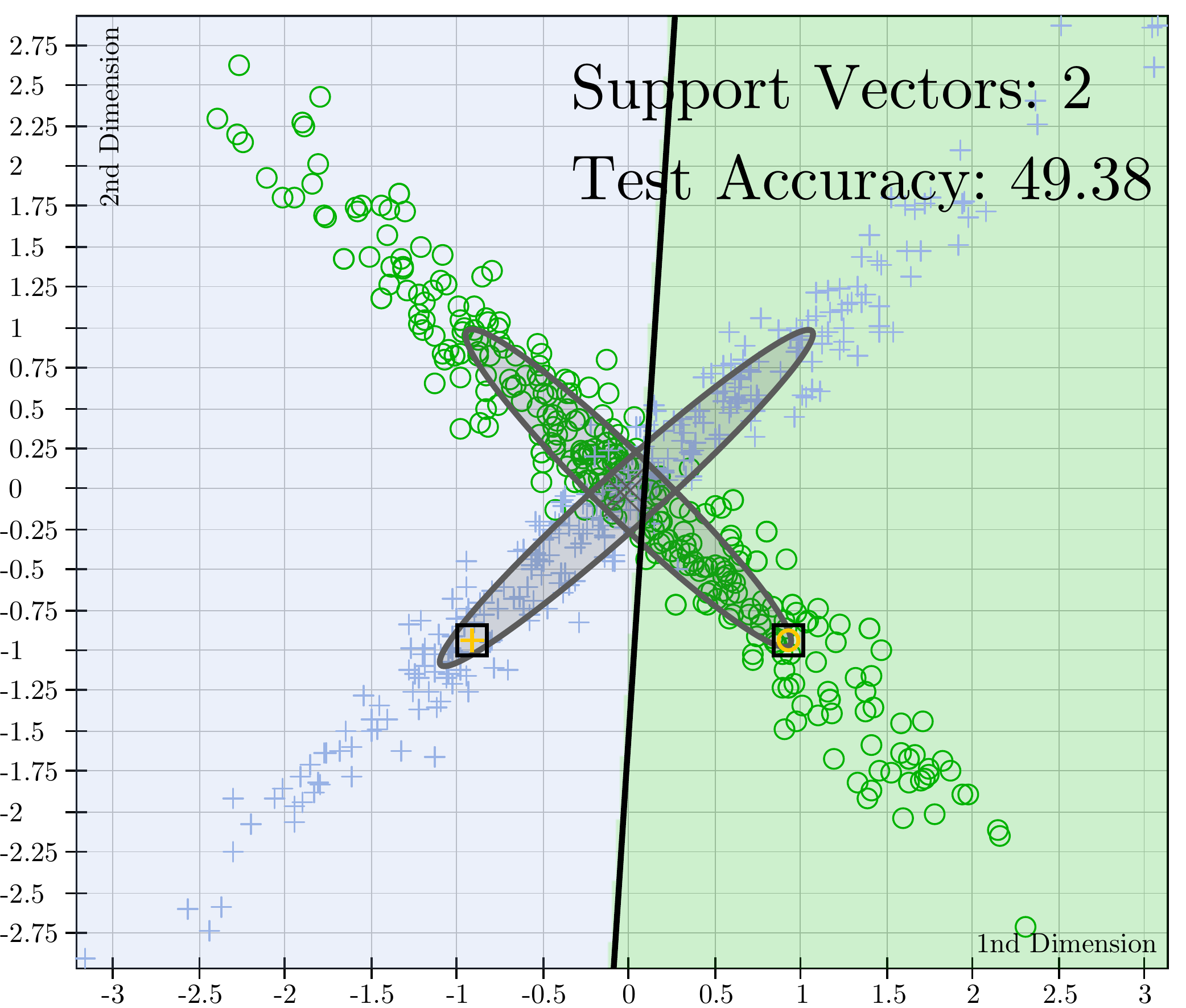}}\hfill
\subfigure[RBF kernel.]{\includegraphics[width=0.24\textwidth]{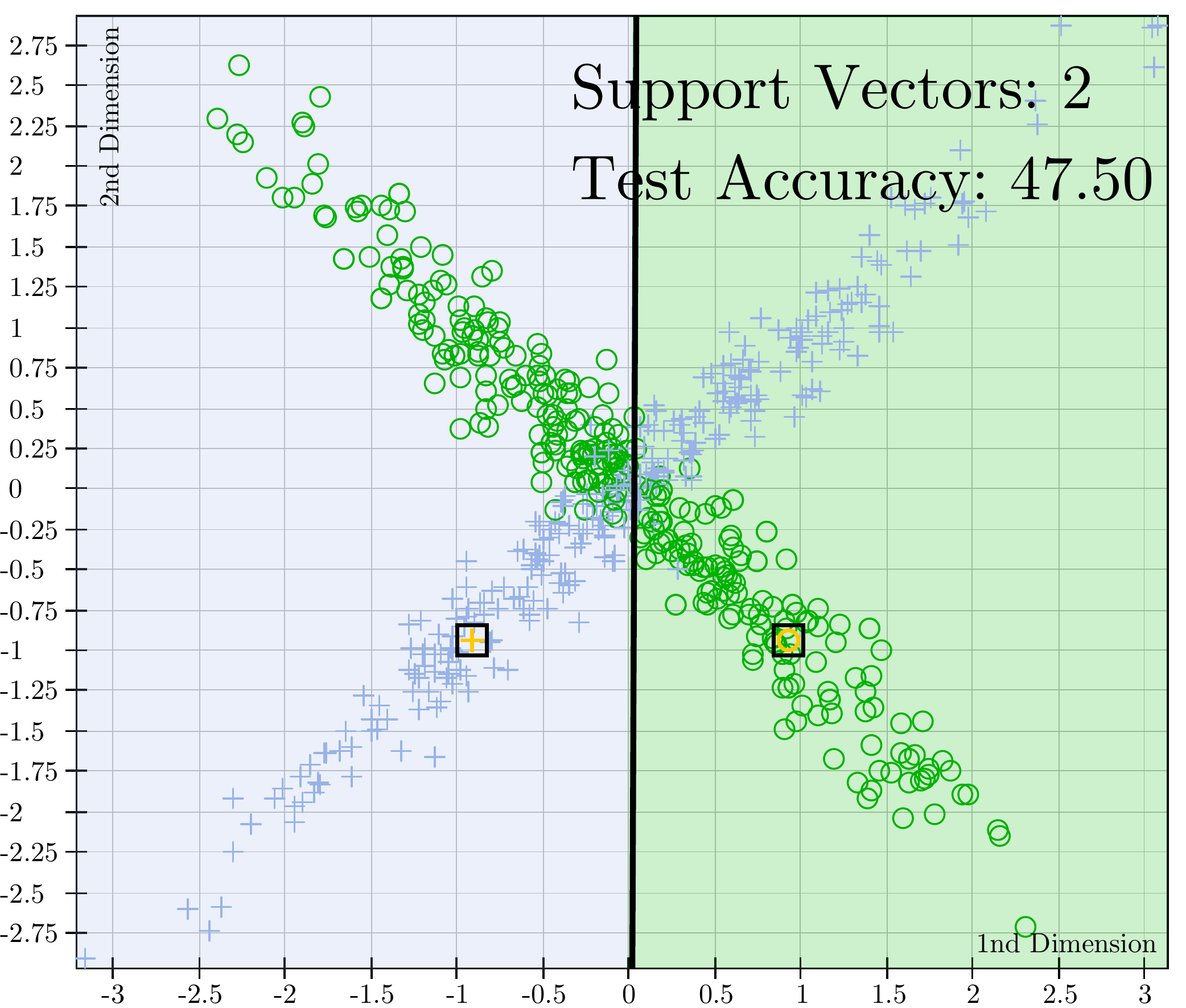}}
\vfill
\vspace{-0.1em}
\subfigure[RWM kernel.]{\includegraphics[width=0.24\textwidth]{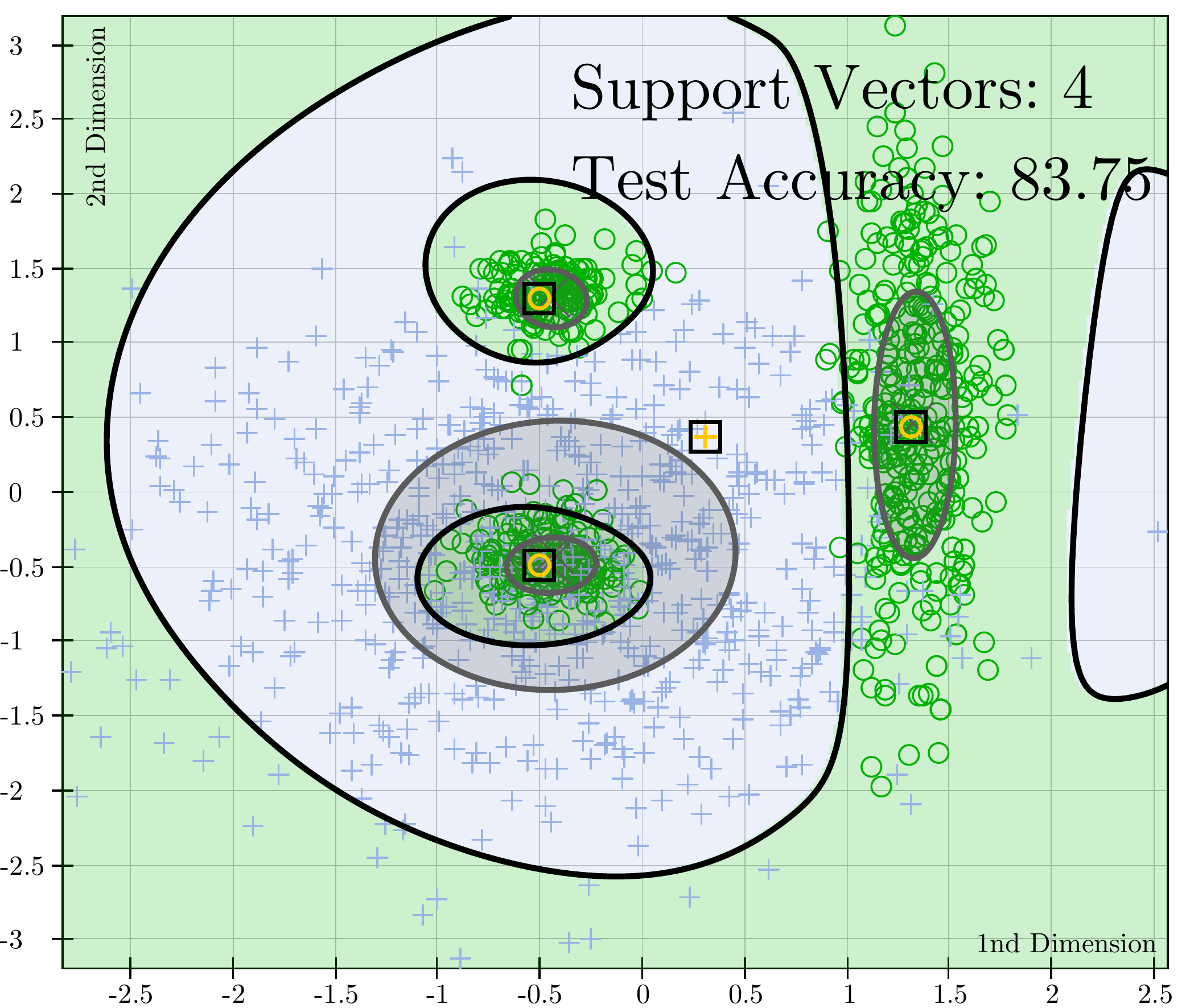}}\hfill
\subfigure[LAP kernel.]{\includegraphics[width=0.24\textwidth]{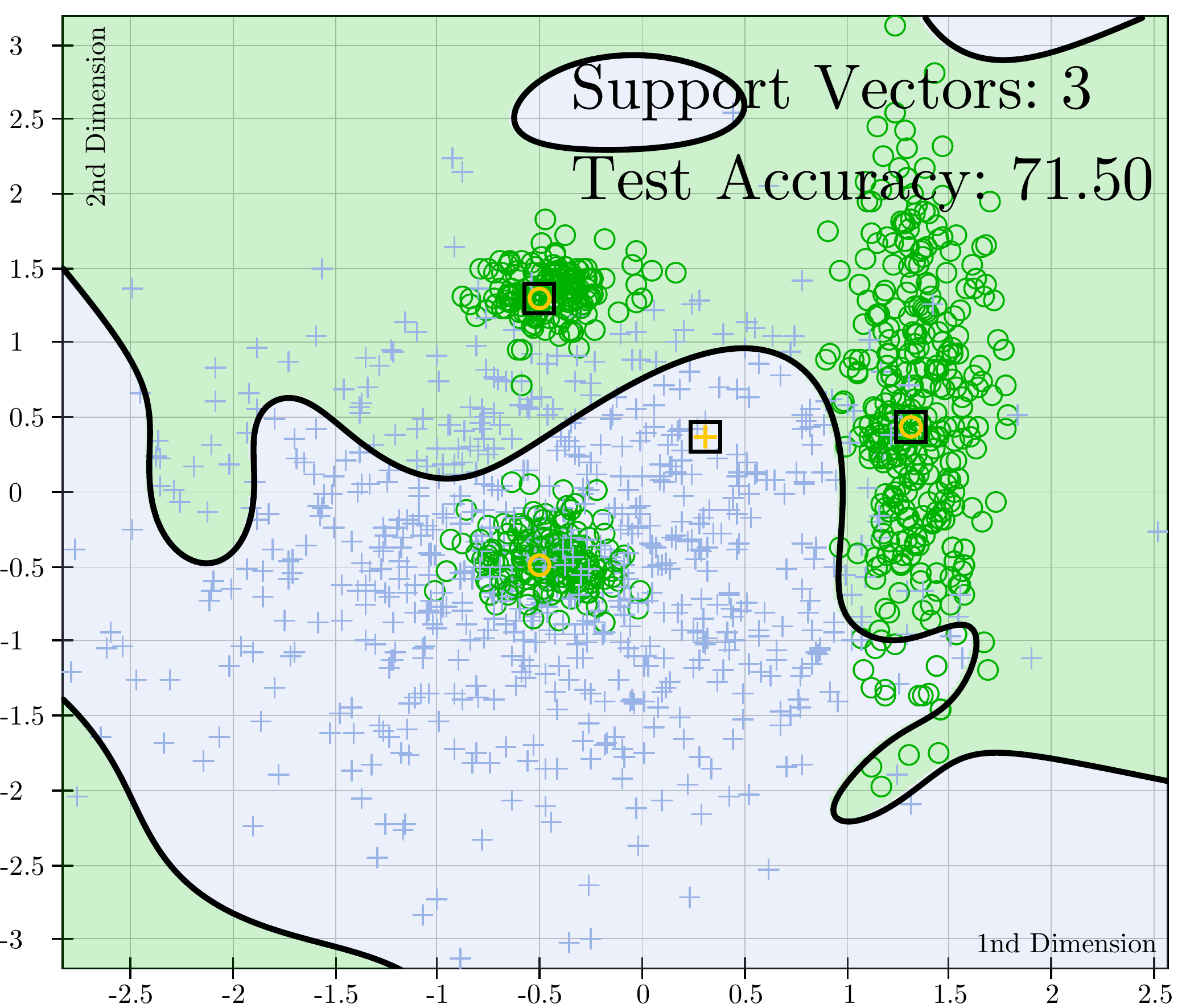}}\hfill
\subfigure[GMM kernel.]{\includegraphics[width=0.24\textwidth]{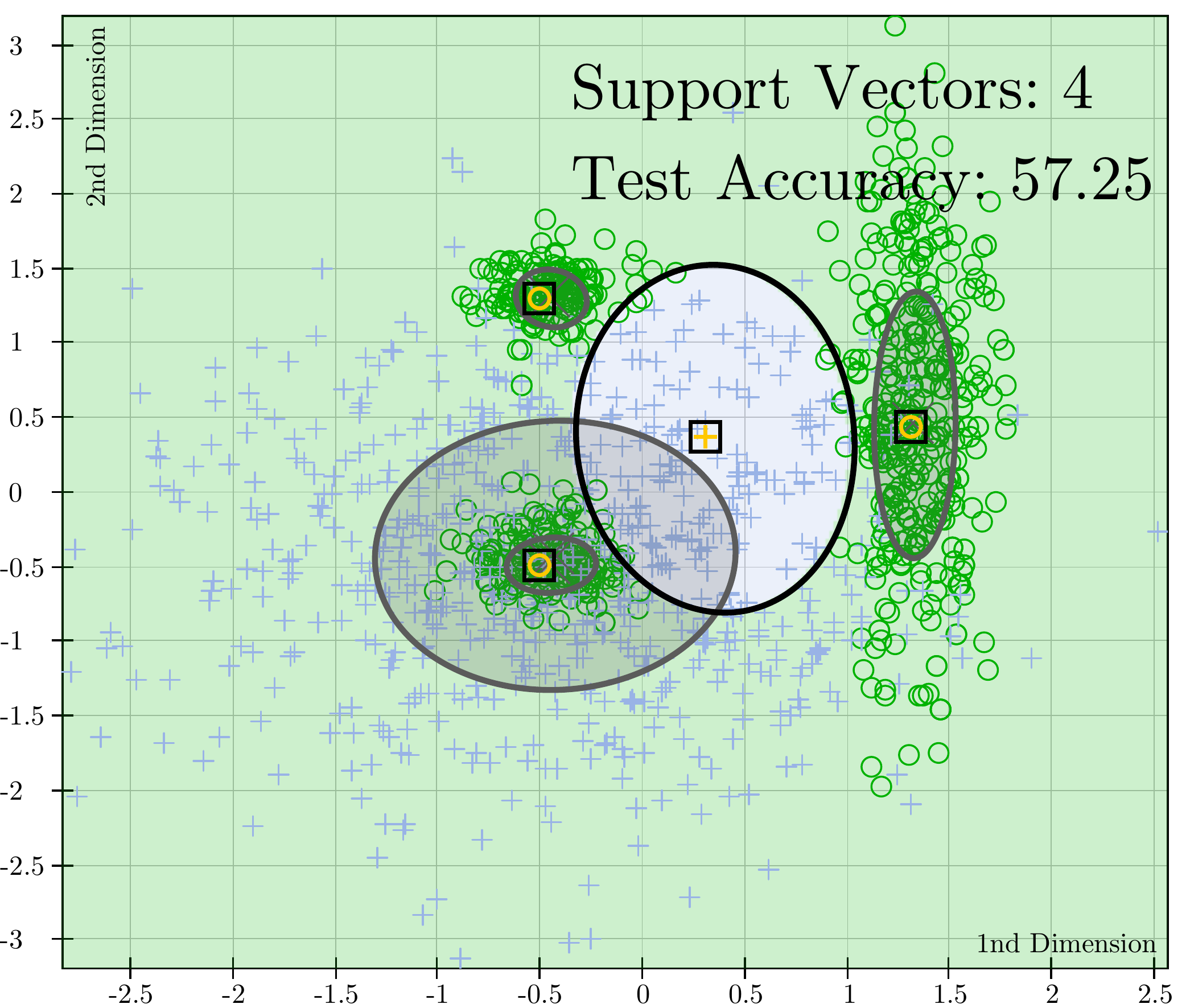}}\hfill
\subfigure[RBF kernel.]{\includegraphics[width=0.24\textwidth]{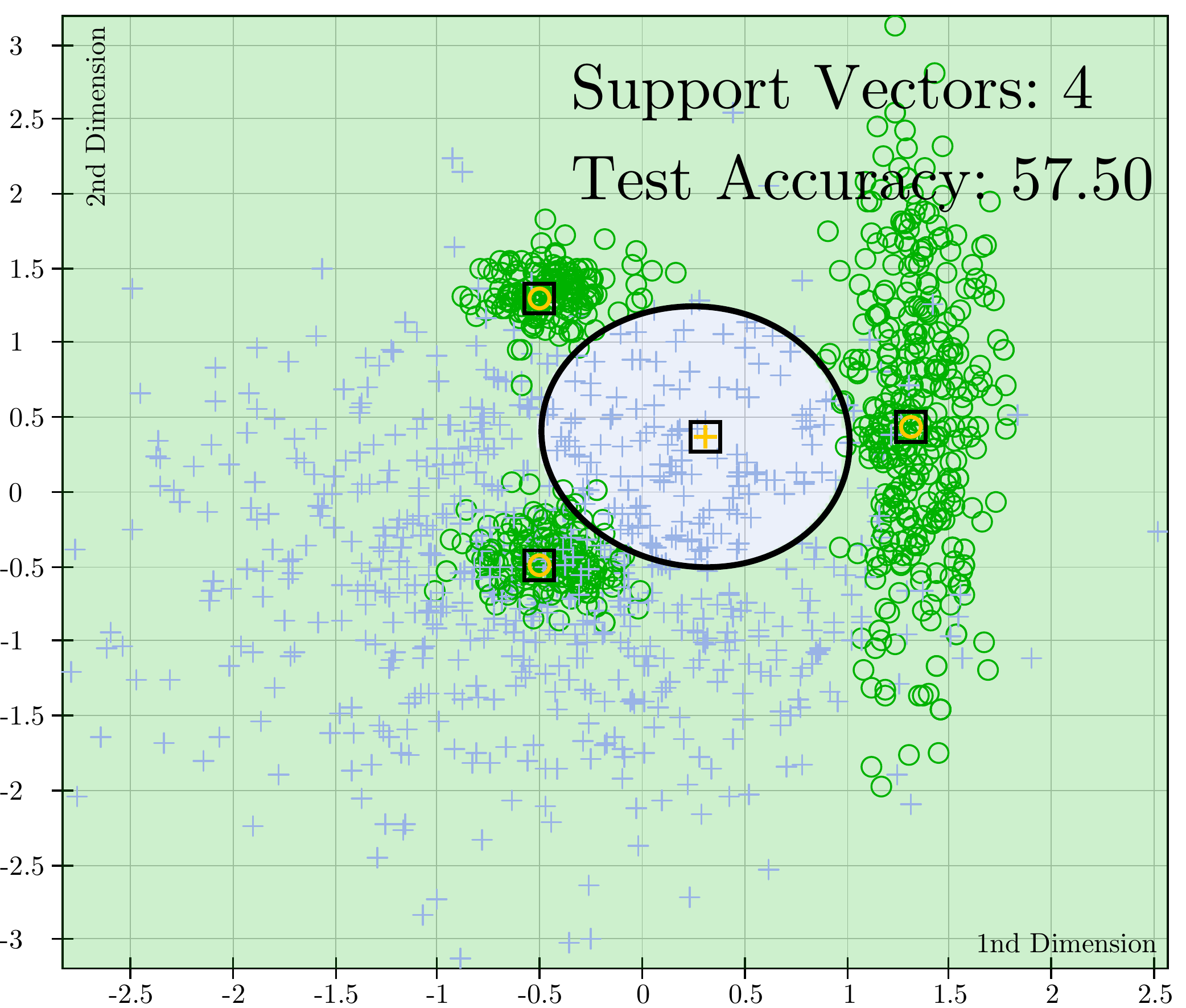}}
\vfill
\vspace{-0.1em}
\subfigure[RWM kernel.]{\includegraphics[width=0.24\textwidth]{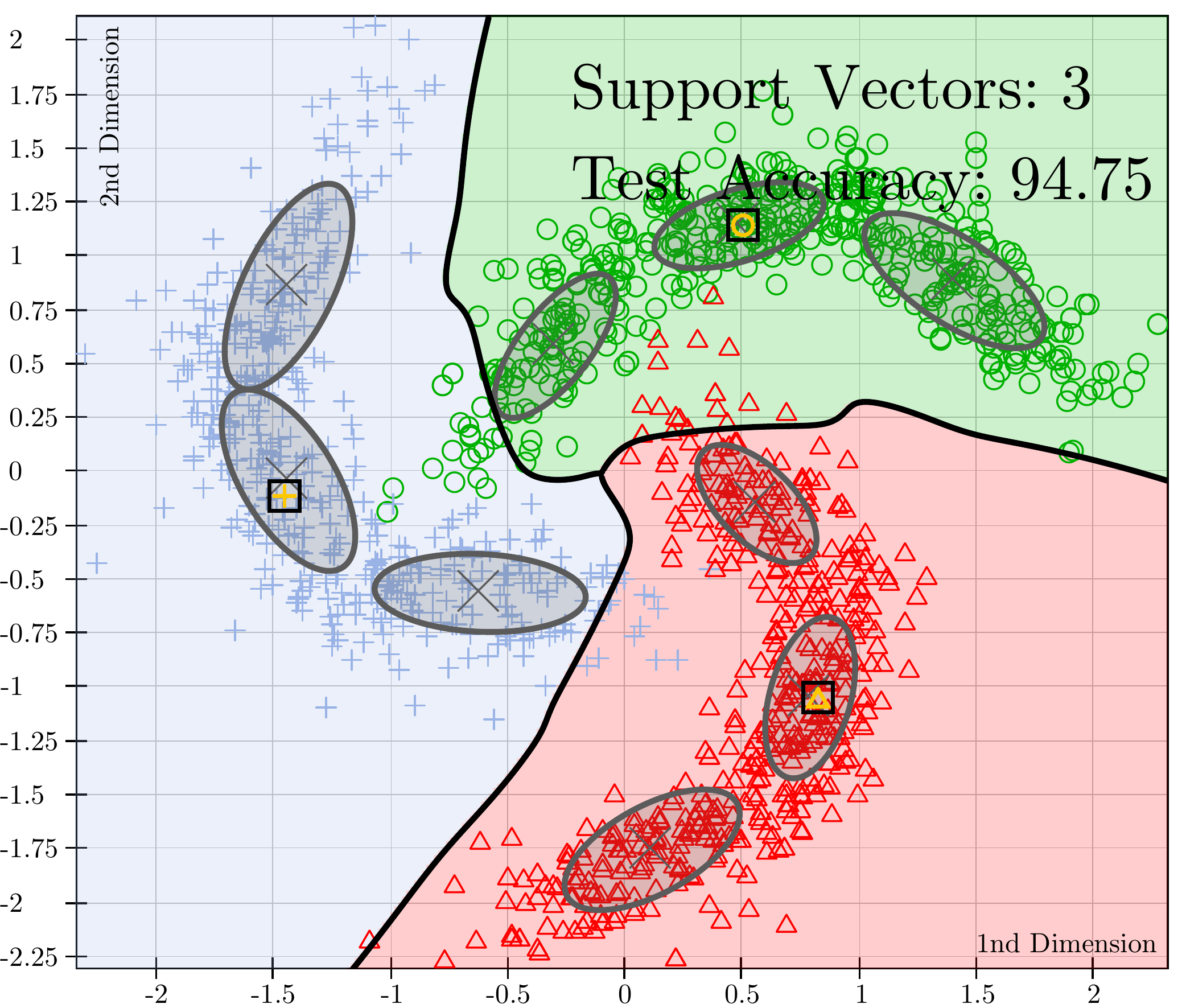}}\hfill
\subfigure[LAP kernel.]{\includegraphics[width=0.24\textwidth]{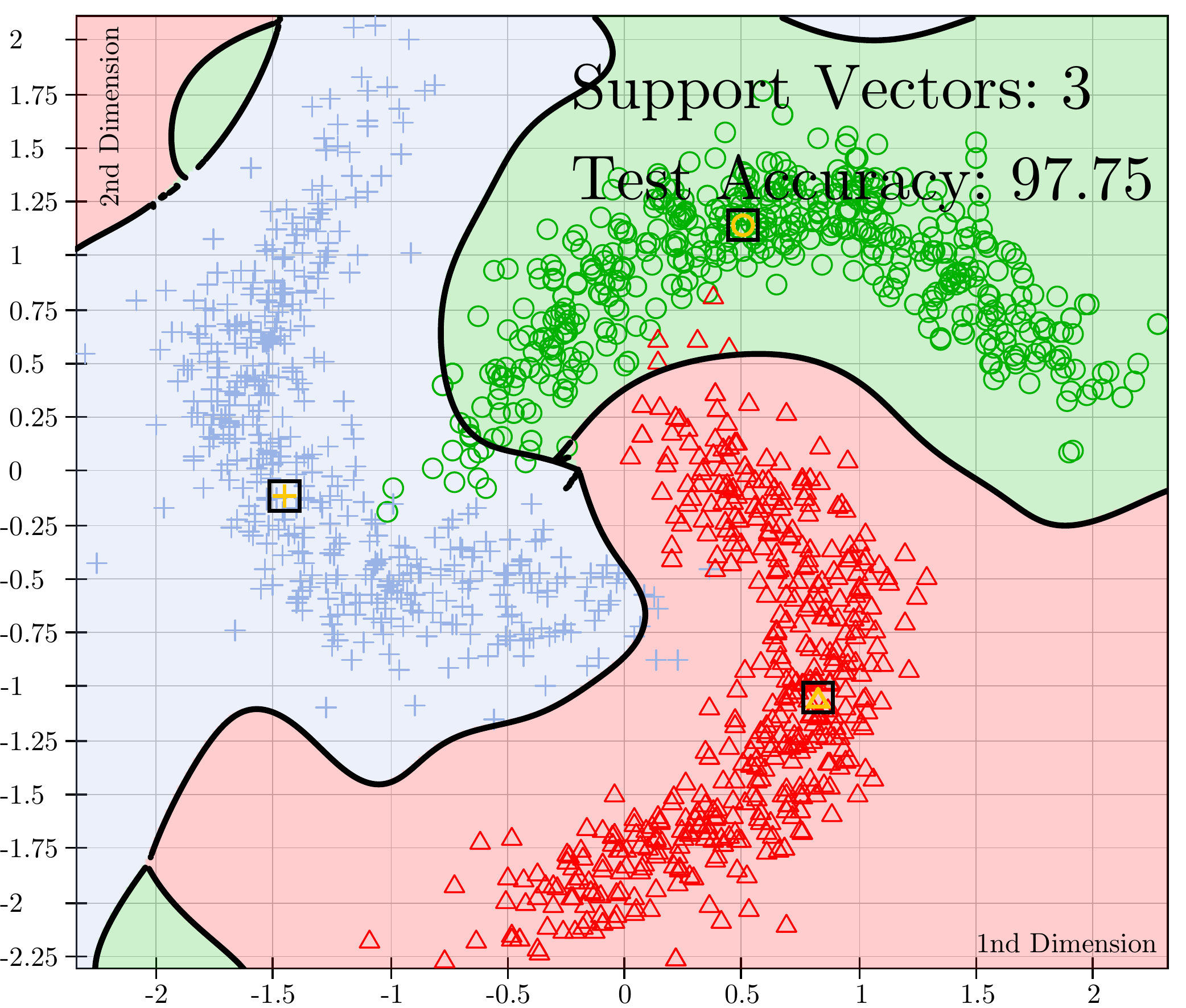}}\hfill
\subfigure[GMM kernel.]{\includegraphics[width=0.24\textwidth]{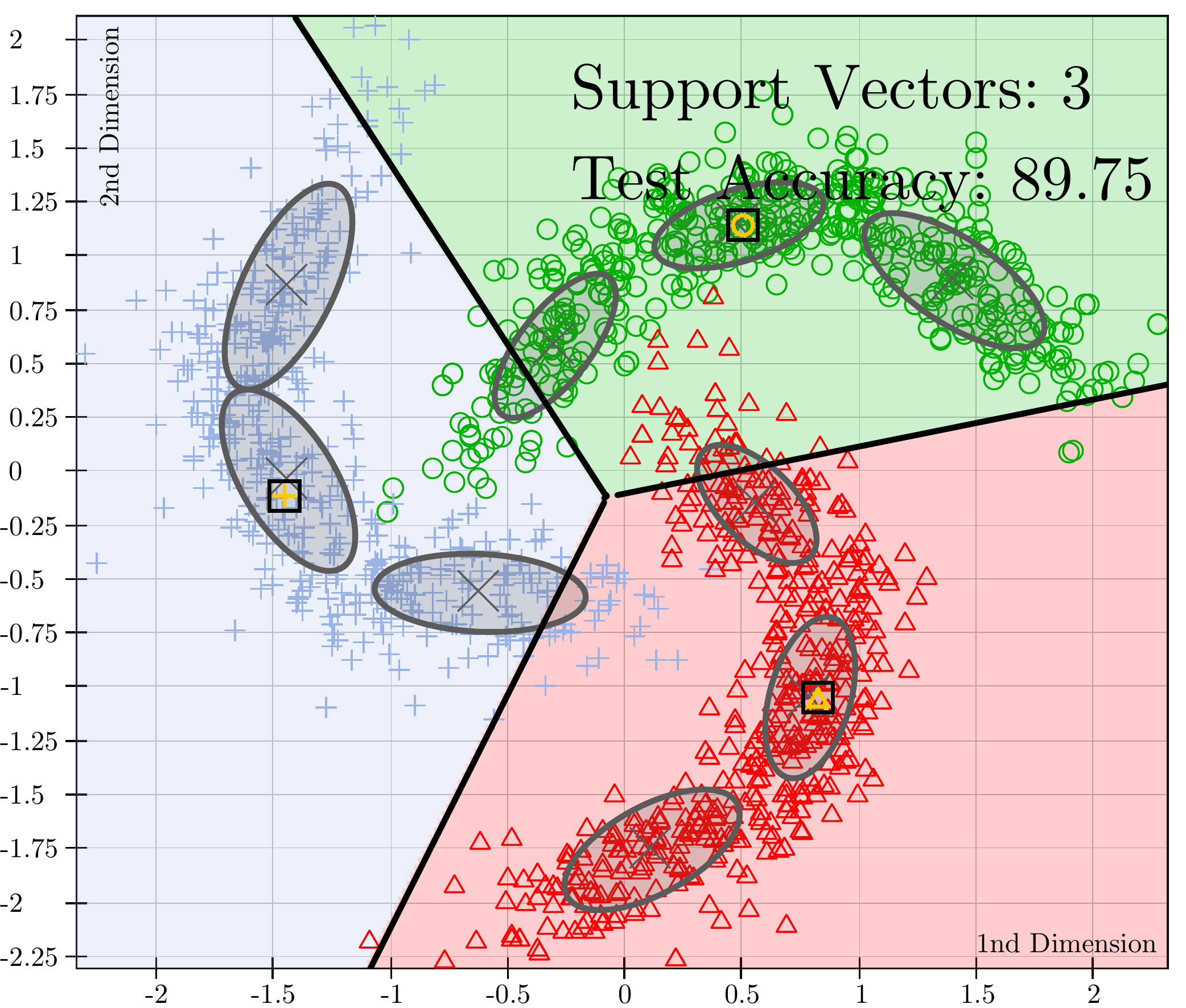}}\hfill
\subfigure[RBF kernel.]{\includegraphics[width=0.24\textwidth]{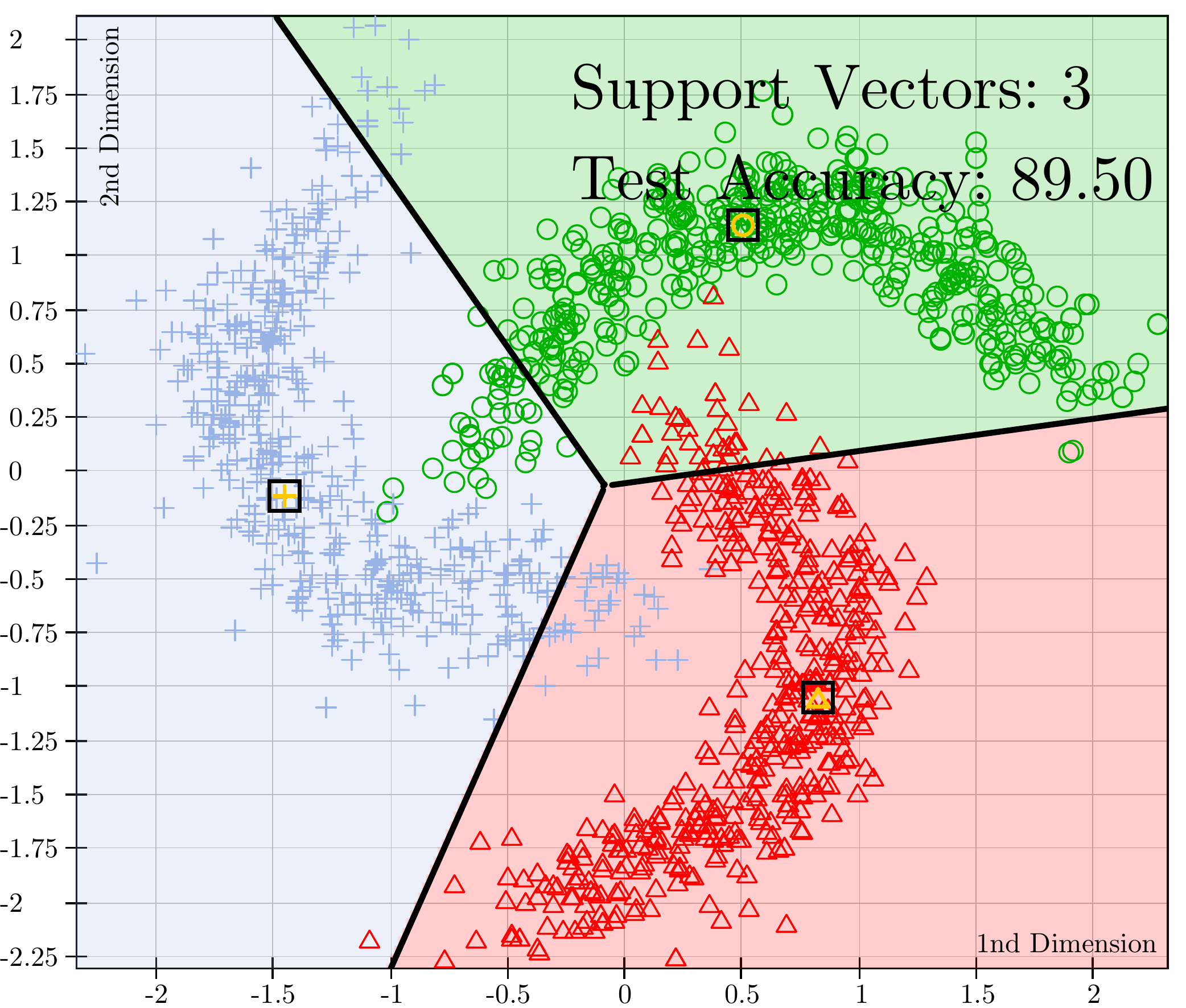}}
\vfill
\vspace{-0.1em}
\subfigure[RWM kernel.]{\includegraphics[width=0.24\textwidth]{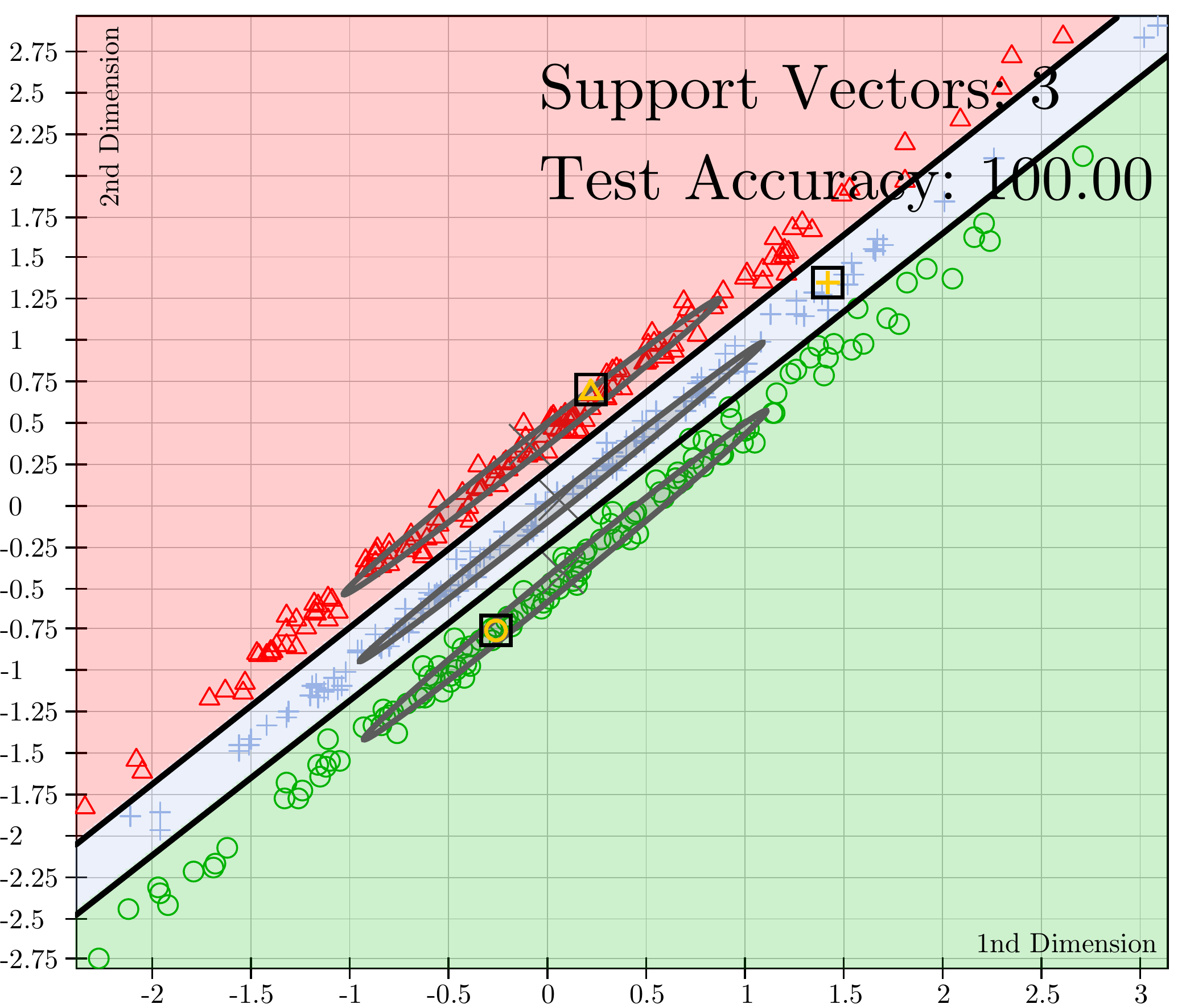}}\hfill
\subfigure[LAP kernel.]{\includegraphics[width=0.24\textwidth]{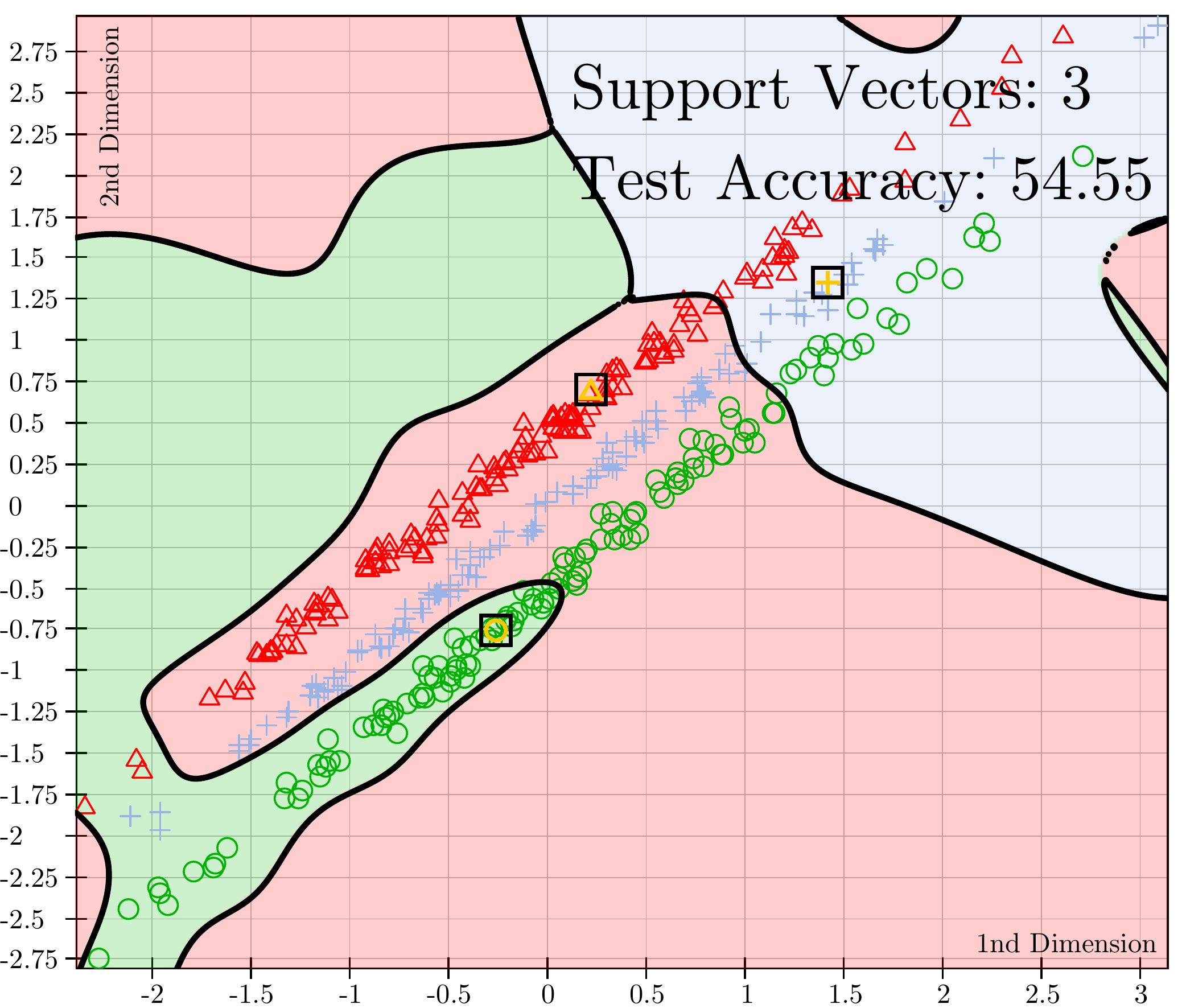}}\hfill
\subfigure[GMM kernel.]{\includegraphics[width=0.24\textwidth]{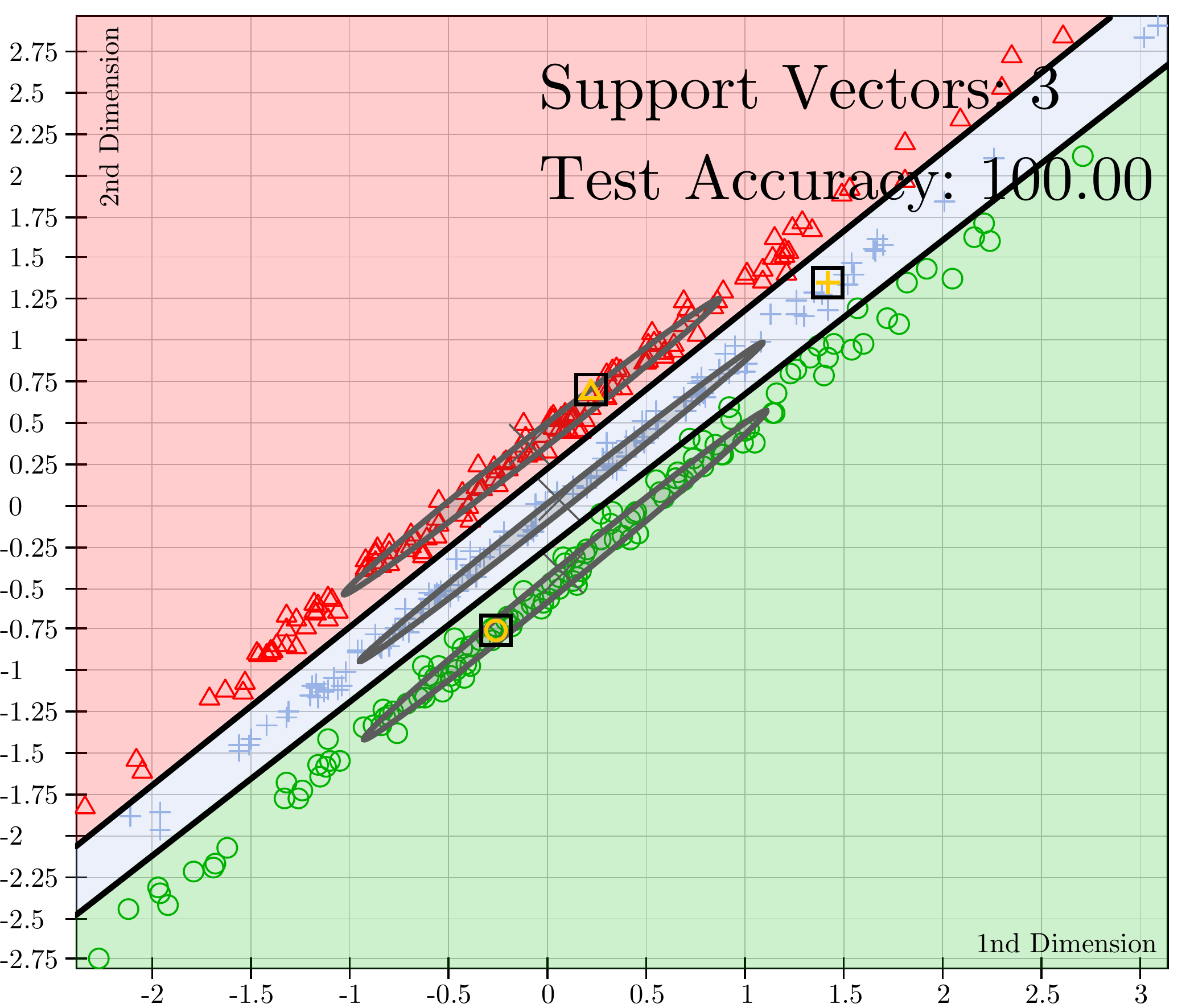}}\hfill
\subfigure[RBF kernel.]{\includegraphics[width=0.24\textwidth]{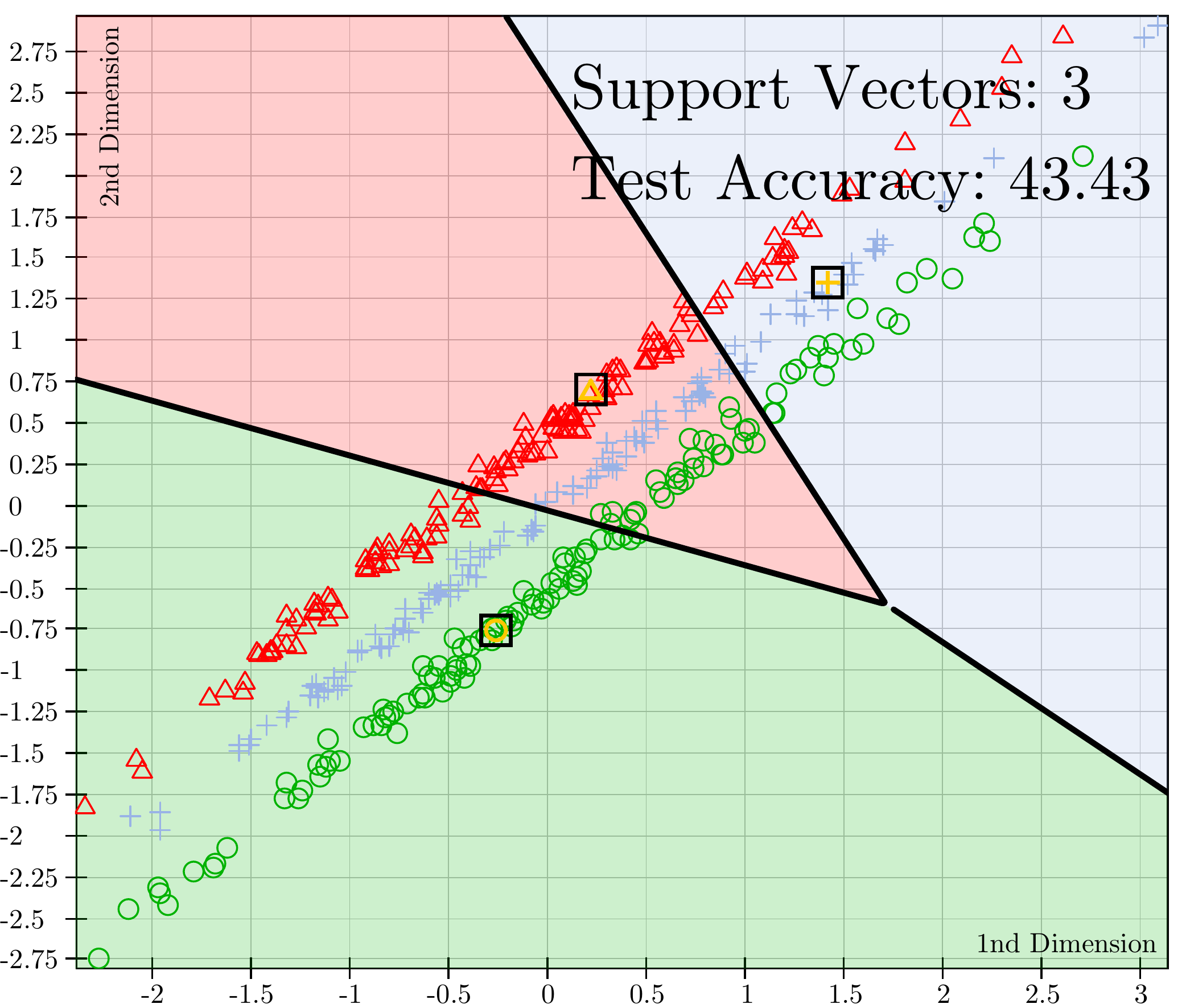}}
\begin{center}
\vspace{-0.5em}
\caption{Performance comparison of SVM classifiers with RWM, LAP, GMM, and RBF kernels trained on different synthetically generated data sets. For each row the same data set is used, in the following order (from top to bottom): Two Moons, Cross, Clouds, Three Moons, and Adidas.}
\label{fig:comparison_best_case}
\end{center}
\end{figure*}

\subsection{Behavior of SVM using RWM Kernels}
\label{sec:behavior}

To visualize the behavior of an SVM with RWM and other kernels in the presence of very sparse data we took five artificial data sets: The well-known data sets Two Moons (suggested in \cite{Melacci12}) and Clouds (from the UCI Machine Learning Repository \cite{AN07}), and three additional data sets, called Cross, Three Moons, and Adidas, generated by mixtures of Gaussians (for more information, please send an email to one of the authors). We performed a $z$-score normalization for all five data sets and conducted a stratified 5-fold cross-validation. To get the best possible classification result for each kernel function, we exceptionally (i.e., other than in Section~\ref{sec:benchmarkDataSets}) optimized the parameters with respect to the test set. For this, we applied an exhaustive search by varying $C = 10^{i}$ and $\gamma = 10^{i}$ for $i \in \{-3,-2, \dots, 2\}$ and the additional parameters of the LAP kernel $\gamma_{I} = 10^{j}$, $\gamma_{A} = 10^{j}$ for $j \in \{-7,-6, \dots, 4\}$, $k \in \{5,7,9\}$ and $p=1$. Information about the parametrization of the mixture density model underlying the RWM and GMM kernels can be found in Section~\ref{sec:benchmarkDataSets}.

Fig.~\ref{fig:comparison_best_case} shows for each data set the resulting SVM with RWM, LAP, GMM, and RBF kernels from the first cross-\-vali\-da\-tion fold. The orange colored samples correspond to the labeled training samples used by the SMO algorithm, whereas the remaining samples are used to construct the kernel in the case of RWM, LAP, and GMM. A training sample framed by a black square indicates that this sample is a support vector. The black solid line is the decision boundary and gray colored ellipses (in the case of the RWM and GMM kernels) correspond to level curves of Gaussians that are located at centers indicated by large $\times$s. 

In Fig.~\ref{fig:comparison_best_case} we can see that SVM with RBF kernels perform worst. This is not surprising as this kernel does not take advantage from the unlabeled data at all. That is, in the presence of sparsely labeled data the usage of structure information derived from unlabeled data helps to achieve significantly better classification results. One might assume further that a kernel based on a non-parametric density modeling approach such as the LAP kernel performs relatively well even if the generating processes of the data produce clusters with non-convex shapes. Actually, SVM with LAP kernel yield on the data sets Two Moons and Three Moons better results than an SVM using our new RWM kernel based on a parametric density modeling approach. However, the results are only slightly better than with a GMM or an RWM kernel because convex and non-convex clusters can both be modeled with mixtures of Gaussians. On the remaining data sets, SVM with RWM kernel achieve noticeably better results than an SVM with one of the other kernel functions. It is obvious that the LAP kernel has problems if clusters with different class affiliations are overlapping (Cross and Clouds data sets) or if the clusters are not clearly separated (Adidas data set). Interestingly, the new RWM kernel performs either better (four data sets) or equal (one data set) than its relative, the GMM kernel. The GMM kernel is derived from the Gaussian mixture model in straight forward way, while the RWM kernel gives higher weights to components that are responsible for any two samples that are assessed.


\subsection{Comparison based on 20 Benchmark Data Sets}
\label{sec:benchmarkDataSets}

To evaluate the performance of an SVM using our new RWM kernel numerically and in more detail, we conduct experiments with 20 publicly available data sets. Thus, we are able to come to statistically significant conclusions concerning the new kernel. In addition, we conduct run-time measurements and analyze the computational complexity of our new approach.

\subsubsection{Data Sets and Experimental Setup}
\label{sec:labelAndSetup}

For our experiments, we use 20 data sets: 14 real-world data sets (Australian, Credit A, Credit G, Ecoli, Glass, Heart, Iris, Page Blocks, Pima, Seeds, Vehicle, Vowel, Wine, and Yeast) from the UCI Machine Learning Repository \cite{AN07}, two real-world (Phoneme and Satimage) and two artificial data sets (Clouds and Concentric) from the UCL Machine Learning Group \cite{UCL14}, and in addition two artificial data sets, (Ripley) suggested in \cite{Ripley96} and (Two Moons) suggested in \cite{Melacci12}. In order to obtain meaningful results regarding the performance of our new RWM kernel, we consider three requirements for the selection of the data sets: First, the majority of the data sets should come from real life applications. Second, the data sets should have very different numbers of classes. And third, some of the data sets should have unbalanced class distributions. The description of the data sets is summarized in Table~\ref{tab:DataSetsBasic}.

\begin{table*}[htbp!]
\caption[Data set information]{General data set information.}
\label{tab:DataSetsBasic}
\ifpreprint
\tiny
\else
\scriptsize
\fi
\begin{center}
\newcolumntype{C}[1]{>{\centering}m{#1}}
\begin{tabular}{|l|C{4.7em} C{4em} C{4em} C{4.7em} c|} 
\hline
\multirow{3}{*}{Data Set} & \multicolumn{5}{c|}{Description}\\
&  Number\ of & Continuous & Categorical & Number\ of & Class \\
&  Samples &  Attributes\ & Attributes & Classes & Distribution (in \%)\\
\hline
\hline
Australian 	&690		&6		&8		&2	& 55.5,44.5\\
\hline
Clouds 			&5000		&2		&--		&2	& 52.2,50.0\\
\hline
Concentric    &2500      &2     &–-    & 2& 36.8, 63.2\\
\hline
Credit A	   &690		   &6		&9		&2	& 44.5,55.5\\
\hline
Credit G 		&1000		&7		&13		&2	&70.0,30.0\\
\hline
Ecoli			&336		&7		&--		&8 &42.6,22.9,15.5,10.4,5.9,1.5,0.6,0.6\\
\hline
Glass			&214		&9		&--		&6&32.7,35.5,7.9,6.1,4.2,13.6\\
\hline
Heart			&270		&6		&7		&2 & 44.4,55.6\\
\hline
Iris         & 150      &4      & --     & 3 & 33.3,33.3,33.3 \\
\hline
Page Blocks     &5473   &10     & --    & 5 &89.8,6.0,0.5,1.6,2.1\\
\hline
Phoneme			&5404		&5		&--		&2	& 70.7,29.3\\
\hline
Pima 			&768		&--		&8		&2	& 65.0,35.0\\
\hline
Ripley			&1250		&2		&--		&2	&50.0,50.0\\
\hline
Satimage 		&6345		&5		&--		&6	&24.1,11.1,20.3,9.7,11.1,23.7\\
\hline
Seeds         & 210       & 7    & -- & 3 & 33.3,33.3,33.3 \\
\hline
Two Moons 		&800		&2	    &--		&2	&50,50\\
\hline
Vehicle			&846		&18		&--		&4	&23.5,25.7,25.8,25.0\\
\hline
Vowel			&528		&10		&--		&11&9.1,9.1,9.1,9.1,9.1,9.1,9.1,9.1,9.1,9.1,9.1\\
\hline
Wine 			&178		&13		&--		&3&33.1,39.8,26.9\\
\hline
Yeast & 1484 & 8 & -- & 10 & 16.4,28.1,31.2,2.9,2.3,3.4,10.1,2.0,1.3,0.3 \\
\hline
\end{tabular}
\end{center}
\end{table*}

To find good estimates for the hyper-parameters of the VI algorithm (training of the mixture density models capturing structure information in unlabeled data) we used an exhaustive search. To rate a considered set of VI parameters we applied an interestingness measure, called \textit{representativity} \cite{FKS11}. It measures the dissimilarity of the mixture density model trained with VI and a density estimate resulting from a non-parametric Parzen window estimation. As dissimilarity measure we used the symmetric Kullback-Leiber divergence instead of the Hellinger distance mentioned in \cite{FKS11}.

\begin{figure}[htbp!]
	\begin{center}
	\subfigure[Outer cross-validation (one fold).]{\label{fig:subfig1stCrossValidation}\includegraphics[width=0.46\textwidth]{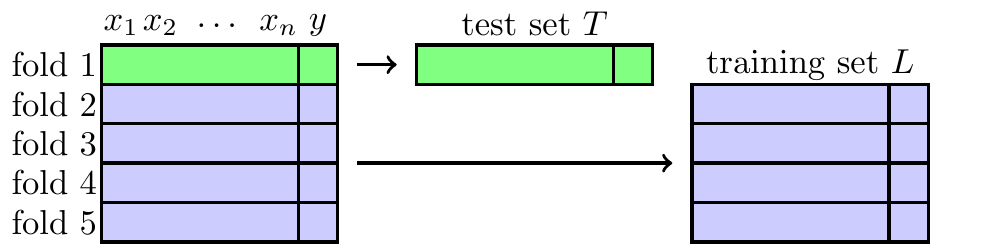}}
	\quad	
	\subfigure[Inner cross-validation (one fold).]{\label{fig:subfig2ndCrossValidation}\includegraphics[width=0.46\textwidth]{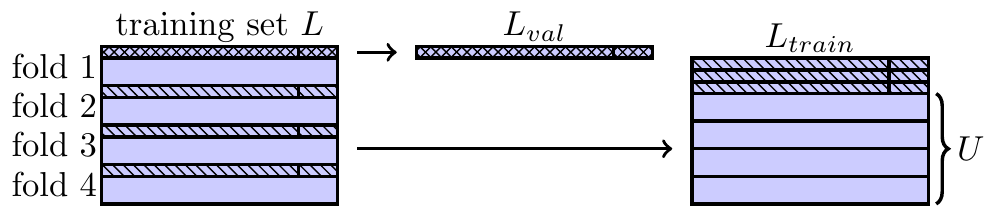}}
	\caption{Disjoint subsets of one fold of the (outer) 5-fold cross-validation. $L_{val} \cup L_{train}$ correspond to a randomly chosen subset from the ``training'' folds (training set $L$). The remaining samples of $L$ build the subset $U$.}
	\label{fig:crossValidation}
	\end{center}
\end{figure}

In our experiments, we performed a $z$-score normalization for all data sets and conducted a stratified 5-fold cross-validation evaluation, as sketched in Fig.~\ref{fig:crossValidation}. In each round of the outer cross-validation, one fold is kept out as \textit{test set} $T$. Of course, $T$ is not considered for any parametrization purposes. The other four folds are used as training set $L$ (cf.\ part (a) of Fig.~\ref{fig:crossValidation}). To simulate the presence of sparsely labeled data, we selected subsets of different sizes -- $4\times \text{the number of classes}$ (experiment 1), $10\%$ of $|L|$ (experiment 2), and $100\%$ of $|L|$ (experiment 3) -- from the training set folds (cf.\ dashed boxes in part (b) of Fig.~\ref{fig:crossValidation}). Precisely, in experiment 1, we chose samples lying in high density regions randomly ($p(\mathbf{x})$ given by the mixture density model). In experiment 2, the selected samples of experiment 1 are enriched with randomly selected samples until the number of ten percent of $L$ is obtained. To get good parametrization results we applied an inner 4-fold cross-validation to these subsets. Here, one fold is used as validation set $L_{val}$ and the other three folds as training set $L_{train}$. The non-randomly chosen samples from $L$ build a subset $U$ that is only considered for capturing structure information (i.e., without class assignments). Consequently, the whole training set $L= L_{val} \cup L_{train} \cup U$ is used to determine the Laplacian graph in case of the LAP kernel and to determine the Gaussian mixture model in case of the RWM and GMM kernels. To rate a considered parameter combination we determined the classification performance by considering $L_{val}$ and $U$ (to determine the expected error) simultaneously.

The penalty parameter $C = 10^i$ and the kernel width $\gamma = 10^i$ were varied for $i \in \{-3,-2, \dots 2\}$, the additional parameters of the LAP kernel $\gamma_{I} =10^{j}$ and $\gamma_{A} = 10^{j}$ for $j \in \{-6,-5, \dots, 2\}$, the neighborhood size was fixed to $k=7$ and the degree to $p=1$. To account for information from catgorical input dimensions we adapted all kernel functions in the same way as the RWM similarity (described in Section~\ref{sec:extensionRWM}). Consequently, to find the best values ​​of $\alpha$ (weighting factor of continuous input dimensions) and $\beta$ (weighting factor of discrete input dimensions) we varied $ \alpha$ and $\beta $ from $0$ to $1$ in step sizes of $0.1$ (for the data sets Australian, Credit A, Credit G, Heart, and Pima that have categorical attributes, cf.~Table~\ref{tab:DataSetsBasic}).


To assess our results numerically, we rank the classification paradigms based on a (non-parametric statistical) Friedman test \cite{Friedman40}. The Friedman test ranks -- considering a given significance value $\alpha$ -- $S$ classifiers for each of $N$ data sets separately, in the sense that the best performing classifier gets the lowest rank, a rank of 1, and the worst classifier the highest rank, a rank of $S$. In case of ties, the Friedman test assigns averaged ranks. Let $r_{i}^{j}$ be the rank of of the $i$-th classifier on the $j$-th data set, then the Friedman test compares the classifiers based on the averaged ranks $R_j = \frac{1}{N}\sum_{i=1}^{S}{r_{i}^{j}}$. Under the null hypothesis, which claims that all classifiers are equivalent in their performance and hence their averaged ranks $R_j$ should be equal, the Friedman statistic is distributed according to the $\chi_{F}^{2}$ distribution with $S-1$ degrees of freedom \cite{JS11}. The Friedman test rejects the null hypothesis if Friedman's $\chi_{F}^{2}$ is greater than the $p$-value of the $\chi_{F}^{2}$ distribution. If the null hypothesis can be rejected we proceed with the Nemenyi test \cite{Nemenyi63} as post hoc test in order to show which classifier performs significantly different. Here, the performance differences of two classifiers are significant if the corresponding average ranks differ by at least the critical difference $\text{CD} = q_{\alpha} \sqrt{\sfrac{S(S+1)}{6N}}$ where the critical value $q_{\alpha}$ is based on the Studentized range statistic divided by $\sqrt{2}$. Dem\v{s}ar \cite{Demsar06} suggests that the results of the Nemenyi test can be visualized with help of critical difference plots. In these plots, non-significantly different classifiers are connected in groups (their rank difference is smaller than $\text{CD}$). To summarize the classification results over all data sets, the average ranks and the numbers of wins are shown. A \textit{number of wins} outlines the number of data sets for which a paradigm performs best. Wins can be ``shared'' when different classifiers perform equally on the same data set. That is, a good paradigm yields a low average rank and a large number of wins.

\subsubsection{Results}
\label{subsec:results}
We compare the classification performance achieved by an SVM with RWM kernel to that of an SVM with GMM, LAP, and RBF kernels. The evaluation criterion for our comparison of the four kernel functions is the classification accuracy on the test set $T$, the data set never used for any modeling or other parametrization purposes (averaged over five folds of the cross-validation). In each experiment we used significance values $\alpha$ of $0.01$, $0.05$, and $0.1$ and present the lowest value (if any) of $\alpha$ for which the significant difference of at least one of kernel functions to the other kernels can be stated.

A general observation, which is holds for all kernel functions, is the higher the fraction of labeled samples the higher the number of samples that are used as support vectors. However, this correlation is not linear, because the number of support vectors highly depends on the difficulty to classify a considered data set correctly (e.g., compare data sets Two Moons and Yeast in Tables~\ref{tab:ResultsFractionMin}, \ref{tab:ResultsFraction01} and \ref{tab:ResultsFraction10}).

\begin{figure*}[htbp!]
\centering
\subfigure[Experiment 1 with $\alpha = 0.01$.\label{fig:criticalDifferencePlotsMin}]{\includegraphics[width=0.48\textwidth]{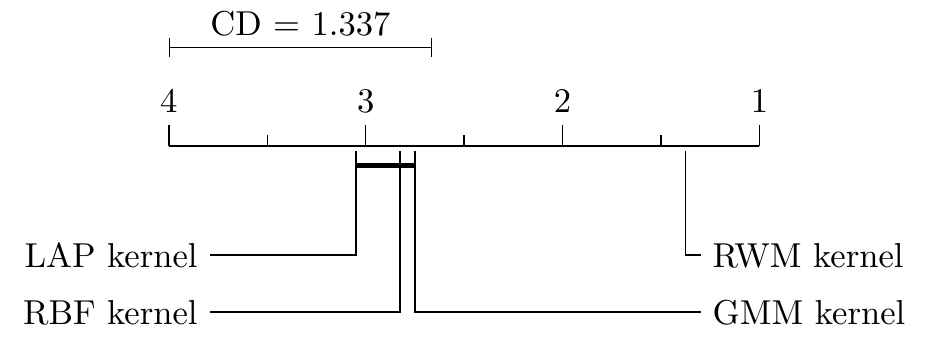}}\hfill
\subfigure[Experiment 2 with $\alpha = 0.1$.\label{fig:criticalDifferencePlots01}]{\includegraphics[width=0.48\textwidth]{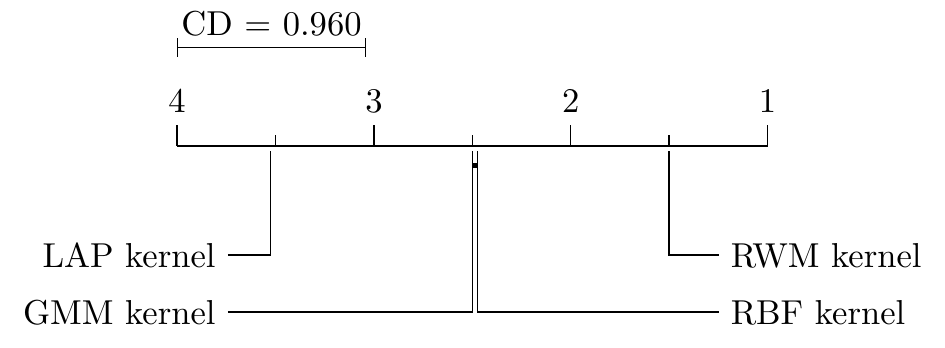}}\vfill
\subfigure[Experiment 3 with $\alpha = 0.1$.\label{fig:criticalDifferencePlots10}]{\includegraphics[width=0.48\textwidth]{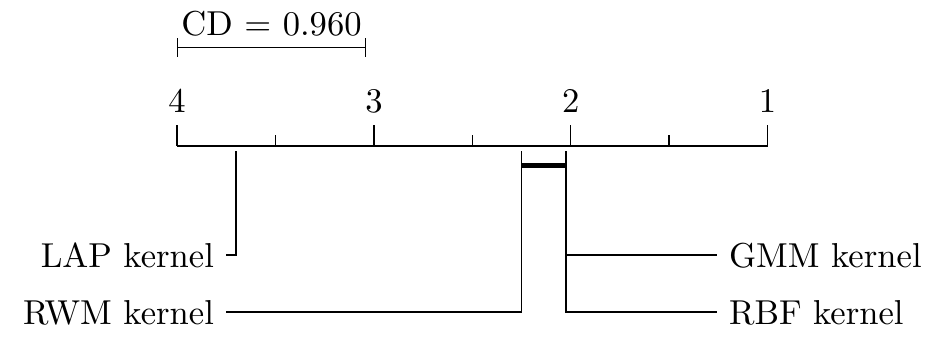}}
\caption{Comparison of SVM combined with different kernel functions with the Nemenyi test for different fractions of labeled samples (experiments 1, 2, and 3). Groups of SVM that are not significantly different are connected.}
\label{fig:criticalDifferencePlots}
\end{figure*}

In experiment 1 we limited the number of labeled samples to $4 \times$ the number of classes for each data set. Table~\ref{tab:ResultsFractionMin} shows the classification accuracies for an SVM combined with each kernel function on the 20 data sets. The best results (classifiers that received the smallest ranks according to the Friedman test) for each data set are highlighted in bold face. With four classifiers and 20 data sets, Friedman's $\chi_{F}^{2}$ is distributed according to a $\chi_{F}^{2}$ distribution with $4 - 1$ degrees of freedom. The critical value of $\chi_{F}^{2}(3)$ for $\alpha = 0.01$ is $11.1$ and, thus, smaller than Friedman's $\chi_{F}^{2} = 20.83$, so we can reject the null hypothesis. With the Nemenyi test, we compute the critical difference $\text{CD}= 3.275 \sqrt{\sfrac{4 \cdot 5}{6 \cdot 20}} = 1.337$ to investigate which methods perform significantly different. The corresponding critical difference plot is shown in Fig.~\ref{fig:criticalDifferencePlotsMin}. On a significance level of $\alpha=0.01$, an SVM with RWM kernel performs significantly better than an SVM combined with GMM, LAP, or RBF kernels, that build a group of not significantly different classifiers. The superior performance of the RWM kernel is also visible in the last two rows of Table~\ref{tab:ResultsFractionMin}. There, we can notice that an SVM with RWM kernel wins more than 14 of the 20 data sets and yields the smallest average rank. Despite the significantly better performance of the RWM kernel, an SVM combined with this kernel does not require more support vectors than with other kernel functions. Thus, with regard to the needed number of support vectors no significant difference is visible (cf.~columns 4, 6, 8, and 10 of Table~\ref{tab:ResultsFractionMin}).
\begin{table*}[htbp!]%
\caption{Classification accuracies on the test data (with standard deviations), average ranks, and wins for each data set for SVM combined with RWM, GMM, RBF, and LAP kernels. The training set size (i.e., the number of labeled samples, cf.~second column) is $4 \times \text{the number of classes}$. The columns 4, 6, 8, and 10 show the number of support vectors used by an SVM with one of the considered kernel functions.}
\label{tab:ResultsFractionMin}
\begin{center}
\ifpreprint
\tiny
\begin{tabular}{|l|r|p{0.9cm}|r|p{0.9cm}|r|p{0.9cm}|r|p{0.9cm}|r|} 
\hline
Data set & \multicolumn{1}{c|}{$4\times|C|$} & RWM kernel & \multicolumn{1}{c|}{\#SV} & GMM kernel & \multicolumn{1}{c|}{\#SV} & RBF kernel & \multicolumn{1}{c|}{\#SV} & LAP kernel & \multicolumn{1}{c|}{\#SV} \\
\else
\scriptsize
\begin{tabular}{|l|c|c|c|c|c|c|c|c|c|}
\hline
Data set & \multicolumn{1}{c|}{$4\times|C|$} & RWM kernel & \multicolumn{1}{c|}{\#SV} & GMM kernel & \multicolumn{1}{c|}{\#SV} & RBF kernel & \multicolumn{1}{c|}{\#SV} & LAP kernel & \multicolumn{1}{c|}{\#SV} \\
\fi

\hline
\hline
Australian & 8 & $\mathbf{82.46}$ $ \mathbf{(\pm 5.715)}$ & $7.8$ & $79.86$ $ (\pm 10.253)$ & $7.6$ & $79.86$ $ (\pm 10.253)$ & $7.6$ & $81.74$ $ (\pm 5.940)$ & $8.0$ \\
Clouds & 8 & $\mathbf{83.72}$ $ \mathbf{(\pm 2.722)}$ & $6.6$ & $66.48$ $ (\pm 8.551)$ & $5.8$ & $71.50$ $ (\pm 5.075)$ & $7.4$ & $74.88$ $ (\pm 4.612)$ & $8.0$ \\
Concentric & 8 & $\mathbf{87.92}$ $ \mathbf{(\pm 2.995)}$ & $7.8$ & $78.64$ $ (\pm 4.727)$ & $7.8$ & $84.28$ $ (\pm 6.225)$ & $7.6$ & $86.52$ $ (\pm 7.260)$ & $6.4$ \\
Credit A & 8 & $\mathbf{76.09}$ $ \mathbf{(\pm 6.317)}$ & $6.6$ & $73.77$ $ (\pm 5.210)$ & $7.4$ & $75.51$ $ (\pm 8.205)$ & $8.0$ & $70.43$ $ (\pm 5.552)$ & $7.6$ \\
Credit G & 8 & $65.90$ $ (\pm 3.975)$ & $7.4$ & $65.90$ $ (\pm 3.975)$ & $7.4$ & $65.90$ $ (\pm 3.975)$ & $7.4$ & $\mathbf{69.30}$ $ \mathbf{(\pm 1.891)}$ & $8.0$ \\
Ecoli & 32 & $75.31$ $ (\pm 1.337)$ & $29.4$ & $72.92$ $ (\pm 1.690)$ & $27.6$ & $\mathbf{75.97}$ $ \mathbf{(\pm 5.969)}$ & $27.4$ & $71.45$ $ (\pm 4.200)$ & $26.2$ \\
Glass & 24 & $\mathbf{50.94}$ $ \mathbf{(\pm 4.538)}$ & $22.8$ & $47.65$ $ (\pm 4.665)$ & $23.8$ & $48.59$ $ (\pm 4.065)$ & $23.2$ & $36.46$ $ (\pm 2.213)$ & $24.0$ \\
Heart & 8 & $\mathbf{82.96}$ $ \mathbf{(\pm 4.223)}$ & $8.0$ & $81.85$ $ (\pm 4.223)$ & $7.4$ & $81.85$ $ (\pm 3.562)$ & $6.4$ & $82.59$ $ (\pm 3.840)$ & $7.0$ \\
Iris & 12 & $92.00$ $ (\pm 6.055)$ & $9.4$ & $\mathbf{92.67}$ $ \mathbf{(\pm 4.944)}$ & $8.8$ & $89.33$ $ (\pm 7.601)$ & $7.2$ & $90.67$ $ (\pm 4.346)$ & $8.0$ \\
Page Blocks & 20 & $\mathbf{92.36}$ $ \mathbf{(\pm 0.929)}$ & $12.4$ & $89.77$ $ (\pm 0.077)$ & $20.0$ & $90.63$ $ (\pm 1.070)$ & $12.2$ & $89.93$ $ (\pm 0.241)$ & $15.0$ \\
Phoneme & 8 & $\mathbf{72.41}$ $ \mathbf{(\pm 0.750)}$ & $7.0$ & $72.30$ $ (\pm 1.811)$ & $7.0$ & $71.37$ $ (\pm 1.001)$ & $7.4$ & $70.63$ $ (\pm 0.156)$ & $8.0$ \\
Pima & 8 & $\mathbf{69.14}$ $ \mathbf{(\pm 4.469)}$ & $7.2$ & $66.66$ $ (\pm 1.987)$ & $7.4$ & $66.66$ $ (\pm 3.676)$ & $6.8$ & $68.87$ $ (\pm 3.138)$ & $6.8$ \\
Ripley & 8 & $\mathbf{90.16}$ $ \mathbf{(\pm 1.345)}$ & $7.6$ & $87.60$ $ (\pm 1.356)$ & $4.8$ & $87.68$ $ (\pm 2.876)$ & $7.2$ & $86.48$ $ (\pm 2.748)$ & $4.0$ \\
Satimage & 24 & $79.16$ $ (\pm 2.293)$ & $18.0$ & $\mathbf{80.37}$ $ \mathbf{(\pm 1.956)}$ & $18.2$ & $74.79$ $ (\pm 4.356)$ & $21.6$ & $61.58$ $ (\pm 6.293)$ & $21.8$ \\
Seeds & 12 & $\mathbf{93.33}$ $ \mathbf{(\pm 3.104)}$ & $7.6$ & $\mathbf{93.33}$ $ \mathbf{(\pm 4.880)}$ & $7.4$ & $90.48$ $ (\pm 4.124)$ & $8.2$ & $88.57$ $ (\pm 4.580)$ & $12.0$ \\
Two Moons & 8 & $\mathbf{99.12}$ $ \mathbf{(\pm 1.957)}$ & $5.4$ & $93.38$ $ (\pm 3.992)$ & $6.6$ & $92.12$ $ (\pm 2.054)$ & $7.2$ & $98.25$ $ (\pm 1.355)$ & $6.0$ \\
Vehicle & 16 & $\mathbf{49.64}$ $ \mathbf{(\pm 3.233)}$ & $15.8$ & $44.66$ $ (\pm 17.545)$ & $16.0$ & $43.96$ $ (\pm 9.547)$ & $15.6$ & $43.85$ $ (\pm 3.870)$ & $15.8$ \\
Vowel & 44 & $\mathbf{50.51}$ $ \mathbf{(\pm 2.550)}$ & $44.0$ & $43.64$ $ (\pm 7.488)$ & $44.0$ & $41.72$ $ (\pm 7.530)$ & $44.0$ & $37.98$ $ (\pm 4.321)$ & $44.0$ \\
Wine & 12 & $\mathbf{96.05}$ $ \mathbf{(\pm 2.539)}$ & $11.8$ & $93.81$ $ (\pm 3.654)$ & $11.4$ & $92.13$ $ (\pm 5.378)$ & $10.8$ & $95.48$ $ (\pm 4.331)$ & $12.0$ \\
Yeast & 40 & $42.26$ $ (\pm 2.431)$ & $39.0$ & $46.50$ $ (\pm 2.536)$ & $37.6$ & $\mathbf{47.10}$ $ \mathbf{(\pm 2.965)}$ & $36.6$ & $36.45$ $ (\pm 4.207)$ & $39.0$ \\
\hline
\hline
Mean & 15.8 & $76.57$ $(\pm 3.556)$ & $14.1$ & $73.59$ $(\pm 6.110)$ & $14.2$ & $73.57$ $(\pm 5.590)$ & $14.0$ & $72.11$ $(\pm 4.182)$ & $14.4$ \\
\hline
\hline
Rank & &$1.375$ &  & $2.750$ &  & $2.825$ &  & $3.050$ &  \\
Win & &$14.5$ &  & $2.5$ &  & $2.0$ &  & $1.0$ &  \\
\hline
\end{tabular}
\end{center}
\end{table*}

In experiment 2, we increased the number of labeled samples to $10\%$ of $|L|$ for each data set. The corresponding classification results are summarized in Table~\ref{tab:ResultsFraction01}.
For a significance value $\alpha=0.1$, the critical value of $\chi_{F}^{2}(3)$ is $6.24$ and thus smaller than Friedman's $\chi_{F}^{2} = 24.62$, so we can also reject the null hypothesis. The Nemenyi test with critical difference $\text{CD}= 2.351 \sqrt{\sfrac{4 \cdot 5}{6 \cdot 20}} = 0.960$ shows that an SVM with RWM kernel performs significantly better than an SVM combined with one of the other three kernels and, consequently, it confirms the results obtained in the first experiment. The respective CD plot is shown in Fig.~\ref{fig:criticalDifferencePlots01}. It shows that an SVM with GMM or RBF kernels performs significantly better than an SVM with LAP kernel. However, for these two kernel functions, no significant difference is observed. The last two rows of Table~\ref{tab:ResultsFraction01} show again that an SVM with RWM kernel performs better than an SVM in combination with one of the other kernels on more than 13 data sets (wins) and also performs best on average (rank). Note that the significance level $\alpha$ is $0.1$ in contrast to experiment 1 (there: $0.01$). That is, no significant advantage of RWM kernels was stated for $\alpha=0.01$ and $\alpha=0.05$. Despite the significantly better performance of the RWM kernel it requires slightly more support vectors than the RBF and GMM kernels, but less support vectors than the LAP kernel for more than a half of the data sets (cf.~columns 4, 6, 8, and 10 of Table~\ref{tab:ResultsFraction01}).
\begin{table*}[htbp!]%
\caption{Classification accuracies on the test data, average ranks, and wins for each data set for SVM combined with RWM, GMM, RBF, and LAP kernel. The training set size (i.e., the number of labeled samples, cf.~second column) is $10\%$ of the number $|L|$. The columns 4, 6, 8, and 10 show the number of support vectors used by an SVM with one of the considered kernel functions.}
\label{tab:ResultsFraction01}
\begin{center}
\ifpreprint
\tiny
\begin{tabular}{|l|r|p{0.9cm}|r|p{0.9cm}|r|p{0.9cm}|r|p{0.9cm}|r|} 
\hline
Data set & \multicolumn{1}{c|}{$|L|/10$} & RWM kernel & \multicolumn{1}{c|}{\#SV} & GMM kernel & \multicolumn{1}{c|}{\#SV} & RBF kernel & \multicolumn{1}{c|}{\#SV} & LAP kernel & \multicolumn{1}{c|}{\#SV} \\
\else
\scriptsize
\begin{tabular}{|l|c|c|c|c|c|c|c|c|c|}
\hline
Data set & \multicolumn{1}{c|}{$|L|/10$} & RWM kernel & \multicolumn{1}{c|}{\#SV} & GMM kernel & \multicolumn{1}{c|}{\#SV} & RBF kernel & \multicolumn{1}{c|}{\#SV} & LAP kernel & \multicolumn{1}{c|}{\#SV} \\
\fi
\hline
\hline
Australian & 56 & $\mathbf{86.23}$ $ \mathbf{(\pm 3.321)}$ & $42.0$ & $\mathbf{86.23}$ $ \mathbf{(\pm 3.321)}$ & $42.0$ & $\mathbf{86.23}$ $ \mathbf{(\pm 3.321)}$ & $42.0$ & $\mathbf{86.17}$ $ \mathbf{(\pm 2.079)}$ & $45.4$ \\
Clouds & 400 & $\mathbf{88.38}$ $ \mathbf{(\pm 1.016)}$ & $118.4$ & $86.62$ $ (\pm 1.117)$ & $113.4$ & $87.24$ $ (\pm 1.092)$ & $115.8$ & $74.30$ $ (\pm 1.208)$ & $361.6$ \\
Concentric & 200 & $\mathbf{97.12}$ $ \mathbf{(\pm 0.415)}$ & $54.6$ & $96.72$ $ (\pm 0.756)$ & $37.2$ & $97.00$ $ (\pm 1.183)$ & $36.2$ & $95.52$ $ (\pm 0.576)$ & $28.2$ \\
Credit A & 56 & $\mathbf{85.51}$ $ \mathbf{(\pm 2.233)}$ & $37.8$ & $84.64$ $ (\pm 4.449)$ & $42.8$ & $81.74$ $ (\pm 2.633)$ & $33.2$ & $83.19$ $ (\pm 2.582)$ & $33.2$ \\
Credit G & 80 & $71.10$ $ (\pm 2.043)$ & $59.2$ & $70.20$ $ (\pm 2.253)$ & $57.6$ & $\mathbf{72.60}$ $ \mathbf{(\pm 2.408)}$ & $63.8$ & $70.90$ $ (\pm 0.894)$ & $74.0$ \\
Ecoli & 32 & $\mathbf{79.14}$ $ \mathbf{(\pm 4.495)}$ & $30.0$ & $77.39$ $ (\pm 5.162)$ & $28.4$ & $78.30$ $ (\pm 3.544)$ & $29.4$ & $71.37$ $ (\pm 6.365)$ & $29.2$ \\
Glass & 24 & $\mathbf{52.80}$ $ \mathbf{(\pm 4.159)}$ & $23.4$ & $47.65$ $ (\pm 4.665)$ & $23.8$ & $48.59$ $ (\pm 4.065)$ & $23.2$ & $43.93$ $ (\pm 7.638)$ & $24.0$ \\
Heart & 22 & $\mathbf{82.22}$ $ \mathbf{(\pm 4.829)}$ & $15.0$ & $80.37$ $ (\pm 4.263)$ & $13.2$ & $81.48$ $ (\pm 4.900)$ & $16.4$ & $80.37$ $ (\pm 2.111)$ & $18.8$ \\
Iris & 12 & $\mathbf{93.33}$ $ \mathbf{(\pm 2.357)}$ & $9.0$ & $92.67$ $ (\pm 4.944)$ & $8.8$ & $89.33$ $ (\pm 7.601)$ & $7.2$ & $89.33$ $ (\pm 1.491)$ & $9.6$ \\
Page Blocks & 438 & $95.23$ $ (\pm 0.335)$ & $216.8$ & $95.18$ $ (\pm 0.644)$ & $176.8$ & $\mathbf{95.71}$ $ \mathbf{(\pm 0.344)}$ & $152.6$ & $94.32$ $ (\pm 0.273)$ & $169.8$ \\
Phoneme & 433 & $\mathbf{80.31}$ $ \mathbf{(\pm 0.367)}$ & $189.2$ & $78.48$ $ (\pm 1.134)$ & $279.2$ & $79.50$ $ (\pm 2.407)$ & $225.8$ & $73.63$ $ (\pm 4.100)$ & $295.2$ \\
Pima & 62 & $\mathbf{69.91}$ $ \mathbf{(\pm 4.191)}$ & $52.8$ & $64.06$ $ (\pm 3.672)$ & $52.0$ & $65.51$ $ (\pm 7.634)$ & $50.4$ & $67.32$ $ (\pm 4.202)$ & $45.4$ \\
Ripley & 100 & $\mathbf{90.00}$ $ \mathbf{(\pm 1.442)}$ & $30.4$ & $89.52$ $ (\pm 1.635)$ & $33.2$ & $89.04$ $ (\pm 2.032)$ & $47.8$ & $86.40$ $ (\pm 6.505)$ & $54.0$ \\
Satimage & 515 & $\mathbf{86.51}$ $ \mathbf{(\pm 0.490)}$ & $263.8$ & $86.25$ $ (\pm 0.935)$ & $215.2$ & $85.24$ $ (\pm 0.927)$ & $197.6$ & $80.17$ $ (\pm 1.583)$ & $408.6$ \\
Seeds & 17 & $\mathbf{93.33}$ $ \mathbf{(\pm 3.104)}$ & $8.0$ & $\mathbf{93.33}$ $ \mathbf{(\pm 3.912)}$ & $10.0$ & $89.05$ $ (\pm 4.325)$ & $9.0$ & $85.71$ $ (\pm 4.762)$ & $13.2$ \\
Two Moons & 64 & $\mathbf{100.00}$ $ \mathbf{(\pm 0.000)}$ & $51.8$ & $99.38$ $ (\pm 0.884)$ & $32.4$ & $99.00$ $ (\pm 1.630)$ & $29.8$ & $\mathbf{100.00}$ $ \mathbf{(\pm 0.000)}$ & $13.6$ \\
Vehicle & 68 & $66.28$ $ (\pm 2.746)$ & $62.8$ & $\mathbf{69.26}$ $ \mathbf{(\pm 5.158)}$ & $48.6$ & $61.45$ $ (\pm 7.432)$ & $59.0$ & $56.84$ $ (\pm 6.305)$ & $53.6$ \\
Vowel & 80 & $64.75$ $ (\pm 1.532)$ & $79.0$ & $\mathbf{65.86}$ $ \mathbf{(\pm 3.304)}$ & $77.6$ & $64.95$ $ (\pm 2.914)$ & $77.6$ & $40.00$ $ (\pm 3.983)$ & $79.8$ \\
Wine & 15 & $\mathbf{95.49}$ $ \mathbf{(\pm 1.581)}$ & $13.6$ & $94.94$ $ (\pm 1.280)$ & $14.0$ & $92.67$ $ (\pm 3.287)$ & $13.0$ & $92.63$ $ (\pm 6.282)$ & $15.0$ \\
Yeast & 119 & $51.96$ $ (\pm 1.205)$ & $113.6$ & $52.29$ $ (\pm 0.737)$ & $106.2$ & $\mathbf{54.38}$ $ \mathbf{(\pm 1.755)}$ & $101.6$ & $36.73$ $ (\pm 4.299)$ & $111.2$ \\
\hline
\hline
Mean & 139.7 & $81.48$ $(\pm 2.563)$ & $73.6$ & $80.55$ $(\pm 3.188)$ & $70.6$ & $79.95$ $(\pm 3.907)$ & $66.6$ & $75.44$ $(\pm 4.091)$ & $94.2$ \\
\hline
\hline
Rank & &$1.500$ &  & $2.500$ &  & $2.475$ &  & $3.525$ &  \\
Win & &$13.2$ &  & $2.8$ &  & $3.2$ &  & $0.8$ &  \\
\hline

\end{tabular}
\end{center}
\end{table*}

In experiment 3, we used all samples in $L$ as labeled training samples for each data set (i.e., we train the SVM completely supervised and not semi-supervised as above). Table~\ref{tab:ResultsFraction10} summarizes the classification accuracies of SVM with RWM, GMM, LAP, and RBF kernels on all 20 data sets. Here, with $\alpha=0.1$ the null hypothesis is rejected again (the critical value of $\chi_{F}^{2}(3)$ is $6.24$ which is smaller than Friedman's $\chi_{F}^{2} = 23.44$). The Nemenyi test with $\text{CD} = 0.960$ shows that an SVM with RWM kernel belongs to the ``top'' group (a group of not significantly different, but best performing classifiers) together with an SVM combined with GMM or RBF kernel, cf.\ Fig.~\ref{fig:criticalDifferencePlots10}. With a closer look at Table~\ref{tab:ResultsFraction10}, we see that an SVM with GMM or RBF kernels performs best regarding the highest number of wins ($6.8$) and the smallest average ranks ($2.025$). In comparison, the SVM with RWM kernel yields a win of $5.8$ and an average rank of $2.250$. Altogether we can state that, the results of SVM with RWM, GMM, and RBF kernels are not significantly different for $\alpha=0.01$, $\alpha=0.05$, and $\alpha=0.1$. Experiment 3 shows that considering structure information resulting from a parametric or non-parametric density estimation brings no further benefit if the data set is completely labeled. In addition, the columns 4, 6, 8, and 10 of Table~\ref{tab:ResultsFraction10} show that an SVM combined with a kernel function that uses structure information requires a higher number of support vectors than with an RBF kernel that neglects this information if all samples are labeled.

Experiments 1 to 3 have shown that in the presence of sparse\-ly labeled data the use of an RWM kernel may lead to significantly better results compared to the  three kernels LAP, GMM, and RBF. Thus, the RWM kernel seems to be perfectly suited for SSL.


\begin{table*}[htbp!]%
\caption{Classification accuracies on the test data, average ranks, and wins for each data set for SVM combined with RWM, GMM, RBF, and LAP kernel. The training set size (i.e., the number of labeled samples, cf.~second column) is $|L|$. The columns 4, 6, 8, and 10 show the number of support vectors used by an SVM with one of the considered kernel functions.}
\label{tab:ResultsFraction10}
\begin{center}
\ifpreprint
\tiny
\begin{tabular}{|l|r|p{0.8cm}|r|p{0.8cm}|r|p{0.8cm}|r|p{0.8cm}|r|} 
\hline
Data set & \multicolumn{1}{c|}{$|L|$} & RWM kernel & \multicolumn{1}{c|}{\#SV} & GMM kernel & \multicolumn{1}{c|}{\#SV} & RBF kernel & \multicolumn{1}{c|}{\#SV} & LAP kernel & \multicolumn{1}{c|}{\#SV} \\
\else
\scriptsize
\begin{tabular}{|l|c|c|c|c|c|c|c|c|c|}
\hline
Data set & \multicolumn{1}{c|}{$|L|$} & RWM kernel & \multicolumn{1}{c|}{\#SV} & GMM kernel & \multicolumn{1}{c|}{\#SV} & RBF kernel & \multicolumn{1}{c|}{\#SV} & LAP kernel & \multicolumn{1}{c|}{\#SV} \\
\fi
\hline 
\hline
Australian & 552 & $86.03$ $ (\pm 1.451)$ & $406.0$ & $\mathbf{86.16}$ $ \mathbf{(\pm 1.987)}$ & $332.6$ & $\mathbf{86.23}$ $ \mathbf{(\pm 2.613)}$ & $374.8$ & $\mathbf{86.22}$ $ \mathbf{(\pm 1.847)}$ & $276.6$ \\
Clouds & 4000 & $89.34$ $ (\pm 0.550)$ & $1181.2$ & $89.40$ $ (\pm 0.809)$ & $1015.8$ & $\mathbf{89.56}$ $ \mathbf{(\pm 0.844)}$ & $1027.2$ & $88.84$ $ (\pm 1.234)$ & $964.2$ \\
Concentric & 2000 & $\mathbf{99.48}$ $ \mathbf{(\pm 0.335)}$ & $69.0$ & $\mathbf{99.48}$ $ \mathbf{(\pm 0.363)}$ & $158.0$ & $\mathbf{99.56}$ $ \mathbf{(\pm 0.167)}$ & $113.8$ & $\mathbf{99.52}$ $ \mathbf{(\pm 0.228)}$ & $69.0$ \\
Credit A & 552 & $\mathbf{85.07}$ $ \mathbf{(\pm 1.099)}$ & $248.2$ & $84.35$ $ (\pm 1.819)$ & $307.2$ & $\mathbf{85.07}$ $ \mathbf{(\pm 1.668)}$ & $365.4$ & $82.17$ $ (\pm 6.886)$ & $339.6$ \\
Credit G & 800 & $75.20$ $ (\pm 2.515)$ & $522.8$ & $73.40$ $ (\pm 1.387)$ & $569.8$ & $\mathbf{76.00}$ $ \mathbf{(\pm 2.622)}$ & $564.4$ & $73.30$ $ (\pm 4.522)$ & $568.6$ \\
Ecoli & 270 & $85.72$ $ (\pm 4.621)$ & $140.0$ & $86.34$ $ (\pm 4.706)$ & $132.6$ & $\mathbf{86.96}$ $ \mathbf{(\pm 3.914)}$ & $129.8$ & $69.91$ $ (\pm 4.395)$ & $133.6$ \\
Glass & 171 & $67.76$ $ (\pm 1.794)$ & $143.4$ & $67.73$ $ (\pm 4.736)$ & $133.8$ & $\mathbf{69.17}$ $ \mathbf{(\pm 2.891)}$ & $127.6$ & $63.53$ $ (\pm 4.422)$ & $154.6$ \\
Heart & 216 & $\mathbf{85.19}$ $ \mathbf{(\pm 3.928)}$ & $115.2$ & $83.33$ $ (\pm 1.309)$ & $104.4$ & $82.59$ $ (\pm 3.364)$ & $139.6$ & $82.59$ $ (\pm 5.172)$ & $133.0$ \\
Iris & 120 & $97.33$ $ (\pm 1.491)$ & $66.6$ & $\mathbf{98.00}$ $ \mathbf{(\pm 1.826)}$ & $44.8$ & $95.33$ $ (\pm 1.826)$ & $33.2$ & $89.33$ $ (\pm 2.789)$ & $94.4$ \\
Page Blocks & 4377 & $95.94$ $ (\pm 0.556)$ & $596.6$ & $96.18$ $ (\pm 0.620)$ & $420.2$ & $\mathbf{96.73}$ $ \mathbf{(\pm 0.754)}$ & $385.0$ & $93.68$ $ (\pm 1.172)$ & $610.8$ \\
Phoneme & 4323 & $\mathbf{89.86}$ $ \mathbf{(\pm 0.746)}$ & $2376.0$ & $88.90$ $ (\pm 0.794)$ & $1432.2$ & $89.25$ $ (\pm 1.112)$ & $1179.2$ & $80.68$ $ (\pm 5.177)$ & $2174.6$ \\
Pima & 614 & $73.96$ $ (\pm 3.289)$ & $399.6$ & $\mathbf{76.82}$ $ \mathbf{(\pm 3.297)}$ & $328.6$ & $76.56$ $ (\pm 2.343)$ & $354.4$ & $73.18$ $ (\pm 5.220)$ & $373.2$ \\
Ripley & 1000 & $\mathbf{90.80}$ $ \mathbf{(\pm 1.766)}$ & $417.0$ & $90.56$ $ (\pm 1.615)$ & $483.0$ & $90.64$ $ (\pm 1.565)$ & $230.4$ & $89.68$ $ (\pm 1.968)$ & $719.2$ \\
Satimage & 5147 & $88.83$ $ (\pm 0.243)$ & $1648.0$ & $\mathbf{89.57}$ $ \mathbf{(\pm 0.588)}$ & $1599.6$ & $88.72$ $ (\pm 0.590)$ & $1386.8$ & $76.55$ $ (\pm 2.457)$ & $1868.6$ \\
Seeds & 168 & $\mathbf{96.19}$ $ \mathbf{(\pm 2.715)}$ & $32.6$ & $\mathbf{96.19}$ $ \mathbf{(\pm 2.715)}$ & $79.2$ & $93.81$ $ (\pm 2.130)$ & $35.4$ & $91.43$ $ (\pm 5.216)$ & $122.2$ \\
Two Moons & 640 & $\mathbf{100.00}$ $ \mathbf{(\pm 0.000)}$ & $13.0$ & $\mathbf{100.00}$ $ \mathbf{(\pm 0.000)}$ & $15.4$ & $\mathbf{100.00}$ $ \mathbf{(\pm 0.000)}$ & $15.4$ & $\mathbf{100.00}$ $ \mathbf{(\pm 0.000)}$ & $12.8$ \\
Vehicle & 677 & $79.90$ $ (\pm 4.971)$ & $629.6$ & $\mathbf{84.99}$ $ \mathbf{(\pm 2.615)}$ & $283.6$ & $83.57$ $ (\pm 2.117)$ & $331.8$ & $63.35$ $ (\pm 4.574)$ & $666.4$ \\
Vowel & 792 & $97.07$ $ (\pm 1.152)$ & $580.6$ & $\mathbf{98.99}$ $ \mathbf{(\pm 0.357)}$ & $579.2$ & $98.69$ $ (\pm 0.576)$ & $535.2$ & $88.59$ $ (\pm 12.973)$ & $764.6$ \\
Wine & 142 & $\mathbf{99.44}$ $ \mathbf{(\pm 1.242)}$ & $68.4$ & $98.32$ $ (\pm 1.536)$ & $60.4$ & $98.33$ $ (\pm 1.521)$ & $74.2$ & $97.75$ $ (\pm 3.087)$ & $126.4$ \\
Yeast & 1186 & $58.62$ $ (\pm 2.121)$ & $993.0$ & $\mathbf{59.03}$ $ \mathbf{(\pm 2.271)}$ & $969.8$ & $\mathbf{58.96}$ $ \mathbf{(\pm 2.215)}$ & $935.4$ & $42.99$ $ (\pm 4.427)$ & $1133.0$ \\
\hline
\hline
Mean & 1387.3 & $87.09$ $(\pm 2.309)$ & $532.3$ & $87.39$ $(\pm 2.194)$ & $452.5$ & $87.29$ $(\pm 2.027)$ & $416.9$ & $81.66$ $(\pm 4.781)$ & $565.3$ \\
\hline
\hline
Rank & &$2.250$ &  & $2.025$ &  & $2.025$ &  & $3.700$ &  \\
Win & &$5.5$ &  & $6.8$ &  & $6.8$ &  & $0.8$ &  \\
\hline

\end{tabular}
\end{center}
\end{table*}

To evaluate the run-time of an SVM with RWM kernel we conducted run-time measurements on an Intel Xeon Processor E5-2670 v2. Here, we measure the run-time averaged over a five-fold cross-validation for three tasks: (1) constructing the kernel matrix (building time), (2) training the SVM with RWM, GMM, RBF, and LAP kernels by solving the optimization problem with the SMO algorithm (training time), and (3) testing the unlabeled samples with trained SVM (testing time). For the RWM and GMM kernels we measure the run-time for the model estimation, too. All run-times are summarized in Table~\ref{tab:times}. The sizes of the different sample sets that are used to train the SVM ($|L|/10$), to test the SVM ($|U|$) and to estimate the (parametric or non-parametric) density model in case of the RWM, GMM, and LAP kernel ($|L|$) correspond to the sample set sizes used in experiment 2 above. Across all kernels, the training and testing times are very similar. Interestingly, the LAP kernel yields the smallest testing time. This is due to the fact that the LapSVM use an different evaluation function (in comparison to the \textsf{libsvm}). The main difference can be seen if we compare the building times. For a fair comparison of the RWM and GMM kernels to the other two kernels we have to extend their building times by the model estimation times. However, averaged over all data sets the sum of these two times is much lower than the kernel building time of the LapSVM. Moreover, the model must be estimated only once (i.e.,\ offline, before the SVM is trained). Averaged over all data sets, we see that the SVM with RBF kernel yield the smallest run-times, but an SVM with the new RWM kernel takes comparable times for training and testing, and five times longer for building the kernel matrix if we assume that the density model is available.    

\begin{table*}[htbp!]
\caption{Run-times in seconds averaged over a five fold cross-validation to construct the kernel matrix (building), to solve the optimization problem with SMO based on $|L|/10$ labeled samples (training) and to classify $|U|$ unlabeled samples  (testing) by means of an SVM with RWM, GMM, RBF and LAP kernels. Column 2 gives the additional run-times for estimating the density model based on $|L|$ samples that is used by the RWM and GMM kernels.}
\label{tab:times}
\begin{center}
\setlength{\tabcolsep}{3.5pt}
\scriptsize
\begin{tabular}{|l |c|c|r |r r r| r r r| r r r |r r r|}\hline
\multirow{3}{*}{Data set} & \multirow{3}{*}{$|L|/10$}& \multirow{3}{*}{$|U|$}& \multicolumn{13}{c|}{run-time in $s$ for}\\
& & &\multicolumn{1}{c|}{model} & \multicolumn{3}{c|}{RWM kernel}& \multicolumn{3}{c|}{GMM kernel}& \multicolumn{3}{c|}{RBF kernel}& \multicolumn{3}{c|}{LAP kernel}\\
& & &\multicolumn{1}{c|}{estimation} & building & training& testing & building & training& testing & building & training& testing & building & training& testing \\
\hline
\hline
Australian &56  &140    & 2.699& 0.668& 0.009& 0.686& 0.695& 0.009& 0.725& 0.278& 0.010& 0.619& 2.256& 0.005& 0.128\\
Clouds     &400 &1000   & 24.264& 2.501& 0.035& 4.875& 2.877& 0.057& 5.745& 1.665& 0.043& 5.259& 420.871& 0.074& 1.798\\
Concentric &200 &500    & 2.783& 1.755& 0.019& 2.242& 1.740& 0.014& 2.289& 0.679& 0.018& 2.444& 47.650& 0.023& 0.462\\
Credit A   &56  &140    & 4.212& 0.613& 0.010& 0.738& 0.720& 0.007& 0.668& 0.275& 0.010& 0.592& 2.357& 0.007& 0.121\\
Credit G   &80  &200    & 18.537& 1.653& 0.008& 0.694& 1.777& 0.010& 0.703& 0.322& 0.008& 0.630& 5.046& 0.007& 0.260\\
Ecoli      &32  &80     & 1.098& 0.218& 0.016& 0.259& 0.226& 0.010& 0.244& 0.063& 0.013& 0.292& 0.474& 0.017& 0.051\\
Glass      &24  &60     & 0.769& 0.163& 0.008& 0.143& 0.126& 0.009& 0.144& 0.048& 0.010& 0.156& 0.218& 0.016& 0.038\\
Heart      &22  &55     & 0.762& 0.122& 0.005& 0.274& 0.115& 0.008& 0.283& 0.084& 0.006& 0.288& 0.216& 0.003& 0.039\\
Iris       &12  &33     & 0.278& 0.076& 0.006& 0.125& 0.067& 0.005& 0.139& 0.035& 0.007& 0.170& 0.118& 0.005& 0.027\\
Page Blocks&438 &1095   & 37.669& 25.380& 0.102& 7.300& 25.397& 0.092& 7.019& 3.861& 0.129& 5.693& 567.914& 0.077& 3.939\\
Phoneme    &433 &1083   & 30.392& 14.805& 0.103& 6.170& 14.697& 0.080& 6.591& 2.697& 0.087& 5.332& 527.290& 0.060& 2.722\\
Pima       &62  &155    & 8.231& 0.921& 0.011& 0.737& 0.978& 0.010& 0.681& 0.260& 0.009& 0.651& 2.904& 0.007& 0.092\\
Ripley     &100 &250    & 1.285& 0.442& 0.010& 1.175& 0.474& 0.008& 1.090& 0.281& 0.011& 1.224& 7.879& 0.006& 0.114\\
Satimage   &515 &1288   & 44.915& 12.342& 0.084& 7.366& 12.523& 0.091& 7.963& 2.743& 0.083& 6.917& 926.585& 0.156& 4.367\\
Seeds      &17  &43     & 0.172& 0.065& 0.005& 0.176& 0.087& 0.008& 0.171& 0.055& 0.006& 0.235& 0.183& 0.006& 0.034\\
Two Moons  &64  &160    & 0.808& 0.316& 0.007& 0.664& 0.321& 0.006& 0.573& 0.134& 0.007& 0.649& 3.103& 0.004& 0.054\\
Vehicle    &68  &170    & 6.266& 1.701& 0.012& 0.858& 1.814& 0.011& 0.661& 0.424& 0.016& 0.655& 3.894& 0.009& 0.151\\
Vowel      &80  &200    & 21.439& 3.782& 0.027& 1.249& 3.767& 0.024& 1.190& 0.458& 0.036& 0.858& 5.082& 0.019& 0.208\\
Wine       &15  &38     & 0.807& 0.130& 0.004& 0.125& 0.140& 0.007& 0.164& 0.041& 0.007& 0.155& 0.161& 0.004& 0.019\\
Yeast      &119 &298    & 6.643& 1.521& 0.037& 1.311& 1.558& 0.028& 1.456& 0.572& 0.040& 1.259& 12.184& 0.023& 0.316\\
\hline
\hline
Mean & 139.7 & 349.4&13.430& 3.794& 0.026& 1.955& 3.838& 0.025& 2.025& 0.811& 0.028& 1.796& 121.938& 0.024& 0.753\\
\hline
\end{tabular}
\end{center}
\end{table*}

Finally, we briefly analyze the computational complexity of the RWM kernel in comparison to that of the RBF kernel. The RWM kernel is based on a density model whose parameters are estimated using VI that is comparable to an EM technique: First, the responsibilities (cf.\ Eq.~\ref{eq:resp}) for all samples are estimated (E-step) and second, the parameters of the posterior distribution (M-step) are adapted. Given $N$ samples and $K$ components, the computational cost of one VI-step is $\mathcal{O}(NK(D^2 + E)+K(D^3+E'))$. For most applications, we have $K \ll N$ and $D+E \ll N$. In case of the RWM kernel we have to calculate $K$ Mahalanobis distances and $2K$ responsibility values to compare two considered samples instead of one Euclidean distance in case of the RBF kernel. 

Altogether, we can state that the measured run-times allow for many real applications of the proposed kernel.

\subsection{Lessons Learned}

What lessons did we learn in our simulation experiments? 

First, it is a cumbersome and time-consuming task to optimally parameterize a classifier, especially, if we assume the presence of sparsely labeled data. One reason for this is the fact that often different parameter combinations yield the same ``good'' classification error on the validation set ($L_{val}$, e.g., with a size of only six samples in case of a binary classification problem in our experiment 1), but most of them show a bad performance on the test set. To solve this problem, we used additional information delivered by the expected error calculated with regard to the set $U$, see Fig.~\ref{fig:crossValidation}. Maybe, this approach is not the best solution. In addition, a classifier is difficult to use in practice, if many parameters have to be set by user. Therefore, it is very advantageous to rely on good parametrization heuristics that exist for the RBF and RWM kernels (cf.~Section~\ref{subsec:rwmKernel}).

Second, what kind of density estimation technique to capture structure information shall be used?
The used estimation technique, whether parametric or non-parametric, depends on the application itself and the considered data set. However, our experiments 1 -- 3 show that a parametric estimation is more robust when data generating processes with different class affiliations are overlapping or if the respective clusters are not clearly separable. An SVM combined with the RWM kernel performs very well even if clusters have non-convex shapes (cf.~Section~\ref{sec:results}). For density estimation we applied variational Bayesian inference that, e.g., determines the number of model components by itself, but VI has some adjustable parameters, too \cite{FS09}. However, these parameters can be determined offline and, in combination with our approach, in an unsupervised manner. Besides this, a density estimation such as Gaussian mixture models can also be used for additional tasks, e.g., anomaly detection.

Third, can the RWM kernel be applied to large data sets? In this article we only used rather small data sets due to the larger number of simulation experiments. The answer to this question depends on the GMM parameterization step based on VI. This step is influenced by the number of samples but also by the number of input dimensions (or, more precisely, the number of free parameters of the GMM). To address the former, appropriate sampling techniques can be adopted, to cope with the latter, the number of parameters can be reduced, e.g.\ by restricting the model to diagonal or isotropic covariance matrices. That is, the RWM kernel can be used on large data sets but this was not an issue in this article.

Fourth, in Section~\ref{subsec:rwmKernel} we avoided the discussion about PSD (positive semi-definite) kernels. Clearly, we do not provide a formal proof that the RWM similarity always leads a positive semi-definite kernel matrix such that the optimization problem has a unique solution \cite{Bur98}. However, for all data sets which we used in our experimental studies, we applied a test for positive semi-definiteness to the RWM kernel matrices ($20 \times 5 = 100$ matrices), with the result that each of them was positive semi-definite. 
Besides this, if a kernel is found to be indefinite, different approaches exist to transform the result such that it can be used to solve the optimization problem (see, e.g., \cite{WCZ05}).
Another approach is to use the efficient and numerically stable technique mentioned in Section~\ref{subsec:rwmKernel}.

\section{Conclusion and Outlook}
\label{sec:conclusion}

In this article we proposed and evaluated a new, data dependent kernel function for support vector machines, the responsibility weighted Mahalanobis (RWM) kernel. This kernel considers structure in the data by means of a parametric density modeling approach. We have investigated its properties by evaluating the kernel on a number of benchmark data sets. The key advantages of the RWM kernel can be summarized as follows:
\begin{itemize}
	\item It may lead to better performance (classification accuracy) than some other kernels on partially labeled data sets, i.e., it is well-suited for semi-supervised learning. This is due to the fact that parameters of the RWM kernel can be trained in an unsupervised way.
	\item It is easy to handle. This is due to the facts that (1) standard SVM implementations can be used by just providing them with the kernel matrix and (2) heuristics for the parametrization of the $C$ and $\gamma$ parameters in $C$-SVM known for RBF kernels can easily be adopted.
\end{itemize}

The work presented in this article encourages us to investigate the new kernel in much more detail in our future work. Important questions will be: 
How does the kernel perform in comparison to related approaches such as TSVM, or S$^3$VM (cf.\ Section~\ref{sec:overview})?
Can we use the new kernel in other kernel based techniques (e.g., one-class SVM, support vector regression)? How can we use available class information when we build up the density models, e.g., in a transductive learning step (cf.~\cite{RCS14})? 
Can a self-parametrizing variant of the VI training technique be realized, i.e., a technique that finds parameters based on an analysis of the structure of the training data? 
How can we modify the GMM modeling step to cope with large data sets?
Also, we have to investigate the theoretical properties and the limitations of the RWM kernel in more detail. These are mainly due to limitations of the density based modeling approach, e.g., a difficult parametrization with sparse data.
We expect that it will be possible to combine the advantages of parametric and non-parametric density modeling approaches. We also will adapt the RWM kernel in active learning processes \cite{RS13, RCS14}.
\biboptions{sort&compress}
\bibliographystyle{elsart-num-sort}
\bibliography{bib/bibliography}

\end{document}